\PassOptionsToPackage{table}{xcolor}

\documentclass[letterpaper]{article} 
\usepackage{aaai2027}                

\usepackage[hyphens]{url} 
\usepackage{graphicx}     
\usepackage{natbib}       
\usepackage{caption}      

\usepackage{microtype}
\usepackage{latexsym}
\usepackage{xcolor}

\usepackage{subcaption}
\usepackage{booktabs}
\usepackage{multirow}
\usepackage{graphbox}
\usepackage[export]{adjustbox}

\usepackage{amsmath}
\usepackage{amssymb}
\usepackage{amsfonts}
\usepackage{mathtools}
\usepackage{amsthm}

\usepackage{enumitem}
\usepackage{etoc}

\usepackage{algorithm}
\usepackage{algorithmic}

\usepackage{newfloat}
\usepackage{listings}

\DeclareCaptionStyle{ruled}{
    labelfont=normalfont,
    labelsep=colon,
    strut=off
} 

\floatstyle{ruled}
\newfloat{listing}{tb}{lst}{}
\floatname{listing}{Listing}

\usepackage[most]{tcolorbox}
\tcbuselibrary{skins,breakable,listings}

\newtcblisting{promptbox}[1][]{
    enhanced,
    listing only,
    breakable,
    colframe=black,
    colback=white,
    colbacktitle=white,
    coltitle=black,
    fonttitle=\bfseries,
    title={#1},
    sharp corners,
    boxrule=0.8pt,
    top=2pt,
    bottom=2pt,
    left=4pt,
    right=4pt,
    before skip=1em,
    after skip=1em,
    listing options={
        basicstyle=\small\ttfamily,
        numbers=none,
        xleftmargin=0pt,
        aboveskip=0pt,
        belowskip=0pt,
        breaklines=true,
        breakatwhitespace=true,
        columns=flexible,
        keepspaces=true,
        showstringspaces=false,
        tabsize=2
    }
}

\providecommand{\todo}[2][]{}

\theoremstyle{plain}

\theoremstyle{definition}

\theoremstyle{remark}

\usepackage[capitalize,noabbrev]{cleveref}

\providecommand{\hyperref}[2][]{#2}

\title{
MapTab: A Diagnostic Benchmark for Long-Horizon Multi-Criteria Multimodal Reasoning on Heterogeneous Topological Graphs
}

\author{
    Ziqiao Shang\textsuperscript{\rm 1,2}\equalcontrib,
    Lingyue Ge\textsuperscript{\rm 1,2}\equalcontrib,
    Zian Xu\textsuperscript{\rm 1,2},
    Zi-Jian Cheng\textsuperscript{\rm 1,2},\\
    Shi-Yu Tian\textsuperscript{\rm 1,2},
    Zhenyu Huang\textsuperscript{\rm 1,2},
    Wenbo Fu\textsuperscript{\rm 1,2},
    Weiming Wu\textsuperscript{\rm 1,2},\\
    Yang Chen\textsuperscript{\rm 1,2},
    Xiangwen Zhang\textsuperscript{\rm 3},
    Yulan Hu\textsuperscript{\rm 3},
    Bin Liu\textsuperscript{\rm 4},
    Lan-Zhe Guo\textsuperscript{\rm 1,2}\corresponding
}

\affiliations{
    \textsuperscript{\rm 1}National Key Laboratory for Novel Software Technology, Nanjing University, Nanjing, China\\
    \textsuperscript{\rm 2}School of Intelligence Science and Technology, Nanjing University, Suzhou, China\\
    \textsuperscript{\rm 3}AMAP, Alibaba Group\\
    \textsuperscript{\rm 4}School of Computing and Artificial Intelligence, Southwest Jiaotong University, Chengdu, China\\[1mm]
    \small{\equalcontrib\ Equal contribution \qquad \corresponding\ Corresponding author}\\
    \small{Corresponding author: guolz@lamda.nju.edu.cn}\\
    \small{\textbf{Code:} \url{https://github.com/Ziqiao-Shang/MapTab}}\\
    \small{\textbf{Dataset:} \url{https://huggingface.co/datasets/szq-nju/MapTab}}\\
    \small{\textbf{Project page:} \url{https://ziqiao-shang.github.io/MapTab-Leaderboard/}}
}

\begin{document}

\twocolumn[{%
\renewcommand\twocolumn[1][]{#1}%
\maketitle
\vspace{0.1em}
\begin{center}
    \centering
    \includegraphics[width=0.87\linewidth]{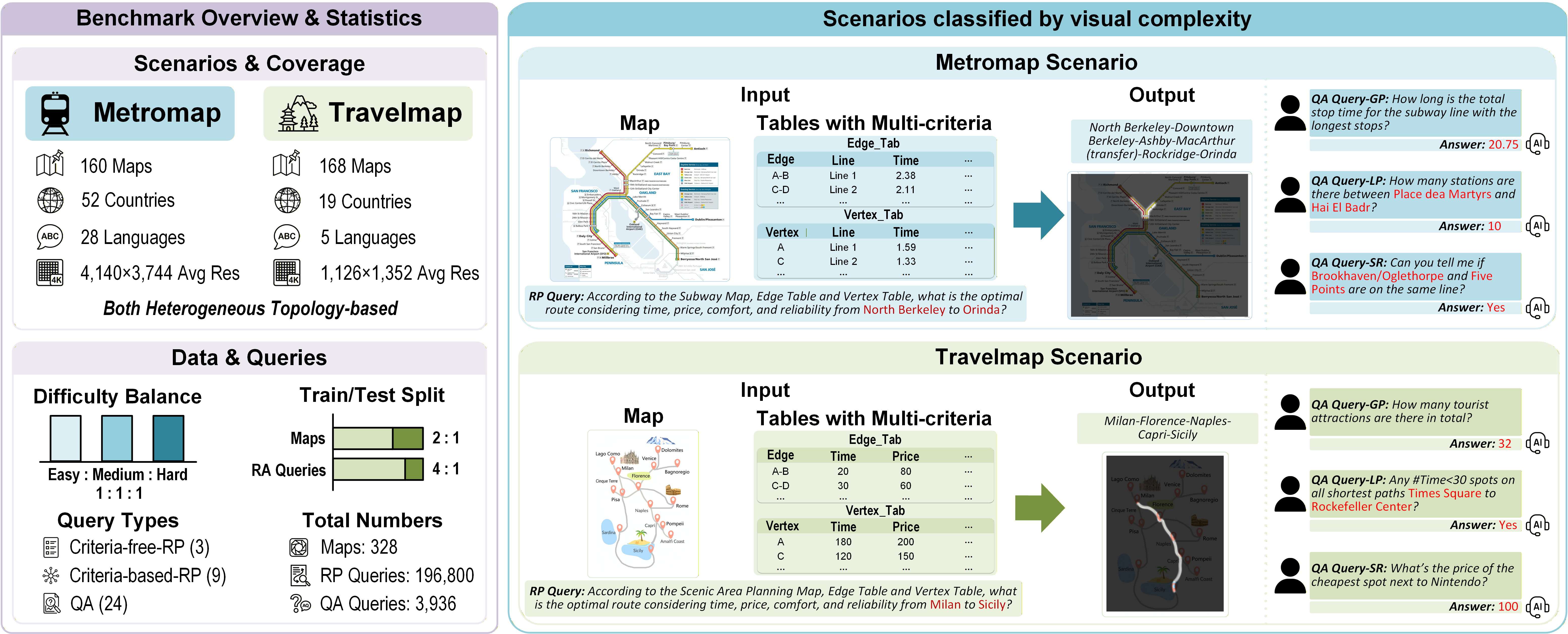}
    \captionsetup{width=0.87\linewidth}
    \captionof{figure}{
    Composition and Statistical Overview of the MapTab Benchmark.
    MapTab features 328 high-resolution maps across Metromap and Travelmap scenarios,
    providing 196,800 RP queries and 3,936 QA queries.
    }
    \label{fig:introduction}
\end{center}
\vspace{1em} 
}]

\begin{abstract}
  Systematically evaluating Multimodal Large Language Models (MLLMs) is essential for advancing Artificial General Intelligence (AGI). Yet existing benchmarks remain inadequate for rigorously measuring their reasoning capabilities under multi-criteria constraints. To address this gap, we introduce \textbf{MapTab}, a multimodal benchmark designed to assess holistic multi-criteria reasoning in MLLMs through route-planning tasks. MapTab requires models to perceive and ground visual information from map images while integrating route attributes, such as Time and Price, from structured tables. It covers two scenarios: \textbf{Metromap}, spanning metro networks in \textbf{160} cities across \textbf{52} countries, and \textbf{Travelmap}, featuring \textbf{168} representative tourist attractions from \textbf{19} countries. Overall, MapTab includes \textbf{328} images, \textbf{196,800} route-planning queries, and \textbf{3,936} QA queries, incorporating four key criteria: Time, Price, Comfort, and Reliability. Extensive evaluations of 21 representative MLLMs show that current models still struggle with multi-criteria multimodal reasoning. Notably, when visual perception is unreliable, multimodal reasoning can even underperform unimodal approaches. MapTab therefore offers a challenging and realistic testbed for systematically evaluating and advancing MLLMs across core perception, integration, numerical comparison, and route planning capabilities. Our code is available at \url{https://anonymous.4open.science/r/MapTab-23883}.
\end{abstract}

\section{Introduction}
Multimodal large language models (MLLMs)~\cite{bai2025qwen3vltechnicalreport} have demonstrated strong capabilities in integrating visual and textual information for complex reasoning and decision-making~\cite{hong2025embodied}. Accordingly, existing benchmarks have examined a wide range of abilities, including visual reasoning~\cite{wu2024v}, spatial understanding~\cite{qiao2025we}, and route planning~\cite{xing2025can}. However, these capabilities are often evaluated separately, and the ability of MLLMs to coordinate them in complex decision-making tasks remains insufficiently explored.

Map-based route planning (RP) provides a natural setting for evaluating such capability, as models must perceive spatial layouts, recover topological connectivity, and search for valid paths from map images. When RP is further extended to multi-criteria decision-making, models must consider not only graph connectivity but also quantitative factors such as time, price, comfort, and reliability. Since these attributes are not fully represented in map images, structured tables can provide the necessary complementary information, forming a heterogeneous multimodal task that requires visual understanding, table reasoning, cross-modal alignment, numerical comparison, and path planning.

Motivated by this gap, we propose MapTab, a multimodal benchmark for multi-criteria route planning over heterogeneous topological graphs. As shown in Fig.~\ref{fig:introduction}, MapTab contains \textbf{328} real-world map images from two scenarios: Metromap, covering \textbf{52} countries and \textbf{160} cities, and Travelmap, covering \textbf{19} countries and \textbf{168} tourist attractions. It combines map images that encode spatial topology with Vertex\_tab and Edge\_tab that provide node- and edge-level quantitative attributes. Since these numerical attributes are synthetically generated while the maps and topology are derived from real-world data, MapTab is a semi-synthetic benchmark. In total, it includes \textbf{16,400} origin--destination pairs, \textbf{196,800} RP queries, and \textbf{3,936} QA queries for diagnosing eight atomic capabilities underlying route planning.

In summary, our contributions are fourfold:
\begin{itemize}[itemsep=0pt, topsep=0pt]
    \item \textbf{Multimodal multi-criteria benchmark:} MapTab is the first benchmark to combine real-world map images and structured tables for multi-criteria route planning over heterogeneous graphs.
    \item \textbf{Capability-integrated RP tasks:} MapTab designs RP tasks that jointly require visual perception, table understanding, cross-modal alignment, topology reasoning, numerical reasoning, and path planning.
    \item \textbf{Atomic capability-diagnostic QA tasks:} MapTab introduces QA tasks that diagnose eight atomic capabilities underlying multimodal route planning.
    \item \textbf{Large-scale benchmark and evaluation:} MapTab contains \textbf{328} maps, \textbf{196,800} RP queries, and \textbf{3,936} QA queries, and evaluates \textbf{21} long-context MLLMs to reveal their key performance limitations.
\end{itemize}

\section{Related Works}\label{sec:related_work}
\subsubsection{Reasoning Abilities and Multimodal Benchmarks.}
Recent MLLMs have advanced from visual perception and cross-modal alignment toward explicit multimodal reasoning. Models such as Qwen3-VL and InternVL3 improve visual grounding and fine-grained understanding~\cite{bai2025qwen3vltechnicalreport,zhu2025internvl3}, while reinforcement learning and visual chain-of-thought further enhance mathematical, logical, spatial, and multimodal reasoning~\cite{guo2025deepseekr1,zhang2025thyme,zheng2025deepeyes,openai2025o3,team2025kimi}. Correspondingly, benchmarks have expanded from abstract visual reasoning~\cite{wu2024v,song2025visualpuzzles,xu2025visulogic} and general multimodal reasoning~\cite{wang2024measuring,yue2024mmmu,zhang2024mathverse} to map understanding and spatial reasoning~\cite{dihan2024mapeval,li2025mapqa,srivastava2025mapiq,zhou2024cartomark,xing2025can}. However, reasoning over complex topology, numerical attributes, and multiple decision criteria remains insufficiently evaluated.

\subsubsection{Map-Based Spatial Reasoning and Planning.}
Recent map-based benchmarks further advance from map understanding to spatial reasoning and navigation. CityBench, DriveBench, and GeoX-Bench, among others, evaluate urban cognition, spatial relations, and navigation-related understanding~\cite{feng2025citybench,xie2025vlms,pyo2025frieda,zheng2025geox,ung2025cartomapqa,cao2024maplm}. Planning-oriented benchmarks, including PlanAgent, NavBench, and GeoBenchX, introduce sequential planning, embodied execution, geographic tools, or environmental feedback~\cite{zheng2024planagent,zeng2024perceive,xu2025geonav,qiao2025navbench,krechetova2025geobenchx,fang2024travellm}. MapBench studies hierarchical map reading, while ReasonMap and RewardMap explore autonomous reasoning and reinforcement-learning-based optimization~\cite{xing2025can,feng2025can,feng2025rewardmap}. In contrast, MapTab is a large-scale, tool-free benchmark for topology-aware, multi-criteria route planning over heterogeneous metro and scenic-area maps. By incorporating criteria such as time, price, comfort, and reliability, it diagnoses MLLMs' limitations in map understanding, cross-modal integration, and decision-making rather than evaluating complete navigation systems. More details for Related Works are provided in Appendix A.

\section{Task Definition}
Given a map image $I$, an edge-attribute table $E$, a vertex-attribute table $V$, and a query $Q=(s,t,\mathbf{w})$, the task is to generate an optimal route from source $s$ to destination $t$. Here, $\mathbf{w}=(w_1,w_2,w_3,w_4)$ represents the relative importance of Time, Price, Comfort, and Reliability. For each query, only the selected criteria are assigned non-zero weights, which are randomly sampled from a predefined set and then normalized to sum to one. 

For a feasible route $r=(v_1,\ldots,v_k)$, where $v_1=s$ and $v_k=t$, Time and Price are summed over all traversed edges and included vertices. Comfort and Reliability are averaged over all non-zero edge and included-vertex values. We define that the source node is included in the counting, the destination node is excluded, and each transfer incurs an additional transfer-time cost. All attributes are normalized using predefined ranges shared across the benchmark.

The route cost is defined as
\begin{equation}
J_{\mathbf{w}}(r)
=
w_1T(r)+w_2P(r)
+w_3\bigl(1-C(r)\bigr)
+w_4\bigl(1-R(r)\bigr),
\end{equation}
where a lower value indicates a better route. The optimal route set is
\begin{equation}
\mathcal{R}^{*}(s,t)
=
\arg\min_{r\in\mathcal{R}(s,t)}J_{\mathbf{w}}(r),
\end{equation}
where $\mathcal{R}(s,t)$ contains all feasible routes from $s$ to $t$. The model may output any route in $\mathcal{R}^{*}(s,t)$; when multiple routes have the same minimum cost, all are treated as valid references.

\begin{figure*}[t!]
	\centering
	\includegraphics[width=0.75\linewidth]{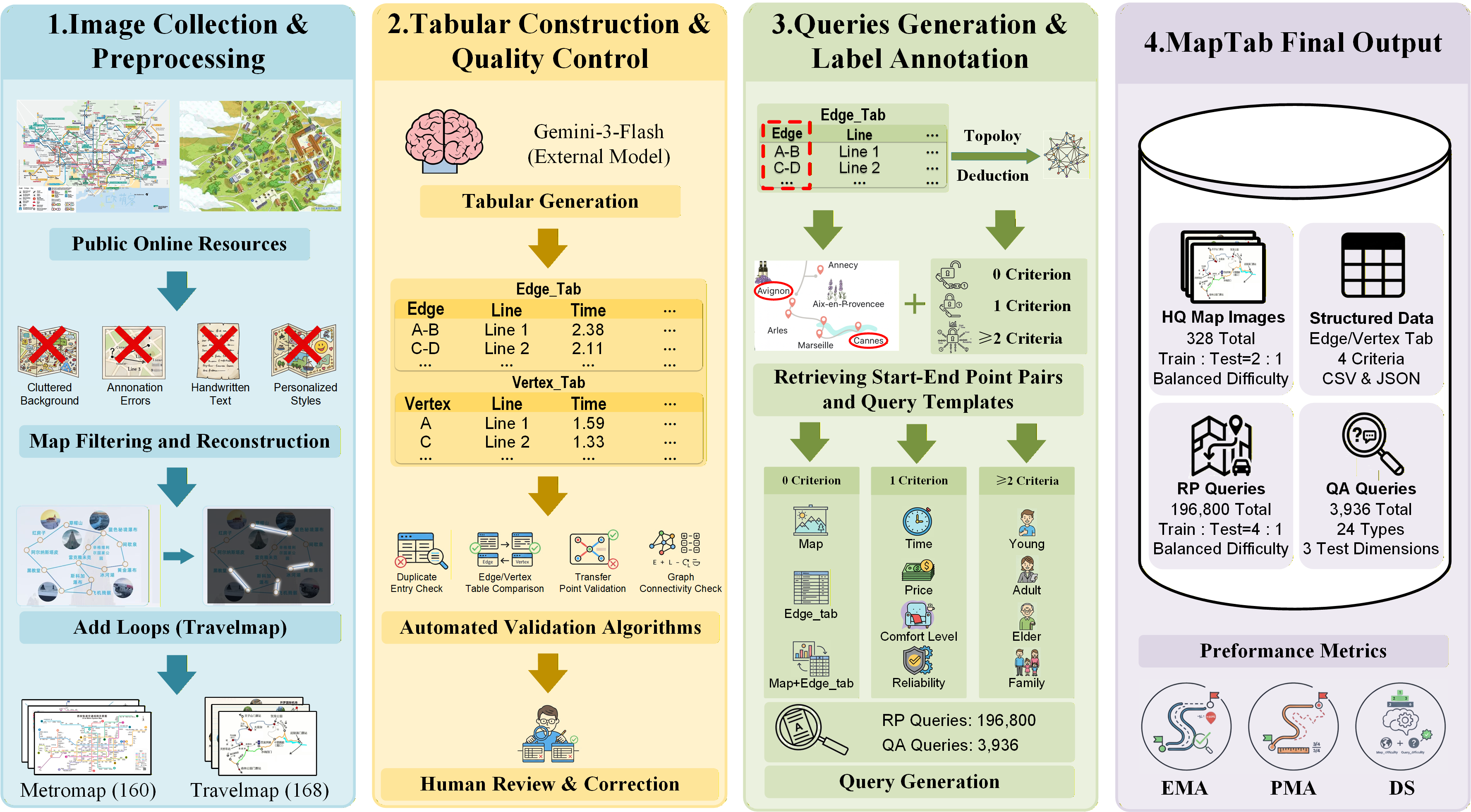}
    \captionsetup{width=0.8\linewidth}
    \captionof{figure}{
    Schematic overview of the MapTab construction pipeline, comprising 5 main steps: Image Collection \& Preprocessing, Tabular Construction, Quality Control, Query Generation, and Label Annotation.
    }
    \label{fig:Pipeline}
\end{figure*}

\begin{table*}[!ht]
\centering

\small
\renewcommand{\arraystretch}{1}
\setlength{\tabcolsep}{3.5pt}

\caption{
Capability taxonomy and corresponding problem types in the proposed MapTab benchmark. Each problem may involve multiple capabilities. M, E, V, and MM denote Map, Edge\_tab, Vertex\_tab, and Map+Mix\_tab, respectively.
}
\label{tab:capability-taxonomy}
\begin{tabular}{
@{}
>{\raggedright\arraybackslash}m{3.6cm}
>{\raggedright\arraybackslash}m{6.0cm}
>{\raggedright\arraybackslash}m{3.8cm}
>{\raggedright\arraybackslash}m{3.4cm}
@{}
}
\toprule

\multicolumn{1}{c}{\textbf{Capability}}
&
\multicolumn{1}{c}{\textbf{Definition}}
&
\multicolumn{1}{c}{\textbf{Metromap Problems}}
&
\multicolumn{1}{c}{\textbf{Travelmap Problems}}
\\

\midrule

\textbf{Visual Perception}
&
Identifying lines, stations, locations, and topological structures from map images.
&
M-GP, M-LP, M-SR
&
M-GP, M-LP, M-SR
\\
\specialrule{0.25pt}{0pt}{0pt}

\textbf{Table Understanding}
&
Retrieving and interpreting information and attributes from Vertex and Edge tables.
&
E-GP, E-LP, E-SR, V-GP, V-LP, V-SR
&
E-GP, E-LP, E-SR, V-GP, V-LP, V-SR
\\
\specialrule{0.25pt}{0pt}{0pt}

\textbf{Cross-modal Alignment}
&
Matching information across maps and tables and checking whether they are consistent.
&
MM-GP, MM-LP, MM-SR
&
MM-GP, MM-LP, MM-SR
\\
\specialrule{0.25pt}{0pt}{0pt}

\textbf{Graph Topology Reasoning}
&
Reasoning about graph structures, adjacency, connectivity, routes, and transfer relationships.
&
M-LP, E-SR, MM-SR
&
M-SR, E-SR
\\
\specialrule{0.25pt}{0pt}{0pt}

\textbf{Spatial Localization}
&
Determining locations, line membership, same-line relationships, and relative positions.
&
M-SR, V-LP, V-SR
&
M-LP, V-SR
\\
\specialrule{0.25pt}{0pt}{0pt}

\textbf{Numerical Reasoning}
&
Performing counting, comparison, aggregation, averaging, and other numerical operations.
&
M-GP, M-LP, E-GP, E-SR, V-GP, V-LP, MM-GP, MM-LP
&
M-GP, E-GP, V-GP, V-LP, MM-SR
\\
\specialrule{0.25pt}{0pt}{0pt}

\textbf{Path Planning}
&
Reasoning about shortest paths, transfer plans, and constraints along candidate routes.
&
MM-LP, MM-SR
&
E-SR, MM-LP, MM-SR
\\
\specialrule{0.25pt}{0pt}{0pt}

\textbf{Global Reasoning}
&
Integrating multiple reasoning steps, modalities, conditions, and global information.
&
MM-GP, MM-SR
&
MM-GP, MM-LP
\\

\bottomrule
\end{tabular}
\end{table*}

\section{MapTab Building Pipeline}
This study introduces MapTab, a multimodal benchmark for multi-criteria reasoning in Metromap and Travelmap scenarios. As shown in Fig.~\ref{fig:Pipeline}, we curate 328 maps with their Vertex\_tab and Edge\_tab. We construct \textbf{Mix\_tab} using Vertex\_tab as its main structure: the first column contains vertex names, while the remaining columns combine numerical values from Vertex\_tab and Edge\_tab. \textbf{The rows are independently shuffled for each sample to prevent order-based topology leakage.} All data and annotations undergo automated validation and human review, followed by deterministic generation of RP and QA queries.

\subsection{Benchmark Statistics}\label{sec:framework}
MapTab contains 328 topological maps from two scenarios. Metromap includes 160 maps from 52 countries and 32 native languages, rendered in 28 languages with an average resolution of $4{,}140 \times 3{,}744$. Travelmap includes 168 maps from 19 countries and 15 native languages, rendered in 5 languages with an average resolution of $1{,}126 \times 1{,}352$, and is visually simpler than Metromap. All maps are stratified into easy, medium, and hard levels at a 1:1:1 ratio and split into 218 training maps and 110 test maps. The split is strictly city/attraction-disjoint, and duplicate or near-duplicate maps are removed before splitting.

\paragraph{Route Planning.}
This subset contains 16,400 origin--destination pairs, including 8,000 from Metromap and 8,400 from Travelmap. The pairs are split 4:1 into 13,120 training and 3,280 test pairs. Each pair generates 3 criteria-free and 9 multi-criteria queries, resulting in 196,800 route-planning queries evenly distributed across the three difficulty levels. These queries evaluate topology understanding, route search, attribute integration, and multi-criteria optimization.

\paragraph{Diagnostic Question Answering.}
MapTab further provides 12 QA queries per map, yielding 3,936 queries in total. As shown in Tabel~\ref{tab:capability-taxonomy}, they assess eight capabilities required for route planning: \textbf{Visual Perception}, \textbf{Table Understanding}, \textbf{Cross-modal Alignment}, \textbf{Graph Topology Reasoning}, \textbf{Spatial Localization}, \textbf{Numerical Reasoning}, \textbf{Path Planning}, and \textbf{Global Reasoning}. This subset diagnoses the intermediate capability bottlenecks behind final route-planning performance. The QA queries are used only for evaluation and are not split by dataset or difficulty.

\subsection{Data Collection and Preprocessing}
\subsubsection{Map Image Collection and Preprocessing}\label{sec:map_curation}
As shown in Fig.~\ref{fig:Pipeline}, we collect high-resolution Metromap and Travelmap images from public online resources and remove maps with cluttered backgrounds, annotation errors, handwritten content, or highly personalized styles. For Travelmap, loop connections are added when necessary to improve the overall topological completeness.

\subsubsection{Multi-Criteria Tabular Construction}
We use Gemini-3-Flash~\cite{google2025gemini3flash} to construct \textbf{Edge\_tab} and \textbf{Vertex\_tab} from the extracted topology, providing edge- and node-level attributes while reducing repetitive annotation. Gemini is used only to assist repetitive manual annotation under predefined fields, value ranges, and generation rules, rather than to freely generate benchmark content. All topology is retained from the original maps, so this process does not alter the underlying topological distribution or introduce model-specific structural bias. The generated tables are automatically checked for duplicated entities, edge--vertex consistency, transfer validity, and graph connectivity, followed by manual review and correction.

\subsubsection{Quality Control}\label{sec:quality_control}
To ensure the accuracy of Edge\_tab and Vertex\_tab, we implement dedicated validation algorithms for Metromap and Travelmap (see Appendix B.5 and B.6) to pre-verify outputs from Gemini-3-Flash-Preview. Each result is then manually checked, ensuring reliability across the entire data pipeline.

\subsection{Query Generation and Label Annotation}\label{sec:query_gen}
As shown in Fig.~\ref{fig:Pipeline}, MapTab generates two complementary query sets: route-planning (RP) queries and diagnostic QA queries. Edge\_tab is first used to recover graph topology and retrieve valid origin--destination pairs, after which query templates and reference labels are generated automatically.

\paragraph{Route-Planning Queries.}
RP queries cover three levels of decision complexity: zero criterion, one criterion, and multiple criteria. Criteria-free queries use three input settings: \textit{Map-only}, \textit{Edge\_tab-only}, and \textit{Map+Edge\_tab}. Single-criterion queries optimize \textit{Time}, \textit{Price}, \textit{Comfort Level}, or \textit{Reliability}, while multi-criteria queries combine these attributes to represent different user preferences. Reference routes are derived from graph topology and the relevant numerical attributes. MapTab contains 12 RP query categories and 196,800 RP queries in total. Detailed sampling, templates, optimization, and label construction are provided in Appendix D.1.

\paragraph{Diagnostic QA Queries.}
MapTab further constructs 12 QA queries for each map, yielding 3,936 queries across 24 task types. These tasks span global perception-based reasoning, local perception-based reasoning, and spatial relationship judgment, and diagnose eight capabilities in Table~\ref{tab:capability-taxonomy} required throughout route planning. Answers are generated automatically when deterministic rules are available; otherwise, model-assisted annotations are manually verified. Further details are provided in Appendix D.2.

\begin{table*}[t]
  \centering

  \renewcommand{\arraystretch}{0.85}
  \setlength{\tabcolsep}{1.4mm}
  \small

  \begin{tabular}{@{}l c ccc ccc ccc ccc@{}}
    \toprule
    \multirow{3}{*}{\textbf{Model}} & \multirow{3}{*}{\textbf{Type}} & \multicolumn{6}{c}{\textbf{Metromap}} & \multicolumn{6}{c}{\textbf{Travelmap}} \\
    \cmidrule(lr){3-8}
    \cmidrule(l){9-14}
    & & \multicolumn{3}{c}{\textbf{Map-only}} & \multicolumn{3}{c}{\textbf{Map+Mix\_tab}} & \multicolumn{3}{c}{\textbf{Map-only}} & \multicolumn{3}{c}{\textbf{Map+Mix\_tab}} \\
    \cmidrule(r){3-5}
    \cmidrule(lr){6-8}
    \cmidrule(lr){9-11}
    \cmidrule(l){12-14}
    & & \textbf{EMA} & \textbf{PMA} & \textbf{DS} & \textbf{EMA} & \textbf{PMA} & \textbf{DS} & \textbf{EMA} & \textbf{PMA} & \textbf{DS} & \textbf{EMA} & \textbf{PMA} & \textbf{DS} \\
    \midrule
    \textbf{Human} & -- & 85.1 & -- & -- & 62.2 & -- & -- & 91.7 & -- & -- & 72.8 & -- & -- \\
    \midrule
    \multicolumn{14}{l}{\textit{\textbf{Open-source Models}}} \\
    Qwen3-VL-8B-Instruct & No-Thinking & 2.75 & 17.58 & 103 & 4.69 & 21.87 & 182 & 19.29 & 42.50 & 1040 & 15.65 & 40.97 & 804 \\
    Qwen3-VL-8B-Thinking & Thinking & 5.12 & 20.99 & 188 & 6.38 & 22.93 & 270 & 22.62 & 45.94 & 4511 & 12.74 & 38.10 & 653 \\
    Qwen3-VL-2B-Instruct & No-Thinking & 0.94 & 15.14 & 35 & 2.00 & 17.82 & 78 & 8.45 & 34.30 & 450 & 3.15 & 30.69 & 159 \\
    Qwen2.5-VL-7B-Instruct & No-Thinking & 0.94 & 15.02 & 32 & 3.38 & 18.09 & 131 & 7.68 & 30.48 & 390 & 4.70 & 28.84 & 228 \\
    Phi-3.5-Vision-Instruct-4B & No-Thinking & 0.06 & 10.40 & 2 & 0.81 & 12.94 & 26 & 0.12 & 20.00 & 6 & 1.31 & 22.68 & 67 \\
    Phi-4-Multimodal-Instruct-6B & No-Thinking & 0.00 & 9.75 & 0 & 0.44 & 9.02 & 14 & 0.42 & 19.26 & 20 & 1.43 & 18.96 & 67 \\
    InternVL3-8B-Instruct & No-Thinking & 0.13 & 13.98 & 4 & 1.75 & 17.00 & 76 & 6.61 & 29.21 & 308 & 2.50 & 24.28 & 132 \\
    Qwen3-VL-30B-A3B-Instruct & No-Thinking & 3.31 & 19.26 & 129 & 6.75 & 26.22 & 288 & 17.86 & 44.15 & 976 & 9.70 & 37.93 & 517 \\
    Qwen3-VL-32B-Instruct & No-Thinking & 6.31 & 22.23 & 250 & 6.56 & 24.43 & 262 & 36.90 & 57.44 & 2091 & 21.67 & 47.34 & 1161 \\
    Qwen3-VL-32B-Thinking & Thinking & \textbf{13.31} & \textbf{29.43} & 558 & 9.19 & 28.89 & 381 & 39.17 & 58.84 & 2262 & 19.94 & 46.73 & 1082 \\
    Qwen3.5-9B & No-Thinking & 5.69 & 22.44 & 231 & 8.75 & 28.15 & 365 & 25.95 & 49.39 & 1424 & 16.49 & 42.32 & 892 \\
    Qwen3.6-35B-A3B & Thinking & 11.56 & 29.16 & 476 & \textbf{14.75} & \textbf{34.76} & 643 & \textbf{42.74} & \textbf{61.74} & 2461 & \textbf{27.08} & \textbf{49.28} & 1532 \\
    \midrule
    \multicolumn{14}{l}{\textit{\textbf{Closed-source Models}}} \\
    GPT-4o & No-Thinking & 6.63 & 25.61 & 257 & 11.31 & 31.11 & 469 & 16.85 & 40.98 & 930 & 12.08 & 38.07 & 675 \\
    GPT-4.1 & No-Thinking & 7.94 & 25.52 & 306 & 14.06 & 35.98 & 608 & 20.30 & 43.24 & 1077 & 15.06 & 40.67 & 793 \\
    GPT-5.5-Instant & No-Thinking & 43.63 & 64.27 & 2320 & \textbf{69.50} & \textbf{79.76} & 4021 & 68.15 & 78.51 & 4242 & 57.98 & 69.87 & 3535 \\
    Doubao-Seed-1.6-w/o-Thinking & No-Thinking & 8.13 & 24.60 & 315 & 13.81 & 35.61 & 579 & 33.04 & 54.15 & 1880 & 25.48 & 49.54 & 1406 \\
    Doubao-Seed-1.6-Thinking & Thinking & 12.06 & 30.49 & 512 & 22.03 & 42.48 & 1029 & 38.45 & 58.46 & 2295 & 25.30 & 48.90 & 1437 \\
    Qwen-VL-Plus-w/o-Thinking & No-Thinking & 4.81 & 21.83 & 186 & 6.94 & 27.69 & 288 & 30.60 & 52.64 & 1691 & 22.92 & 47.65 & 1262 \\
    Qwen-VL-Plus-Thinking & Thinking & 10.75 & 29.11 & 437 & 16.38 & 37.44 & 714 & 38.27 & 58.94 & 2194 & 23.21 & 47.18 & 1289 \\
    Gemini-3-Flash-Preview & No-Thinking & 37.06 & 57.15 & 1881 & 53.87 & 65.84 & 2976 & 60.00 & 73.20 & 3693 & 43.51 & 60.11 & 2687 \\
    Gemini-3.5-Flash & No-Thinking & \textbf{44.19} & \textbf{65.40} & 2435 & 60.50 & 71.67 & 3433 & \textbf{68.21} & \textbf{79.12} & 4186 & \textbf{58.93} & \textbf{71.30} & 3648 \\
    \bottomrule
  \end{tabular}

  \caption{Evaluation results of various Multimodal Large Language Models (MLLMs) on the MapTab route-planning task under the Metromap and Travelmap scenarios. EMA, PMA, and DS denote Exact Match Accuracy, Partial Match Accuracy, and Difficulty-aware Score, respectively. Map-only uses only the map image, while Map+Mix\_tab uses the map image together with Mix\_tab. Bold values indicate the best performance within the open-source and closed-source model groups, respectively.}
  \label{tab:combined-eval}
\end{table*}

\begin{table*}[t]
  \centering

  \renewcommand{\arraystretch}{0.85}
  \setlength{\tabcolsep}{0.85mm}
  \small

  \begin{tabular}{@{}l c cccccccc cccccccc@{}}
    \toprule
    \multirow{2}{*}{\textbf{Model}} & \multirow{2}{*}{\textbf{Type}} & \multicolumn{8}{c}{\textbf{Metromap}} & \multicolumn{8}{c}{\textbf{Travelmap}} \\
    \cmidrule(lr){3-10}
    \cmidrule(l){11-18}
    & & \textbf{VP} & \textbf{TU} & \textbf{CA} & \textbf{GT} & \textbf{SL} & \textbf{NR} & \textbf{PP} & \textbf{GR} & \textbf{VP} & \textbf{TU} & \textbf{CA} & \textbf{GT} & \textbf{SL} & \textbf{NR} & \textbf{PP} & \textbf{GR} \\
    \midrule
    \multicolumn{18}{l}{\textit{\textbf{Open-source Models}}} \\
    Qwen3-VL-8B-Instruct & No-Thinking & 48.5 & 54.4 & 20.6 & 21.3 & 70.6 & 29.4 & 30.6 & 19.7 & 40.1 & 52.2 & 53.4 & 49.1 & 60.7 & 25.7 & 43.3 & 72.9 \\
    Qwen3-VL-8B-Thinking & Thinking & 44.2 & 78.0 & 17.5 & 40.2 & 75.6 & 46.1 & 22.5 & 21.6 & 54.2 & 87.9 & 48.6 & 46.1 & 85.4 & 68.1 & 40.9 & 66.1 \\
    Qwen3-VL-2B-Instruct & No-Thinking & 25.4 & 32.7 & 11.7 & 11.7 & 46.3 & 8.2 & 17.5 & 13.4 & 27.0 & 41.9 & 45.4 & 37.2 & 48.5 & 14.1 & 45.6 & 66.1 \\
    Qwen2.5-VL-7B-Instruct & No-Thinking & 43.5 & 53.0 & 20.0 & 19.6 & 71.5 & 27.7 & 29.4 & 17.2 & 41.5 & 61.6 & 39.5 & 50.9 & 72.6 & 32.7 & 44.1 & 51.2 \\
    Phi-3.5-Vision-Instruct-4B & No-Thinking & 51.7 & 60.5 & 20.6 & 26.9 & 81.0 & 33.0 & 30.6 & 20.3 & 40.3 & 63.9 & 52.4 & 50.3 & 62.8 & 37.9 & 42.1 & 73.2 \\
    Phi-4-Multimodal-Instruct-6B & No-Thinking & 60.4 & 71.3 & 30.2 & 44.2 & 90.6 & 47.2 & 42.5 & 27.5 & \textbf{57.7} & 84.3 & 53.4 & 46.4 & 85.1 & 67.0 & 40.3 & 71.1 \\
    InternVL3-8B-Instruct & No-Thinking & 40.2 & 34.7 & 13.5 & 17.3 & 48.5 & 15.3 & 20.3 & 14.1 & 35.9 & 44.7 & 47.8 & 42.6 & 55.7 & 17.4 & 40.7 & 69.3 \\
    Qwen3-VL-30B-A3B-Instruct & No-Thinking & 32.5 & 23.4 & 9.8 & 11.9 & 43.5 & 6.4 & 14.7 & 9.1 & 21.0 & 40.9 & 31.0 & 29.5 & 47.9 & 14.1 & 30.2 & 44.9 \\
    Qwen3-VL-32B-Instruct & No-Thinking & 14.4 & 30.9 & 10.4 & 9.0 & 38.5 & 5.9 & 15.6 & 11.9 & 19.1 & 37.5 & 32.5 & 26.6 & 44.9 & 16.3 & 36.0 & 45.8 \\
    Qwen3-VL-32B-Thinking & Thinking & 32.1 & 40.2 & 22.5 & 28.3 & 53.5 & 13.7 & 33.8 & 30.3 & 30.6 & 41.4 & 47.2 & 33.9 & 58.3 & 12.9 & 39.1 & 67.9 \\
    Qwen3.5-9B & No-Thinking & 51.0 & 51.5 & 32.3 & 34.0 & 71.3 & 28.9 & 47.8 & 38.4 & 48.6 & 55.8 & 58.5 & 62.2 & 70.8 & 27.3 & 54.2 & 77.1 \\
    Qwen3.6-35B-A3B & Thinking & \textbf{61.9} & \textbf{89.6} & \textbf{52.1} & \textbf{73.8} & \textbf{93.5} & \textbf{62.8} & \textbf{70.6} & \textbf{55.3} & 54.2 & \textbf{95.8} & \textbf{60.3} & \textbf{70.8} & \textbf{87.5} & \textbf{69.2} & \textbf{59.5} & \textbf{78.0} \\
    \midrule
    \multicolumn{18}{l}{\textit{\textbf{Closed-source Models}}} \\
    GPT-4o & No-Thinking & 51.7 & 65.6 & 26.0 & 29.0 & 85.6 & 37.9 & 37.2 & 24.4 & 41.5 & 67.3 & 52.0 & 42.9 & 65.5 & 44.5 & 40.3 & 72.3 \\
    GPT-4.1 & No-Thinking & 55.0 & 72.7 & 26.0 & 38.1 & 86.7 & 44.3 & 37.5 & 26.3 & 43.1 & 70.9 & 56.2 & 49.4 & 69.6 & 46.9 & 46.0 & 75.3 \\
    GPT-5.5-Instant & No-Thinking & 77.7 & \textbf{99.3} & 82.5 & 92.1 & \textbf{97.1} & 86.0 & 89.4 & 82.8 & 72.6 & \textbf{99.7} & \textbf{70.2} & \textbf{90.5} & \textbf{95.8} & 75.4 & \textbf{76.4} & 89.0 \\
    Doubao-Seed-1.6-w/o-Thinking & No-Thinking & 50.8 & 74.2 & 34.6 & 43.5 & 87.3 & 46.9 & 50.0 & 27.2 & 43.9 & 77.9 & 56.2 & 50.9 & 78.9 & 53.2 & 51.4 & 71.4 \\
    Doubao-Seed-1.6-Thinking & Thinking & 57.5 & 89.5 & 43.5 & 54.4 & 92.1 & 63.7 & 51.6 & 40.3 & 50.6 & 91.6 & 57.5 & 63.7 & 85.7 & 68.0 & 55.2 & 75.3 \\
    Qwen-VL-Plus-w/o-Thinking & No-Thinking & 53.5 & 70.0 & 22.5 & 34.2 & 84.6 & 41.4 & 33.1 & 20.9 & 48.2 & 70.6 & 48.0 & 44.3 & 77.4 & 48.9 & 41.9 & 62.2 \\
    Qwen-VL-Plus-Thinking & Thinking & 61.5 & 87.7 & 38.5 & 57.3 & 92.5 & 61.1 & 50.9 & 34.7 & 51.6 & 88.9 & 55.4 & 48.5 & 83.9 & 69.5 & 43.9 & 70.8 \\
    Gemini-3-Flash-Preview & No-Thinking & 78.3 & 91.4 & 74.2 & 84.2 & 95.8 & 77.5 & 87.2 & 71.3 & 69.6 & 97.1 & 64.5 & 82.2 & 92.6 & 73.8 & 64.7 & 83.6 \\
    Gemini-3.5-Flash & No-Thinking & \textbf{79.6} & 99.2 & \textbf{89.4} & \textbf{94.0} & 95.8 & \textbf{89.7} & \textbf{89.7} & \textbf{92.8} & \textbf{75.4} & 98.7 & 69.4 & 84.8 & 94.9 & \textbf{77.9} & 73.0 & \textbf{89.3} \\
    \bottomrule
  \end{tabular}

  \caption{
  Capability-level performance of different MLLMs on the MapTab benchmark.
  Each score is the arithmetic mean accuracy over all problems associated with the corresponding capability.
  VP, TU, CA, GT, SL, NR, PP, and GR denote Visual Perception, Table Understanding,
  Cross-modal Alignment, Graph Topology Reasoning, Spatial Localization,
  Numerical Reasoning, Path Planning, and Global Reasoning, respectively.
  Bold values indicate the best performance within the open-source and closed-source model groups, respectively.
  }
  \label{tab:capability-performance-transposed}
\end{table*}

\section{Experiments}
\subsection{Experimental Setups}
This section systematically evaluates 21 state-of-the-art long-context MLLMs on MapTab, covering both open- and closed-source models as well as instruction-following and reasoning-enhanced variants. We compare their performance on RP and QA tasks across the Metromap and Travelmap scenarios under consistent prompts, inference settings, and multimodal input preprocessing. The complete model list and experimental details are provided in Appendix F.1.

\subsection{Metrics.} 
For model performance evaluation, RP tasks are assessed using three metrics:
\begin{itemize}[itemsep=0pt, topsep=0pt]
    \item \textbf{Exact Match Accuracy (EMA):} EMA evaluates whether the generated route matches any optimal reference path in station order and route content, while allowing minor spelling variations.
    \item \textbf{Partial Match Accuracy (PMA):} PMA measures the longest contiguous correct prefix from the origin. When multiple optimal paths exist, the highest PMA is used.
    \item \textbf{Difficulty-aware Score (DS):} DS weights exact-match results by the combined Map\_Difficulty and Query\_Difficulty levels.
\end{itemize}

Detailed definitions of these metrics are provided in Appendix C.1. In addition, we design multiple indicators to more comprehensively characterize Map\_Difficulty and Query\_Difficulty in Appendix C.2.

\subsection{Experimental Results}
\subsubsection{Performance of MLLMs on RP}\label{subsec:performance_RP}

Table~\ref{tab:combined-eval} reports the RP performance of different MLLMs under the Map-only and Map+Mix\_tab settings on Metromap and Travelmap. Results for the Edge\_tab-only setting and detailed analyses of cross-modal information fusion are provided in Appendix F.2.

\textbf{From the experimental results}, the role of Mix\_tab depends strongly on perceptual difficulty. In the visually dense Metromap scenario, Map+Mix\_tab improves most models; for example, Gemini-3.5-Flash increases from 44.19\% to 60.50\% EMA. Here, Mix\_tab mainly provides symbolic anchors that reduce OCR, entity recognition, and topology errors, although the remaining perceptual burden still limits multi-criteria optimization. In the simpler Travelmap scenario, models can perceive the map more reliably and attend more to the decision attributes in Mix\_tab. However, performance often declines after adding the table, indicating that multi-criteria integration becomes the main bottleneck once perception is less challenging.

\textbf{From the model comparison}, Gemini-3.5-Flash and GPT-5.5-Instant achieve the strongest closed-source results, while Qwen3.6-35B-A3B leads the open-source models in most settings. Their advantage likely comes from stronger visual understanding, cross-modal alignment, and reasoning consistency. Thinking models also generally outperform their Instruct counterparts, especially on Metromap, showing that CoT helps decompose routes, track topology and constraints, and sustain long-horizon planning. Nevertheless, a substantial gap remains between the strongest open-source models and closed-source models.

\textbf{Humans} substantially outperform MLLMs without multi-criteria information, achieving 85.1\% on Metromap and 91.7\% on Travelmap. Their performance drops to 62.2\% and 72.8\% after introducing multiple criteria, indicating the additional difficulty of comparing route attributes. The participant pool consisted of all authors, each of whom independently completed multiple sampled tasks with access to scratch paper, calculators, and translation tools.

\subsubsection{Performance of MLLMs on QA}\label{subsec:qa_results}
Table~\ref{tab:capability-performance-transposed} presents the results obtained by mapping the 24 QA questions to the eight atomic capability categories defined in Table~\ref{tab:capability-taxonomy} and averaging the accuracy of the questions associated with each capability. Detailed results for all 24 QA questions are provided in Appendix F.3.

\textbf{Observation 1: Models remain weak in cross-modal alignment, topology reasoning, numerical reasoning, and global reasoning.} Cross-modal alignment is challenging because route structures are shown in maps, while their attributes are stored in separate tables, making precise entity matching essential. Topology reasoning is hindered by dense lines, transfer stations, and visually crossing but disconnected routes, which can cause incorrect connectivity judgments. Numerical reasoning requires models to aggregate route attributes and consistently handle weights, normalization, and opposite optimization directions. Global reasoning is difficult because models must compare multiple feasible routes over the entire graph, yet they often favor locally salient or seemingly shortest paths.

\textbf{Observation 2: Errors accumulate across the multimodal reasoning pipeline.} Most atomic capabilities still fall short of reliable accuracy. Since multi-criteria route planning requires perception, cross-modal alignment, and numerical computation to work jointly, errors at any stage can propagate to subsequent reasoning and alter the selected route. Consequently, end-to-end planning performance is often substantially lower than the performance of individual capabilities. The near-zero Total Time accuracy is not an evaluation artifact: manual inspection and sampled script verification show that models often omit edge travel time, vertex dwell time, or conditional transfer time, and any missing component leads to an incorrect total. Additional verified examples are provided in Appendix H.

\textbf{Observation 3: MapTab evaluates end-to-end intelligent navigation rather than route optimization alone.} The gap between topology reasoning and path planning shows that recovering the graph structure does not guarantee a correct route. Traditional navigation systems mainly optimize paths over predefined graphs and reliable attributes, but cannot independently understand raw maps, extract structured information, or align multimodal evidence. MapTab instead requires MLLMs to complete the full process from multimodal understanding to multi-criteria route planning, and therefore evaluates a practical end-to-end intelligent navigation capability rather than graph algorithms alone.

\begin{table*}[!t]
    \centering
    
    \renewcommand{\arraystretch}{0.85}
    \setlength{\tabcolsep}{4.8pt}
    
    \small
    \begin{tabular}{lcccccccccccc}
    \toprule
    \multicolumn{1}{c}{\multirow{2}{*}{\textbf{Criteria Settings}}} & \multicolumn{2}{c}{\textbf{FR}} & \multicolumn{2}{c}{\textbf{PR}} & \multicolumn{2}{c}{\textbf{NR}} & \multicolumn{6}{c}{\textbf{All}} \\
    \cmidrule(lr){2-3}
    \cmidrule(lr){4-5}
    \cmidrule(lr){6-7}
    \cmidrule(lr){8-13}
    & \textbf{EMA} & \textbf{PMA} & \textbf{EMA} & \textbf{PMA} & \textbf{EMA} & \textbf{PMA} & \textbf{EMA} & \textbf{PMA} & \textbf{T} & \textbf{P} & \textbf{C} & \textbf{R} \\
    \midrule
    \multicolumn{13}{l}{\textit{\textbf{Metromap}}} \\
    Time-Only & 26.59 & 46.75 & 1.92 & 16.36 & 1.80 & 21.34 & 19.81 & 39.47 & 0.00 & 2.44 & 2.88 & 9.69 \\
    Price-Only & 30.91 & 51.05 & 0.00 & 13.87 & 3.31 & 25.94 & 22.06 & 42.40 & 0.00 & 1.19 & 4.06 & 10.44 \\
    Comfort\_Level-Only & 25.31 & 46.24 & 0.00 & 16.20 & 2.56 & 23.81 & 18.25 & 38.97 & 0.00 & 2.44 & 4.88 & 10.19 \\
    Reliability-Only & 25.73 & 45.57 & 0.00 & 13.84 & 3.59 & 27.69 & 19.38 & 39.53 & 0.00 & 1.19 & 4.06 & 10.44 \\
    Time+Price+Reliability & 24.73 & 44.79 & 1.52 & 15.44 & 2.35 & 23.12 & 19.25 & 38.91 & 0.00 & 2.19 & 3.87 & 9.81 \\
    Time+Comfort\_Level+Reliability & 24.51 & 44.74 & 1.53 & 15.22 & 1.63 & 23.31 & 19.13 & 39.04 & 0.00 & 2.00 & 4.88 & 10.19 \\
    Price+Comfort\_Level+Reliability & 26.06 & 47.06 & 0.00 & 12.68 & 3.36 & 26.23 & 19.44 & 40.18 & 0.00 & 1.75 & 4.31 & 11.00 \\
    Time+Price+Comfort\_Level+Reliability & 24.35 & 44.42 & 0.75 & 17.14 & 3.04 & 24.82 & 19.31 & 39.31 & 0.00 & 2.13 & 5.19 & 10.69 \\
    \textbf{Average} & \textbf{25.96} & \textbf{46.26} & \textbf{0.73} & \textbf{15.08} & \textbf{2.74} & \textbf{24.59} & \textbf{19.58} & \textbf{39.73} & \textbf{0.00} & \textbf{1.92} & \textbf{4.27} & \textbf{10.31} \\
    \midrule
    \multicolumn{13}{l}{\textit{\textbf{Travelmap}}} \\
    Time-Only & 36.24 & 50.57 & 8.40 & 29.30 & 9.09 & 29.02 & 29.94 & 45.70 & 0.00 & 0.12 & 1.07 & 4.46 \\
    Price-Only & 45.20 & 62.32 & 13.36 & 36.56 & 15.49 & 39.44 & 38.69 & 57.11 & 0.06 & 0.06 & 1.25 & 4.05 \\
    Comfort\_Level-Only & 39.05 & 59.99 & 18.45 & 44.78 & 12.24 & 38.47 & 34.17 & 56.28 & 0.12 & 0.00 & 1.43 & 5.00 \\
    Reliability-Only & 43.32 & 62.76 & 19.18 & 44.42 & 8.73 & 36.12 & 37.20 & 58.08 & 0.06 & 0.06 & 1.19 & 3.87 \\
    Time+Price+Reliability & 39.81 & 59.41 & 12.36 & 37.07 & 7.84 & 33.60 & 34.35 & 54.97 & 0.12 & 0.06 & 1.07 & 4.29 \\
    Time+Comfort\_Level+Reliability & 38.63 & 59.07 & 13.24 & 37.85 & 14.81 & 36.26 & 33.75 & 54.90 & 0.12 & 0.00 & 1.49 & 4.35 \\
    Price+Comfort\_Level+Reliability & 41.03 & 60.17 & 11.90 & 37.06 & 12.50 & 36.78 & 35.42 & 55.69 & 0.06 & 0.06 & 1.67 & 4.11 \\
    Time+Price+Comfort\_Level+Reliability & 39.28 & 59.41 & 14.75 & 39.42 & 13.33 & 39.07 & 34.52 & 55.56 & 0.06 & 0.18 & 1.49 & 4.82 \\
    \textbf{Average} & \textbf{40.32} & \textbf{59.23} & \textbf{13.91} & \textbf{38.27} & \textbf{11.25} & \textbf{35.57} & \textbf{34.76} & \textbf{54.79} & \textbf{0.08} & \textbf{0.07} & \textbf{1.33} & \textbf{4.37} \\
    \bottomrule
    \end{tabular}
    
    \caption{
    Ablation studies of route repetition and criteria understanding in the Metromap and Travelmap scenarios.
    Query types are categorized into Fully Repeat (FR), Partially Repeat (PR), and Not Repeat (NR), according to the deviation between the original shortest paths and the optimal paths after introducing different criteria.
    Under All, EMA and PMA report the overall route planning performance, while T, P, C, and R denote the understanding accuracy of Total Time, Total Price, Average Comfort Level, and Average Reliability, respectively.
    }
    \label{tab:route-repetition-criteria-ablation}
\end{table*}

\subsubsection{Ablation Study for Multi-Criteria}

Table~\ref{tab:capability-performance-transposed} evaluates whether Qwen3-VL-8B-Instruct truly uses the given criteria rather than simply selecting the shortest path. With fixed start and end stations, samples are divided into three categories: \textbf{Fully Repeat (FR)}, where the criteria-optimal path is the shortest path; \textbf{Partially Repeat (PR)}, where it matches one of multiple shortest paths; and \textbf{Not Repeat (NR)}, where it differs from all shortest paths. We also require the model to report Total Time (T), Total Price (P), Average Comfort (C), and Average Reliability (R) to evaluate its criterion understanding and computation.

We analyze the results from four aspects:

\textbf{Observation 1: ``Shortest-path trap''.} Performance is mainly concentrated in the FR category. On Metromap, the average EMA values for PR and NR are only 0.73\% and 2.74\%, respectively, compared with 25.96\% for FR. Similar results are observed on Travelmap. This suggests that models mainly rely on the original shortest path and often succeed only when it happens to coincide with the criteria-optimal path, rather than truly using the given criteria.

\textbf{Observation 2: Counting, numerical computation, and multi-step reasoning deficiencies.}
The low accuracy of T, P, C, and R shows that models struggle to correctly calculate route-level criteria. Compared with criteria involving only a single computation, performance drops sharply when transfer time is introduced as an additional condition in total-time calculation. This shows that models may handle isolated numerical operations, but struggle to preserve correctness once an additional dependency is introduced, revealing a substantial gap between single-step and multi-step reasoning.

\textbf{Observation 3: Limited understanding of weighted and heterogeneous graphs.} Model performance remains low even under single-criterion settings, with limited improvement after reducing the number of criteria. This indicates that the main difficulty lies not only in combining multiple criteria, but also in understanding and reasoning over weighted relations in heterogeneous graphs.

\subsubsection{Other Experimental Results}
The following summarizes 4 groups of experimental results presented in Appendix G:

\begin{itemize}[itemsep=0pt, topsep=0pt]
    \item \textit{Impact of image resolution on RP and QA tasks} \textbf{(Appendix G.1)}: While structured table understanding provides a reliable lower bound for model performance, effective image understanding is what determines the achievable upper bound.
    
    \item \textit{Analysis across different Map Difficulty and Query Difficulty levels} \textbf{(Appendix G.2)}: Performance degradation trends vary substantially with increasing Map and Query Difficulty levels, especially between Metromap scenario and Travelmap scenario.
    
    \item \textit{Impact of tabular modality formats on RP and QA tasks} \textbf{(Appendix G.3)}: The CSV format significantly reduces token overhead without inducing any noticeable performance degradation.
    
    \item \textit{Impact of language distribution on RP tasks} \textbf{(Appendix G.4)}: Models perform better when the image language is high-frequency languages (e.g., Chinese or English), while differences in native language have a relatively limited impact on performance.
\end{itemize}

Moreover, Appendix H presents a more detailed error case analysis to reveal representative model failure patterns, Appendix I summarizes the current limitations of MapTab in benchmark design and evaluation, and Appendix J discusses several promising directions for future research on multimodal multi-criteria reasoning.

\subsubsection{Route Planning Error Case Analysis}
\begin{table}[t]
\centering

\renewcommand{\arraystretch}{0.85}
\setlength{\tabcolsep}{1pt}

\small
\begin{tabular}{lccc}
\toprule
\textbf{Error Type} &
\textbf{Count} &
\textbf{Error (\%)} &
\textbf{Overall (\%)} \\
\midrule
Perception & 267 & 20.7 & 16.7 \\
Transfer Indeterminate & 109 & 8.4 & 6.8 \\
Path Search & 460 & 35.6 & 28.8 \\
Topology Extraction & \textbf{491} & \textbf{38.0} & \textbf{30.7} \\
Output Format & 34 & 2.6 & 2.1 \\
Table Lookup & 39 & 3.0 & 2.4 \\
\bottomrule
\end{tabular}

\caption{
Distribution of different error types of Qwen3-VL-8B on MetroMap.
Error (\%) denotes the percentage among all error cases,
while Overall (\%) denotes the percentage among the entire evaluation set. Multiple error labels may be assigned to the same prediction; therefore, percentages across categories do not sum to 100\%.
}
\label{tab:error_distribution}
\end{table}

Table~\ref{tab:error_distribution} shows that \textbf{Topology Extraction} (38.0\%) and \textbf{Path Search} (35.6\%) account for most failures, identifying topology reasoning and route planning as the main bottlenecks. \textbf{Perception} errors contribute another 20.7\%, while \textbf{Transfer Indeterminate}, \textbf{Output Format}, and \textbf{Table Lookup} are much less frequent. Overall, current MLLMs struggle more with understanding and reasoning over map structures than with table lookup or output formatting.

The appendix~ref{xx} provides detailed analyses and examples for each error type. The cases show that failures arise at multiple stages: perception errors affect station recognition and visual grounding, while topology reconstruction and incomplete path exploration remain the dominant sources of routing errors. Moreover, a correct station sequence does not guarantee accurate transfer tracking or table understanding, and unnecessary reasoning steps may lead to unstable termination or invalid outputs.

\section{Conclusion}
This study introduces \textbf{MapTab}, a multimodal benchmark for holistic multi-criteria reasoning over heterogeneous topological graphs by integrating \emph{visual maps} and \emph{structured tables}. Across \textit{Metromap} and \textit{Travelmap}, MapTab contains 328 de-duplicated maps, 196,800 RP samples, and 3,936 diagnostic QA samples. Its \textbf{controlled construction}, \textbf{deterministic reference generation}, and complementary evaluations enable \textbf{fair, reproducible, and interpretable comparisons} across 21 MLLMs. The results consistently expose bottlenecks in visual perception, cross-modal alignment, numerical computation, and global route planning, while separating perceptual limitations from reasoning failures. Rather than overextending to all real-world navigation settings, MapTab provides a \textbf{rigorous, well-scoped, and diagnostically informative testbed} for studying capability coordination under explicit decision criteria. Despite limitations in map diversity, synthetic attributes, and static evaluation, it establishes a reliable foundation for dynamic benchmarks, modular perception--reasoning systems, selective tool use, and targeted multimodal post-training.
\bibliography{aaai2027}

@misc{bai2025qwen3vltechnicalreport,
      title={Qwen3-VL Technical Report}, 
      author={Shuai Bai and Yuxuan Cai and Ruizhe Chen and Keqin Chen and Xionghui Chen and Zesen Cheng and Lianghao Deng and Wei Ding and Chang Gao and Chunjiang Ge and Wenbin Ge and Zhifang Guo and Qidong Huang and Jie Huang and Fei Huang and Binyuan Hui and Shutong Jiang and Zhaohai Li and Mingsheng Li and Mei Li and Kaixin Li and Zicheng Lin and Junyang Lin and Xuejing Liu and Jiawei Liu and Chenglong Liu and Yang Liu and Dayiheng Liu and Shixuan Liu and Dunjie Lu and Ruilin Luo and Chenxu Lv and Rui Men and Lingchen Meng and Xuancheng Ren and Xingzhang Ren and Sibo Song and Yuchong Sun and Jun Tang and Jianhong Tu and Jianqiang Wan and Peng Wang and Pengfei Wang and Qiuyue Wang and Yuxuan Wang and Tianbao Xie and Yiheng Xu and Haiyang Xu and Jin Xu and Zhibo Yang and Mingkun Yang and Jianxin Yang and An Yang and Bowen Yu and Fei Zhang and Hang Zhang and Xi Zhang and Bo Zheng and Humen Zhong and Jingren Zhou and Fan Zhou and Jing Zhou and Yuanzhi Zhu and Ke Zhu},
      year={2025},
      eprint={2511.21631},
      archivePrefix={arXiv},
      primaryClass={cs.CV},
      url={https://arxiv.org/abs/2511.21631}, 
}

@misc{google2025gemini3flash,
  title = {Gemini 3 Flash: Frontier Intelligence Built for Speed},
  author = {Google},
  year = {2025},
  month = {December},
  howpublished = {\url{https://blog.google/products-and-platforms/products/gemini/gemini-3-flash/}},
  note = {Model ID: gemini-3-flash-preview. Accessed: 2025-12-25}
}

@misc{google2026gemini35flashdoc,
  title = {Gemini 3.5 Flash Model Documentation},
  author = {Google AI for Developers},
  year = {2026},
  howpublished = {\url{https://ai.google.dev/gemini-api/docs/models/gemini-3.5-flash}},
  note = {Accessed: 2026-05-25}
}

@misc{qwen2026qwen35,
  title = {Qwen3.5 Model Collection},
  author = {Qwen Team},
  year = {2026},
  howpublished = {\url{https://huggingface.co/Qwen/Qwen3.5-9B}},
  note = {Hugging Face Model Hub}
}

@misc{qwen2026github,
  title = {Qwen 3.6 Repository and Technical Documentation},
  author = {Qwen Team},
  year = {2026},
  howpublished = {\url{https://github.com/QwenLM/Qwen3.6}},
  note = {Software repository}
}

@misc{openai2026gpt55instantdoc,
  title = {GPT-5.5 Instant Model Documentation},
  author = {OpenAI},
  year = {2026},
  howpublished = {\url{https://platform.openai.com/docs/models/gpt-5.5-instant}},
  note = {OpenAI API Documentation. Accessed: 2026-05-15}
}

@article{bai2025qwen2,
  title={Qwen2. 5-vl technical report},
  author={Bai, Shuai and Chen, Keqin and Liu, Xuejing and Wang, Jialin and Ge, Wenbin and Song, Sibo and Dang, Kai and Wang, Peng and Wang, Shijie and Tang, Jun and others},
  journal={arXiv preprint arXiv:2502.13923},
  year={2025}
}

@article{zhu2025internvl3,
  title={Internvl3: Exploring advanced training and test-time recipes for open-source multimodal models},
  author={Zhu, Jinguo and Wang, Weiyun and Chen, Zhe and Liu, Zhaoyang and Ye, Shenglong and Gu, Lixin and Tian, Hao and Duan, Yuchen and Su, Weijie and Shao, Jie and others},
  journal={arXiv preprint arXiv:2504.10479},
  year={2025}
}

@misc{abdin2024phi3technicalreporthighly,
      title={Phi-3 Technical Report: A Highly Capable Language Model Locally on Your Phone}, 
      author={Marah Abdin and Jyoti Aneja and Hany Awadalla and Ahmed Awadallah and Ammar Ahmad Awan and Nguyen Bach and Amit Bahree and Arash Bakhtiari and Jianmin Bao and Harkirat Behl and Alon Benhaim and Misha Bilenko and Johan Bjorck and Sébastien Bubeck and Martin Cai and Qin Cai and Vishrav Chaudhary and Dong Chen and Dongdong Chen and Weizhu Chen and Yen-Chun Chen and Yi-Ling Chen and Hao Cheng and Parul Chopra and Xiyang Dai and Matthew Dixon and Ronen Eldan and Victor Fragoso and Jianfeng Gao and Mei Gao and Min Gao and Amit Garg and Allie Del Giorno and Abhishek Goswami and Suriya Gunasekar and Emman Haider and Junheng Hao and Russell J. Hewett and Wenxiang Hu and Jamie Huynh and Dan Iter and Sam Ade Jacobs and Mojan Javaheripi and Xin Jin and Nikos Karampatziakis and Piero Kauffmann and Mahoud Khademi and Dongwoo Kim and Young Jin Kim and Lev Kurilenko and James R. Lee and Yin Tat Lee and Yuanzhi Li and Yunsheng Li and Chen Liang and Lars Liden and Xihui Lin and Zeqi Lin and Ce Liu and Liyuan Liu and Mengchen Liu and Weishung Liu and Xiaodong Liu and Chong Luo and Piyush Madan and Ali Mahmoudzadeh and David Majercak and Matt Mazzola and Caio César Teodoro Mendes and Arindam Mitra and Hardik Modi and Anh Nguyen and Brandon Norick and Barun Patra and Daniel Perez-Becker and Thomas Portet and Reid Pryzant and Heyang Qin and Marko Radmilac and Liliang Ren and Gustavo de Rosa and Corby Rosset and Sambudha Roy and Olatunji Ruwase and Olli Saarikivi and Amin Saied and Adil Salim and Michael Santacroce and Shital Shah and Ning Shang and Hiteshi Sharma and Yelong Shen and Swadheen Shukla and Xia Song and Masahiro Tanaka and Andrea Tupini and Praneetha Vaddamanu and Chunyu Wang and Guanhua Wang and Lijuan Wang and Shuohang Wang and Xin Wang and Yu Wang and Rachel Ward and Wen Wen and Philipp Witte and Haiping Wu and Xiaoxia Wu and Michael Wyatt and Bin Xiao and Can Xu and Jiahang Xu and Weijian Xu and Jilong Xue and Sonali Yadav and Fan Yang and Jianwei Yang and Yifan Yang and Ziyi Yang and Donghan Yu and Lu Yuan and Chenruidong Zhang and Cyril Zhang and Jianwen Zhang and Li Lyna Zhang and Yi Zhang and Yue Zhang and Yunan Zhang and Xiren Zhou},
      year={2024},
      eprint={2404.14219},
      archivePrefix={arXiv},
      primaryClass={cs.CL},
      url={https://arxiv.org/abs/2404.14219}, 
}

@article{abdin2024phi,
  title={Phi-4 technical report},
  author={Abdin, Marah and Aneja, Jyoti and Behl, Harkirat and Bubeck, S{\'e}bastien and Eldan, Ronen and Gunasekar, Suriya and Harrison, Michael and Hewett, Russell J and Javaheripi, Mojan and Kauffmann, Piero and others},
  journal={arXiv preprint arXiv:2412.08905},
  year={2024}
}

@article{lu2025ovis2,
  title={Ovis2. 5 technical report},
  author={Lu, Shiyin and Li, Yang and Xia, Yu and Hu, Yuwei and Zhao, Shanshan and Ma, Yanqing and Wei, Zhichao and Li, Yinglun and Duan, Lunhao and Zhao, Jianshan and others},
  journal={arXiv preprint arXiv:2508.11737},
  year={2025}
}

@article{cheng2025glyph,
  title={Glyph: Scaling context windows via visual-text compression},
  author={Cheng, Jiale and Liu, Yusen and Zhang, Xinyu and Fei, Yulin and Hong, Wenyi and Lyu, Ruiliang and Wang, Weihan and Su, Zhe and Gu, Xiaotao and Liu, Xiao and others},
  journal={arXiv preprint arXiv:2510.17800},
  year={2025}
}

@article{wu2024deepseek,
  title={Deepseek-vl2: Mixture-of-experts vision-language models for advanced multimodal understanding},
  author={Wu, Zhiyu and Chen, Xiaokang and Pan, Zizheng and Liu, Xingchao and Liu, Wen and Dai, Damai and Gao, Huazuo and Ma, Yiyang and Wu, Chengyue and Wang, Bingxuan and others},
  journal={arXiv preprint arXiv:2412.10302},
  year={2024}
}

@misc{liu2023improved,
      title={Improved Baselines with Visual Instruction Tuning}, 
      author={Haotian Liu and Chunyuan Li and Yuheng Li and Yong Jae Lee},
      year={2023},
      eprint={2310.03744},
      archivePrefix={arXiv},
      primaryClass={cs.CV}
}

@article{achiam2023gpt,
  title={Gpt-4 technical report},
  author={Achiam, Josh and Adler, Steven and Agarwal, Sandhini and Ahmad, Lama and Akkaya, Ilge and Aleman, Florencia Leoni and Almeida, Diogo and Altenschmidt, Janko and Altman, Sam and Anadkat, Shyamal and others},
  journal={arXiv preprint arXiv:2303.08774},
  year={2023}
}

@article{hurst2024gpt,
  title={Gpt-4o system card},
  author={Hurst, Aaron and Lerer, Adam and Goucher, Adam P and Perelman, Adam and Ramesh, Aditya and Clark, Aidan and Ostrow, AJ and Welihinda, Akila and Hayes, Alan and Radford, Alec and others},
  journal={arXiv preprint arXiv:2410.21276},
  year={2024}
}

@misc{openai2025gpt41,
  title={GPT-4.1 Model Card},
  author={OpenAI},
  year={2025},
  month={April},
  howpublished={\url{https://platform.openai.com/docs/models/gpt-4.1}},
  note={Released on April 14, 2025}
}

@article{team2023gemini,
  title={Gemini: a family of highly capable multimodal models},
  author={Team, Gemini and Anil, Rohan and Borgeaud, Sebastian and Alayrac, Jean-Baptiste and Yu, Jiahui and Soricut, Radu and Schalkwyk, Johan and Dai, Andrew M and Hauth, Anja and Millican, Katie and others},
  journal={arXiv preprint arXiv:2312.11805},
  year={2023}
}

@misc{doubao2025seed16,
  title = {Seed1.6: Tech Introduction},
  author = {{ByteDance Seed Team}},
  year = {2025},
  month = {June},
  day = {25},
  howpublished = {\url{https://seed.bytedance.com/en/seed1_6}},
  note = {Model ID: doubao-seed-1-6-251015. Accessed: 2025-12-25}
}

@article{peng2023kosmos,
  title={Kosmos-2: Grounding multimodal large language models to the world},
  author={Peng, Zhiliang and Wang, Wenhui and Dong, Li and Hao, Yaru and Huang, Shaohan and Ma, Shuming and Wei, Furu},
  journal={arXiv preprint arXiv:2306.14824},
  year={2023}
}

@article{shang2025bridging,
  title={Bridging text and vision: A multi-view text-vision registration approach for cross-modal place recognition},
  author={Shang, Tianyi and Li, Zhenyu and Xu, Pengjie and Qiao, Jinwei and Chen, Gang and Ruan, Zihan and Hu, Weijun},
  journal={arXiv preprint arXiv:2502.14195},
  year={2025}
}

@article{guo2025enhanced,
  title={Enhanced Natural Language Annotation and Query for Semantic Mapping in Visual SLAM Using Large Language Models},
  author={Guo, Lingfeng and Li, Zihan and Min, Shengjie},
  journal={Journal of Sustainability, Policy, and Practice},
  volume={1},
  number={3},
  pages={131--143},
  year={2025}
}

@article{huang2025mllm,
  title={MLLM-For3D: Adapting Multimodal Large Language Model for 3D Reasoning Segmentation},
  author={Huang, Jiaxin and Chen, Runnan and Li, Ziwen and Gao, Zhengqing and He, Xiao and Guo, Yandong and Gong, Mingming and Liu, Tongliang},
  journal={arXiv preprint arXiv:2503.18135},
  year={2025}
}

@article{lu2025rsvp,
  title={RSVP: Reasoning Segmentation via Visual Prompting and Multi-modal Chain-of-Thought},
  author={Lu, Yi and Cao, Jiawang and Wu, Yongliang and Li, Bozheng and Tang, Licheng and Ji, Yangguang and Wu, Chong and Wu, Jay and Zhu, Wenbo},
  journal={arXiv preprint arXiv:2506.04277},
  year={2025}
}

@article{zhang2025openmaskdino3d,
  title={OpenMaskDINO3D: Reasoning 3D Segmentation via Large Language Model},
  author={Zhang, Kunshen},
  journal={arXiv preprint arXiv:2506.04837},
  year={2025}
}

@inproceedings{yue2025instruction,
  title={Instruction-augmented multimodal alignment for image-text and element matching},
  author={Yue, Xinli and Sun, JianHui and Lu, Junda and Yao, Liangchao and Xia, Fan and Wang, Tianyi and Rao, Fengyun and Lyu, Jing and Deng, Yuetang},
  booktitle={Proceedings of the Computer Vision and Pattern Recognition Conference},
  pages={1379--1388},
  year={2025}
}

@article{wang2024multi,
  title={Multi-level Symmetric Semantic Alignment Network for image--text matching},
  author={Wang, Wenzhuang and Di, Xiaoguang and Liu, Maozhen and Gao, Feng},
  journal={Neurocomputing},
  volume={599},
  pages={128082},
  year={2024},
  publisher={Elsevier}
}

@article{yarom2023you,
  title={What you see is what you read? improving text-image alignment evaluation},
  author={Yarom, Michal and Bitton, Yonatan and Changpinyo, Soravit and Aharoni, Roee and Herzig, Jonathan and Lang, Oran and Ofek, Eran and Szpektor, Idan},
  journal={Advances in Neural Information Processing Systems},
  volume={36},
  pages={1601--1619},
  year={2023}
}

@misc{openaio1,
  author    = {OpenAI},
  title     = {{OpenAI o1}},
  howpublished = {\url{https://openai.com/o1/}},
  year      = {2024}
}

@misc{openai2025o3,
  author       = {OpenAI},
  title        = {{OpenAI o3 and o4-mini System Card}},
  howpublished = {\url{https://cdn.openai.com/pdf/2221c875-02dc-4789-800b-e7758f3722c1/o3-and-o4-mini-system-card.pdf}},
  year         = {2025},
}

@misc{doubao,
  author    = {ByteDance},
  title     = {{doubao‑seed‑1.6‑thinking}},
  howpublished = {\url{https://www.volcengine.com/docs/82379/1593702?utm_source=chatgpt.com&lang=zh}},
  year      = {2025}
}

@article{guo2025deepseekr1,
  title={Deepseek-r1: Incentivizing reasoning capability in llms via reinforcement learning},
  author={Guo, Daya and Yang, Dejian and Zhang, Haowei and Song, Junxiao and Zhang, Ruoyu and Xu, Runxin and Zhu, Qihao and Ma, Shirong and Wang, Peiyi and Bi, Xiao and others},
  journal={arXiv preprint arXiv:2501.12948},
  year={2025}
}

@article{team2025kimi,
  title={Kimi-vl technical report},
  author={Team, Kimi and Du, Angang and Yin, Bohong and Xing, Bowei and Qu, Bowen and Wang, Bowen and Chen, Cheng and Zhang, Chenlin and Du, Chenzhuang and Wei, Chu and others},
  journal={arXiv preprint arXiv:2504.07491},
  year={2025}
}

@article{yuan2025gsm8k,
  title={GSM8K-V: Can Vision Language Models Solve Grade School Math Word Problems in Visual Contexts},
  author={Yuan, Fan and Yan, Yuchen and Jiang, Yifan and Zhao, Haoran and Feng, Tao and Chen, Jinyan and Lou, Yanwei and Zhang, Wenqi and Shen, Yongliang and Lu, Weiming and others},
  journal={arXiv preprint arXiv:2509.25160},
  year={2025}
}

@article{yang2024mathglm,
  title={Mathglm-vision: solving mathematical problems with multi-modal large language model},
  author={Yang, Zhen and Chen, Jinhao and Du, Zhengxiao and Yu, Wenmeng and Wang, Weihan and Hong, Wenyi and Jiang, Zhihuan and Xu, Bin and Tang, Jie},
  journal={arXiv preprint arXiv:2409.13729},
  year={2024}
}

@article{wang2024measuring,
  title={Measuring multimodal mathematical reasoning with math-vision dataset},
  author={Wang, Ke and Pan, Junting and Shi, Weikang and Lu, Zimu and Ren, Houxing and Zhou, Aojun and Zhan, Mingjie and Li, Hongsheng},
  journal={Advances in Neural Information Processing Systems},
  volume={37},
  pages={95095--95169},
  year={2024}
}

@article{wu2025spatialscore,
  title={SpatialScore: Towards Unified Evaluation for Multimodal Spatial Understanding},
  author={Wu, Haoning and Huang, Xiao and Chen, Yaohui and Zhang, Ya and Wang, Yanfeng and Xie, Weidi},
  journal={arXiv preprint arXiv:2505.17012},
  year={2025}
}

@inproceedings{yang2025thinking,
  title={Thinking in space: How multimodal large language models see, remember, and recall spaces},
  author={Yang, Jihan and Yang, Shusheng and Gupta, Anjali W and Han, Rilyn and Fei-Fei, Li and Xie, Saining},
  booktitle={Proceedings of the Computer Vision and Pattern Recognition Conference},
  pages={10632--10643},
  year={2025}
}

@article{cai2025holistic,
  title={Holistic Evaluation of Multimodal LLMs on Spatial Intelligence},
  author={Cai, Zhongang and Wang, Yubo and Sun, Qingping and Wang, Ruisi and Gu, Chenyang and Yin, Wanqi and Lin, Zhiqian and Yang, Zhitao and Wei, Chen and Qian, Oscar and others},
  journal={arXiv preprint arXiv:2508.13142},
  year={2025}
}

@article{xu2025visulogic,
  title={Visulogic: A benchmark for evaluating visual reasoning in multi-modal large language models},
  author={Xu, Weiye and Wang, Jiahao and Wang, Weiyun and Chen, Zhe and Zhou, Wengang and Yang, Aijun and Lu, Lewei and Li, Houqiang and Wang, Xiaohua and Zhu, Xizhou and others},
  journal={arXiv preprint arXiv:2504.15279},
  year={2025}
}

@article{xiao2024logicvista,
  title={Logicvista: Multimodal llm logical reasoning benchmark in visual contexts},
  author={Xiao, Yijia and Sun, Edward and Liu, Tianyu and Wang, Wei},
  journal={arXiv preprint arXiv:2407.04973},
  year={2024}
}

@article{chollet2025arc,
  title={Arc-agi-2: A new challenge for frontier ai reasoning systems},
  author={Chollet, Francois and Knoop, Mike and Kamradt, Gregory and Landers, Bryan and Pinkard, Henry},
  journal={arXiv preprint arXiv:2505.11831},
  year={2025}
}

@article{hong2025embodied,
  title={Embodied Web Agents: Bridging Physical-Digital Realms for Integrated Agent Intelligence},
  author={Hong, Yining and Sun, Rui and Li, Bingxuan and Yao, Xingcheng and Wu, Maxine and Chien, Alexander and Yin, Da and Wu, Ying Nian and Wang, Zhecan James and Chang, Kai-Wei},
  journal={arXiv preprint arXiv:2506.15677},
  year={2025}
}

@article{qi2025bear,
  title={Bear: Benchmarking and enhancing multimodal language models for atomic embodied capabilities},
  author={Qi, Yu and Zhao, Haibo and Guo, Ziyu and Ma, Siyuan and Chen, Ziyan and Han, Yaokun and Zhang, Renrui and Lin, Zitiantao and Xin, Shiji and Huang, Yijian and others},
  journal={arXiv preprint arXiv:2510.08759},
  year={2025}
}

@article{yang2025embodiedbench,
  title={Embodiedbench: Comprehensive benchmarking multi-modal large language models for vision-driven embodied agents},
  author={Yang, Rui and Chen, Hanyang and Zhang, Junyu and Zhao, Mark and Qian, Cheng and Wang, Kangrui and Wang, Qineng and Koripella, Teja Venkat and Movahedi, Marziyeh and Li, Manling and others},
  journal={arXiv preprint arXiv:2502.09560},
  year={2025}
}

@article{tan2025lumine,
  title={Lumine: An Open Recipe for Building Generalist Agents in 3D Open Worlds},
  author={Tan, Weihao and Li, Xiangyang and Fang, Yunhao and Yao, Heyuan and Yan, Shi and Luo, Hao and Ao, Tenglong and Li, Huihui and Ren, Hongbin and Yi, Bairen and others},
  journal={arXiv preprint arXiv:2511.08892},
  year={2025}
}

@article{wang2025game,
  title={Game-tars: Pretrained foundation models for scalable generalist multimodal game agents},
  author={Wang, Zihao and Li, Xujing and Ye, Yining and Fang, Junjie and Wang, Haoming and Liu, Longxiang and Liang, Shihao and Lu, Junting and Wu, Zhiyong and Feng, Jiazhan and others},
  journal={arXiv preprint arXiv:2510.23691},
  year={2025}
}

@article{tan2025stardojo,
  title={Stardojo: Benchmarking open-ended behaviors of agentic multimodal llms in production-living simulations with stardew valley},
  author={Tan, Weihao and Jiang, Changjiu and Duan, Yu and Lei, Mingcong and Li, Jiageng and Hong, Yitian and Wang, Xinrun and An, Bo},
  journal={arXiv preprint arXiv:2507.07445},
  year={2025}
}

@inproceedings{cui2024survey,
  title={A survey on multimodal large language models for autonomous driving},
  author={Cui, Can and Ma, Yunsheng and Cao, Xu and Ye, Wenqian and Zhou, Yang and Liang, Kaizhao and Chen, Jintai and Lu, Juanwu and Yang, Zichong and Liao, Kuei-Da and others},
  booktitle={Proceedings of the IEEE/CVF winter conference on applications of computer vision},
  pages={958--979},
  year={2024}
}

@article{hu2025vision,
  title={Vision-language-action models for autonomous driving: Past, present, and future},
  author={Hu, Tianshuai and Liu, Xiaolu and Wang, Song and Zhu, Yiyao and Liang, Ao and Kong, Lingdong and Zhao, Guoyang and Gong, Zeying and Cen, Jun and Huang, Zhiyu and others},
  journal={arXiv preprint arXiv:2512.16760},
  year={2025}
}

@article{lu2025uniugp,
  title={UniUGP: Unifying understanding, generation, and planing for end-to-end autonomous driving},
  author={Lu, Hao and Liu, Ziyang and Jiang, Guangfeng and Luo, Yuanfei and Chen, Sheng and Zhang, Yangang and Chen, Ying-Cong},
  journal={arXiv preprint arXiv:2512.09864},
  year={2025}
}

@article{lu2023mathvista,
  title={Mathvista: Evaluating mathematical reasoning of foundation models in visual contexts},
  author={Lu, Pan and Bansal, Hritik and Xia, Tony and Liu, Jiacheng and Li, Chunyuan and Hajishirzi, Hannaneh and Cheng, Hao and Chang, Kai-Wei and Galley, Michel and Gao, Jianfeng},
  journal={arXiv preprint arXiv:2310.02255},
  year={2023}
}

@inproceedings{qiao2025we,
  title={We-math: Does your large multimodal model achieve human-like mathematical reasoning?},
  author={Qiao, Runqi and Tan, Qiuna and Dong, Guanting and MinhuiWu, MinhuiWu and Sun, Chong and Song, Xiaoshuai and Wang, Jiapeng and Gongque, Zhuoma and Lei, Shanglin and Zhang, Yifan and others},
  booktitle={Proceedings of the 63rd Annual Meeting of the Association for Computational Linguistics (Volume 1: Long Papers)},
  pages={20023--20070},
  year={2025}
}

@article{li2024mmcode,
  title={Mmcode: Benchmarking multimodal large language models for code generation with visually rich programming problems},
  author={Li, Kaixin and Tian, Yuchen and Hu, Qisheng and Luo, Ziyang and Huang, Zhiyong and Ma, Jing},
  journal={arXiv preprint arXiv:2404.09486},
  year={2024}
}

@article{lu2021inter,
  title={Inter-gps: Interpretable geometry problem solving with formal language and symbolic reasoning},
  author={Lu, Pan and Gong, Ran and Jiang, Shibiao and Qiu, Liang and Huang, Siyuan and Liang, Xiaodan and Zhu, Song-Chun},
  journal={arXiv preprint arXiv:2105.04165},
  year={2021}
}

@inproceedings{li2025eee,
  title={Eee-bench: A comprehensive multimodal electrical and electronics engineering benchmark},
  author={Li, Ming and Zhong, Jike and Chen, Tianle and Lai, Yuxiang and Psounis, Konstantinos},
  booktitle={Proceedings of the Computer Vision and Pattern Recognition Conference},
  pages={13337--13349},
  year={2025}
}

@inproceedings{yue2024mmmu,
  title={Mmmu: A massive multi-discipline multimodal understanding and reasoning benchmark for expert agi},
  author={Yue, Xiang and Ni, Yuansheng and Zhang, Kai and Zheng, Tianyu and Liu, Ruoqi and Zhang, Ge and Stevens, Samuel and Jiang, Dongfu and Ren, Weiming and Sun, Yuxuan and others},
  booktitle={Proceedings of the IEEE/CVF Conference on Computer Vision and Pattern Recognition},
  pages={9556--9567},
  year={2024}
}

@inproceedings{zhang2024mathverse,
  title={Mathverse: Does your multi-modal llm truly see the diagrams in visual math problems?},
  author={Zhang, Renrui and Jiang, Dongzhi and Zhang, Yichi and Lin, Haokun and Guo, Ziyu and Qiu, Pengshuo and Zhou, Aojun and Lu, Pan and Chang, Kai-Wei and Qiao, Yu and others},
  booktitle={European Conference on Computer Vision},
  pages={169--186},
  year={2024},
  organization={Springer}
}

@inproceedings{wu2024v,
  title={V*: Guided visual search as a core mechanism in multimodal llms},
  author={Wu, Penghao and Xie, Saining},
  booktitle={Proceedings of the IEEE/CVF Conference on Computer Vision and Pattern Recognition},
  pages={13084--13094},
  year={2024}
}

@article{song2025visualpuzzles,
  title={VisualPuzzles: Decoupling Multimodal Reasoning Evaluation from Domain Knowledge},
  author={Song, Yueqi and Ou, Tianyue and Kong, Yibo and Li, Zecheng and Neubig, Graham and Yue, Xiang},
  journal={arXiv preprint arXiv:2504.10342},
  year={2025}
}

@article{ren2025vgrp,
  title={Vgrp-bench: Visual grid reasoning puzzle benchmark for large vision-language models},
  author={Ren, Yufan and Tertikas, Konstantinos and Maiti, Shalini and Han, Junlin and Zhang, Tong and S{\"u}sstrunk, Sabine and Kokkinos, Filippos},
  journal={arXiv preprint arXiv:2503.23064},
  year={2025}
}

@article{guo2025r,
  title={R-bench: Graduate-level multi-disciplinary benchmarks for llm \& mllm complex reasoning evaluation},
  author={Guo, Meng-Hao and Xu, Jiajun and Zhang, Yi and Song, Jiaxi and Peng, Haoyang and Deng, Yi-Xuan and Dong, Xinzhi and Nakayama, Kiyohiro and Geng, Zhengyang and Wang, Chen and others},
  journal={arXiv preprint arXiv:2505.02018},
  year={2025}
}

@inproceedings{feng2025citybench,
  title={Citybench: Evaluating the capabilities of large language models for urban tasks},
  author={Feng, Jie and Zhang, Jun and Liu, Tianhui and Zhang, Xin and Ouyang, Tianjian and Yan, Junbo and Du, Yuwei and Guo, Siqi and Li, Yong},
  booktitle={Proceedings of the 31st ACM SIGKDD Conference on Knowledge Discovery and Data Mining V. 2},
  pages={5413--5424},
  year={2025}
}

@article{xie2025vlms,
  title={Are vlms ready for autonomous driving? an empirical study from the reliability, data, and metric perspectives},
  author={Xie, Shaoyuan and Kong, Lingdong and Dong, Yuhao and Sima, Chonghao and Zhang, Wenwei and Chen, Qi Alfred and Liu, Ziwei and Pan, Liang},
  journal={arXiv preprint arXiv:2501.04003},
  year={2025}
}

@article{zheng2024planagent,
  title={Planagent: A multi-modal large language agent for closed-loop vehicle motion planning},
  author={Zheng, Yupeng and Xing, Zebin and Zhang, Qichao and Jin, Bu and Li, Pengfei and Zheng, Yuhang and Xia, Zhongpu and Zhan, Kun and Lang, Xianpeng and Chen, Yaran and others},
  journal={arXiv preprint arXiv:2406.01587},
  year={2024}
}

@article{zeng2024perceive,
  title={Perceive, reflect, and plan: Designing llm agent for goal-directed city navigation without instructions},
  author={Zeng, Qingbin and Yang, Qinglong and Dong, Shunan and Du, Heming and Zheng, Liang and Xu, Fengli and Li, Yong},
  journal={arXiv preprint arXiv:2408.04168},
  year={2024}
}

@article{xu2025geonav,
  title={Geonav: Empowering mllms with explicit geospatial reasoning abilities for language-goal aerial navigation},
  author={Xu, Haotian and Hu, Yue and Gao, Chen and Zhu, Zhengqiu and Zhao, Yong and Li, Yong and Yin, Quanjun},
  journal={arXiv preprint arXiv:2504.09587},
  year={2025}
}

@article{qiao2025navbench,
  title={NavBench: Probing Multimodal Large Language Models for Embodied Navigation},
  author={Qiao, Yanyuan and Hong, Haodong and Lyu, Wenqi and An, Dong and Zhang, Siqi and Xie, Yutong and Wang, Xinyu and Wu, Qi},
  journal={arXiv preprint arXiv:2506.01031},
  year={2025}
}

@inproceedings{krechetova2025geobenchx,
  title={GeoBenchX: Benchmarking LLMs in Agent Solving Multistep Geospatial Tasks},
  author={Krechetova, Varvara and Kochedykov, Denis},
  booktitle={Proceedings of the 1st ACM SIGSPATIAL International Workshop on Generative and Agentic AI for Multi-Modality Space-Time Intelligence},
  pages={27--35},
  year={2025}
}

@article{xing2025can,
  title={Can Large Vision Language Models Read Maps Like a Human?},
  author={Xing, Shuo and Sun, Zezhou and Xie, Shuangyu and Chen, Kaiyuan and Huang, Yanjia and Wang, Yuping and Li, Jiachen and Song, Dezhen and Tu, Zhengzhong},
  journal={arXiv preprint arXiv:2503.14607},
  year={2025}
}

@article{feng2025can,
  title={Can MLLMs Guide Me Home? A Benchmark Study on Fine-Grained Visual Reasoning from Transit Maps},
  author={Feng, Sicheng and Wang, Song and Ouyang, Shuyi and Kong, Lingdong and Song, Zikai and Zhu, Jianke and Wang, Huan and Wang, Xinchao},
  journal={arXiv preprint arXiv:2505.18675},
  year={2025}
}

@article{feng2025rewardmap,
  title={RewardMap: Tackling sparse rewards in fine-grained visual reasoning via multi-stage reinforcement learning},
  author={Feng, Sicheng and Tuo, Kaiwen and Wang, Song and Kong, Lingdong and Zhu, Jianke and Wang, Huan},
  journal={arXiv preprint arXiv:2510.02240},
  year={2025}
}

@article{zhang2025thyme,
  title={Thyme: Think beyond images},
  author={Zhang, Yi-Fan and Lu, Xingyu and Yin, Shukang and Fu, Chaoyou and Chen, Wei and Hu, Xiao and Wen, Bin and Jiang, Kaiyu and Liu, Changyi and Zhang, Tianke and others},
  journal={arXiv preprint arXiv:2508.11630},
  year={2025}
}

@article{zheng2025deepeyes,
  title={DeepEyes: Incentivizing" Thinking with Images" via Reinforcement Learning},
  author={Zheng, Ziwei and Yang, Michael and Hong, Jack and Zhao, Chenxiao and Xu, Guohai and Yang, Le and Shen, Chao and Yu, Xing},
  journal={arXiv preprint arXiv:2505.14362},
  year={2025}
}

@article{feng2025efficient,
  title={Efficient reasoning models: A survey},
  author={Feng, Sicheng and Fang, Gongfan and Ma, Xinyin and Wang, Xinchao},
  journal={arXiv preprint arXiv:2504.10903},
  year={2025}
}

@article{shao2024deepseekmath,
  title={Deepseekmath: Pushing the limits of mathematical reasoning in open language models},
  author={Shao, Zhihong and Wang, Peiyi and Zhu, Qihao and Xu, Runxin and Song, Junxiao and Bi, Xiao and Zhang, Haowei and Zhang, Mingchuan and Li, YK and Wu, Yang and others},
  journal={arXiv preprint arXiv:2402.03300},
  year={2024}
}

@article{tan2025reason,
  title={Reason-rft: Reinforcement fine-tuning for visual reasoning},
  author={Tan, Huajie and Ji, Yuheng and Hao, Xiaoshuai and Lin, Minglan and Wang, Pengwei and Wang, Zhongyuan and Zhang, Shanghang},
  journal={arXiv preprint arXiv:2503.20752},
  year={2025}
}

@article{wong2025survey,
  title={A Survey of Robotic Navigation and Manipulation with Physics Simulators in the Era of Embodied AI},
  author={Wong, Lik Hang Kenny and Kang, Xueyang and Bai, Kaixin and Zhang, Jianwei},
  journal={arXiv preprint arXiv:2505.01458},
  year={2025}
}

@inproceedings{shoaib2023survey,
  title={A survey on the applications of frontier ai, foundation models, and large language models to intelligent transportation systems},
  author={Shoaib, Mohamed R and Emara, Heba M and Zhao, Jun},
  booktitle={2023 International Conference on Computer and Applications (ICCA)},
  pages={1--7},
  year={2023},
  organization={IEEE}
}

@inproceedings{fu2024drive,
  title={Drive like a human: Rethinking autonomous driving with large language models},
  author={Fu, Daocheng and Li, Xin and Wen, Licheng and Dou, Min and Cai, Pinlong and Shi, Botian and Qiao, Yu},
  booktitle={2024 IEEE/CVF Winter Conference on Applications of Computer Vision Workshops (WACVW)},
  pages={910--919},
  year={2024},
  organization={IEEE}
}

@article{wen2023dilu,
  title={Dilu: A knowledge-driven approach to autonomous driving with large language models},
  author={Wen, Licheng and Fu, Daocheng and Li, Xin and Cai, Xinyu and Ma, Tao and Cai, Pinlong and Dou, Min and Shi, Botian and He, Liang and Qiao, Yu},
  journal={arXiv preprint arXiv:2309.16292},
  year={2023}
}

@inproceedings{sima2024drivelm,
  title={Drivelm: Driving with graph visual question answering},
  author={Sima, Chonghao and Renz, Katrin and Chitta, Kashyap and Chen, Li and Zhang, Hanxue and Xie, Chengen and Bei{\ss}wenger, Jens and Luo, Ping and Geiger, Andreas and Li, Hongyang},
  booktitle={European conference on computer vision},
  pages={256--274},
  year={2024},
  organization={Springer}
}

@article{dihan2024mapeval,
  title={Mapeval: A map-based evaluation of geo-spatial reasoning in foundation models},
  author={Dihan, Mahir Labib and Hassan, Md Tanvir and Parvez, Md Tanvir and Hasan, Md Hasebul and Alam, Md Almash and Cheema, Muhammad Aamir and Ali, Mohammed Eunus and Parvez, Md Rizwan},
  journal={arXiv preprint arXiv:2501.00316},
  year={2024}
}

@article{fang2024travellm,
  title={Travellm: Could you plan my new public transit route in face of a network disruption?},
  author={Fang, Bowen and Yang, Zixiao and Di, Xuan},
  journal={arXiv preprint arXiv:2407.14926},
  year={2024}
}

@article{feng2024citybench,
  title={Citybench: Evaluating the capabilities of large language model as world model},
  author={Feng, Jie and Zhang, Jun and Yan, Junbo and Zhang, Xin and Ouyang, Tianjian and Liu, Tianhui and Du, Yuwei and Guo, Siqi and Li, Yong},
  journal={arXiv e-prints},
  pages={arXiv--2406},
  year={2024}
}

@inproceedings{cao2024maplm,
  title={Maplm: A real-world large-scale vision-language benchmark for map and traffic scene understanding},
  author={Cao, Xu and Zhou, Tong and Ma, Yunsheng and Ye, Wenqian and Cui, Can and Tang, Kun and Cao, Zhipeng and Liang, Kaizhao and Wang, Ziran and Rehg, James M and others},
  booktitle={Proceedings of the IEEE/CVF conference on computer vision and pattern recognition},
  pages={21819--21830},
  year={2024}
}

@article{li2025mapqa,
  title={MapQA: Open-domain Geospatial Question Answering on Map Data},
  author={Li, Zekun and Grossman, Malcolm and Kulkarni, Mihir and Chen, Muhao and Chiang, Yao-Yi and others},
  journal={arXiv preprint arXiv:2503.07871},
  year={2025}
}

@article{zheng2025geox,
  title={GeoX-Bench: Benchmarking Cross-View Geo-Localization and Pose Estimation Capabilities of Large Multimodal Models},
  author={Zheng, Yushuo and Ying, Jiangyong and Duan, Huiyu and Li, Chunyi and Zhang, Zicheng and Liu, Jing and Liu, Xiaohong and Zhai, Guangtao},
  journal={arXiv preprint arXiv:2511.13259},
  year={2025}
}

@article{srivastava2025mapiq,
  title={MapIQ: Evaluating multimodal large language models for map question answering},
  author={Srivastava, Varun and Lei, Fan and Mukhopadhyay, Srija and Gupta, Vivek and Maciejewski, Ross},
  journal={arXiv preprint arXiv:2507.11625},
  year={2025}
}

@inproceedings{ung2025cartomapqa,
  title={CartoMapQA: A Fundamental Benchmark Dataset Evaluating Vision-Language Models on Cartographic Map Understanding},
  author={Ung, Huy Quang and Habault, Guillaume and Nishimura, Yasutaka and Niu, Hao and Legaspi, Roberto and Oya, Tomoki and Kojima, Ryoichi and Taya, Masato and Ono, Chihiro and Minamikawa, Atsunori and others},
  booktitle={Proceedings of the 33rd ACM International Conference on Advances in Geographic Information Systems},
  pages={440--453},
  year={2025}
}

@article{pyo2025frieda,
  title={FRIEDA: Benchmarking Multi-Step Cartographic Reasoning in Vision-Language Models},
  author={Pyo, Jiyoon and Jiao, Yuankun and Jung, Dongwon and Li, Zekun and Jang, Leeje and Kirsanova, Sofia and Kim, Jina and Lin, Yijun and Liu, Qin and Xie, Junyi and others},
  journal={arXiv preprint arXiv:2512.08016},
  year={2025}
}

@article{venu2025comprehensive,
  title={A comprehensive review of path planning algorithms for autonomous navigation},
  author={Venu, Sangeeth and Gurusamy, Muralimohan},
  journal={Results in Engineering},
  pages={107750},
  year={2025},
  publisher={Elsevier}
}

@article{yan2024survey,
  title={A survey of sustainable development of intelligent transportation system based on urban travel demand},
  author={Yan, Hongyu and Lv, Zhiqiang},
  journal={Development},
  volume={2},
  number={1},
  pages={2399},
  year={2024}
}

@article{chen2025path,
  title={Path planning algorithm for logistics autonomous vehicles at Cainiao stations based on multi-sensor data fusion},
  author={Chen, Yan},
  journal={PLoS One},
  volume={20},
  number={5},
  pages={e0321257},
  year={2025},
  publisher={Public Library of Science San Francisco, CA USA}
}

@inproceedings{xu2025flame,
  title={Flame: Learning to navigate with multimodal llm in urban environments},
  author={Xu, Yunzhe and Pan, Yiyuan and Liu, Zhe and Wang, Hesheng},
  booktitle={Proceedings of the AAAI Conference on Artificial Intelligence},
  volume={39},
  pages={9005--9013},
  year={2025}
}

@inproceedings{ranga2025urbandrivepathway,
  title={UrbanDrivePathway: A Decision-Making Framework for Navigating Urban Autonomous Vehicles in Complex Traffic Systems},
  author={Ranga, Jarabala and PRASATH, A ARUL and Kumar, Neeraj and Naveenkumar, R and Vadar, Parashuram S and Fiaz, AS Syed},
  booktitle={2025 8th International Conference on Trends in Electronics and Informatics (ICOEI)},
  pages={1575--1582},
  year={2025},
  organization={IEEE}
}

@article{shridhar2020alfworld,
  title={Alfworld: Aligning text and embodied environments for interactive learning},
  author={Shridhar, Mohit and Yuan, Xingdi and C{\^o}t{\'e}, Marc-Alexandre and Bisk, Yonatan and Trischler, Adam and Hausknecht, Matthew},
  journal={arXiv preprint arXiv:2010.03768},
  year={2020}
}

@article{cao2023representation,
  title={Representation granularity enables time-efficient autonomous exploration in large, complex worlds},
  author={Cao, Chao and Zhu, Hongbiao and Ren, Zhongqiang and Choset, Howie and Zhang, Ji},
  journal={Science Robotics},
  volume={8},
  number={80},
  pages={eadf0970},
  year={2023},
  publisher={American Association for the Advancement of Science}
}

@article{zhou2024cartomark,
  title={CartoMark: a benchmark dataset for map pattern recognition and map content retrieval with machine intelligence},
  author={Zhou, Xiran and Wen, Yi and Shao, Zhenfeng and Li, Wenwen and Li, Kaiyuan and Li, Honghao and Xie, Xiao and Yan, Zhigang},
  journal={Scientific Data},
  volume={11},
  number={1},
  pages={1205},
  year={2024},
  publisher={Nature Publishing Group UK London}
}

@article{yao2025understanding,
  title={Understanding the repeat curse in large language models from a feature perspective},
  author={Yao, Junchi and Yang, Shu and Xu, Jianhua and Hu, Lijie and Li, Mengdi and Wang, Di},
  journal={arXiv preprint arXiv:2504.14218},
  year={2025}
}

@article{gou2025reasoning,
  title={Reasoning-Aligned Perception Decoupling for Scalable Multi-modal Reasoning},
  author={Gou, Yunhao and Chen, Kai and Liu, Zhili and Hong, Lanqing and Jin, Xin and Li, Zhenguo and Kwok, James T and Zhang, Yu},
  journal={arXiv preprint arXiv:2506.04559},
  year={2025}
}

@article{avogaro2026sparc,
  title={SPARC: Separating Perception And Reasoning Circuits for Test-time Scaling of VLMs},
  author={Avogaro, Niccolo and Debnath, Nayanika and Mi, Li and Frick, Thomas and Wang, Junling and He, Zexue and Hua, Hang and Schindler, Konrad and Rigotti, Mattia},
  journal={arXiv preprint arXiv:2602.06566},
  year={2026}
}

@article{wang2025perception,
  title={Perception-aware policy optimization for multimodal reasoning},
  author={Wang, Zhenhailong and Guo, Xuehang and Stoica, Sofia and Xu, Haiyang and Wang, Hongru and Ha, Hyeonjeong and Chen, Xiusi and Chen, Yangyi and Yan, Ming and Huang, Fei and Ji, Heng},
  journal={arXiv preprint arXiv:2507.06448},
  year={2025}
}

@article{chng2025sensenova,
  title={SenseNova-MARS: Empowering Multimodal Agentic Reasoning and Search via Reinforcement Learning},
  author={Chng, Yong Xien and Hu, Tao and Tong, Wenwen and Li, Xueheng and Chen, Jiandong and Yu, Haojia and Lu, Jiefan and Guo, Hewei and Deng, Hanming and Xie, Chengjun and Huang, Gao and Lin, Dahua and Lu, Lewei},
  journal={arXiv preprint arXiv:2512.24330},
  year={2025}
}

@article{ning2026mc,
  title={MC-Search: Evaluating and Enhancing Multimodal Agentic Search with Structured Long Reasoning Chains},
  author={Ning, Xuying and Fu, Dongqi and Wei, Tianxin and Ai, Mengting and Zou, Jiaru and Li, Ting-Wei and Tong, Hanghang and Zhu, Yada and Hamann, Hendrik and He, Jingrui},
  journal={arXiv preprint arXiv:2603.00873},
  year={2026}
}

@article{song2025codedance,
  title={CodeDance: A Dynamic Tool-integrated MLLM for Executable Visual Reasoning},
  author={Song, Qi and Li, Honglin and Yu, Yingchen and Zhou, Haoyi and Yang, Lin and Bai, Song and She, Qi and Huang, Zilong and Zhao, Yunqing},
  journal={arXiv preprint arXiv:2512.17312},
  year={2025}
}

@article{liu2025acereason,
  title={Acereason-nemotron 1.1: Advancing math and code reasoning through sft and rl synergy},
  author={Liu, Zihan and Yang, Zhuolin and Chen, Yang and Lee, Chankyu and Shoeybi, Mohammad and Catanzaro, Bryan and Ping, Wei},
  journal={arXiv preprint arXiv:2506.13284},
  year={2025}
}

@article{chen2025beyond,
  title={Beyond two-stage training: Cooperative sft and rl for llm reasoning},
  author={Chen, Liang and Han, Xueting and Shen, Li and Bai, Jing and Wong, Kam-Fai},
  journal={arXiv preprint arXiv:2509.06948},
  year={2025}
}

\newpage
\appendix
\setcounter{secnumdepth}{2}

\onecolumn
\clearpage


\section*{\centering \LARGE Table of Contents in Appendix}
\addcontentsline{toc}{section}{Table of Contents in Appendix}

\vspace{0.45cm}

\begingroup

\setlength{\parindent}{0pt}
\setlength{\parskip}{0.10em}
\fontsize{10.5pt}{14pt}\selectfont


\noindent\hyperref[sec:more_related_work]{\textbf{A \quad More Details for Related Works}}\dotfill\pageref{sec:more_related_work}\par
\noindent\hspace*{1.5em}\hyperref[subsec:rel_1]{A.1. \quad Reasoning Abilities in LLMs and MLLMs}\dotfill\pageref{subsec:rel_1}\par
\noindent\hspace*{1.5em}\hyperref[subsec:rel_2]{A.2. \quad Multimodal Reasoning Benchmarks}\dotfill\pageref{subsec:rel_2}\par
\noindent\hspace*{1.5em}\hyperref[subsec:rel_3]{A.3. \quad Map-Based Spatial Reasoning and Planning}\dotfill\pageref{subsec:rel_3}\par

\vspace{0.55em}


\noindent\hyperref[sec:data_construct]{\textbf{B \quad More Details for Dataset Construction}}\dotfill\pageref{sec:data_construct}\par
\noindent\hspace*{1.5em}\hyperref[sec:struct_data]{B.1. \quad Multi-Criteria Tabular Construction}\dotfill\pageref{sec:struct_data}\par
\noindent\hspace*{1.5em}\hyperref[subsec:RP_query_construct]{B.2. \quad RP Queries Construction}\dotfill\pageref{subsec:RP_query_construct}\par
\noindent\hspace*{1.5em}\hyperref[subsec:QA_query_construct]{B.3. \quad QA Queries Construction}\dotfill\pageref{subsec:QA_query_construct}\par
\noindent\hspace*{1.5em}\hyperref[subsec:PR_Query_distribution]{B.4. \quad RP Query Distribution Based on Difficulty Classification Metrics}\dotfill\pageref{subsec:PR_Query_distribution}\par
\noindent\hspace*{1.5em}\hyperref[subsec:Metromap_validation_alg]{B.5. \quad Quality Control: Metromap Validation Algorithm}\dotfill\pageref{subsec:Metromap_validation_alg}\par
\noindent\hspace*{1.5em}\hyperref[subsec:travelmap_val_alg]{B.6. \quad Quality Control: Travelmap Validation Algorithm}\dotfill\pageref{subsec:travelmap_val_alg}\par

\vspace{0.55em}


\noindent\hyperref[sec:metrics]{\textbf{C \quad Metrics}}\dotfill\pageref{sec:metrics}\par
\noindent\hspace*{1.5em}\hyperref[subsec:performance_metrics]{C.1. \quad Performance Metrics}\dotfill\pageref{subsec:performance_metrics}\par
\noindent\hspace*{1.5em}\hyperref[subsec:difficulty_metrics]{C.2. \quad Difficulty Classification Metrics}\dotfill\pageref{subsec:difficulty_metrics}\par

\vspace{0.55em}


\noindent\hyperref[sec:query_template]{\textbf{D \quad Query Template}}\dotfill\pageref{sec:query_template}\par
\noindent\hspace*{1.5em}\hyperref[subsec:PR_query]{D.1. \quad RP Query}\dotfill\pageref{subsec:PR_query}\par
\noindent\hspace*{1.5em}\hyperref[subsec:qa_query]{D.2. \quad QA Query}\dotfill\pageref{subsec:qa_query}\par

\vspace{0.55em}


\noindent\hyperref[sec:prompt_template]{\textbf{E \quad Prompt Template}}\dotfill\pageref{sec:prompt_template}\par
\noindent\hspace*{1.5em}\hyperref[subsec:prompt_template_1]{E.1. \quad Prompts for Generating Edge\_tab and Vertex\_tab}\dotfill\pageref{subsec:prompt_template_1}\par
\noindent\hspace*{1.5em}\hyperref[subsec:prompt_for_PR]{E.2. \quad Prompts for RP Queries}\dotfill\pageref{subsec:prompt_for_PR}\par
\noindent\hspace*{1.5em}\hyperref[subsec:prompt_for_qa]{E.3. \quad Prompts for QA Queries}\dotfill\pageref{subsec:prompt_for_qa}\par

\vspace{0.55em}


\noindent\hyperref[sec:add_exp_analysis]{\textbf{F \quad Additional Experimental Analysis}}\dotfill\pageref{sec:add_exp_analysis}\par
\noindent\hspace*{1.5em}\hyperref[subsec:experiment_settings]{F.1. \quad Experimental Settings}\dotfill\pageref{subsec:experiment_settings}\par
\noindent\hspace*{1.5em}\hyperref[subsec:additional_rp_results]{F.2. \quad Additional Route Planning Results}\dotfill\pageref{subsec:additional_rp_results}\par
\noindent\hspace*{1.5em}\hyperref[subsec:supp_qa_analysis]{F.3. \quad Supplementary QA Analysis}\dotfill\pageref{subsec:supp_qa_analysis}\par

\vspace{0.55em}


\noindent\hyperref[sec:more_experiments]{\textbf{G \quad More Experiments}}\dotfill\pageref{sec:more_experiments}\par
\noindent\hspace*{1.5em}\hyperref[subsec:abl_res_rp_and_qa]{G.1. \quad Ablation Study of RP and QA for Resolution}\dotfill\pageref{subsec:abl_res_rp_and_qa}\par
\noindent\hspace*{1.5em}\hyperref[subsec:abl_map_and_query_dfficulty]{G.2. \quad Ablation Study of Map Difficulty and Query Difficulty}\dotfill\pageref{subsec:abl_map_and_query_dfficulty}\par
\noindent\hspace*{1.5em}\hyperref[subsec:abl_tabular]{G.3. \quad Ablation Study of Tabular Modality}\dotfill\pageref{subsec:abl_tabular}\par
\noindent\hspace*{1.5em}\hyperref[subsec:abl_language]{G.4. \quad Ablation Study of Language}\dotfill\pageref{subsec:abl_language}\par

\vspace{0.55em}


\noindent\hyperref[sec:err_analysis]{\textbf{H \quad Error Case Analysis}}\dotfill\pageref{sec:err_analysis}\par

\vspace{0.55em}


\noindent\hyperref[sec:limitations]{\textbf{I \quad Limitations}}\dotfill\pageref{sec:limitations}\par

\vspace{0.55em}


\noindent\hyperref[sec:future_work]{\textbf{J \quad Future Work}}\dotfill\pageref{sec:future_work}\par

\endgroup

\vspace{0.8cm}
\clearpage


\section{More Details for Related Works}\label{sec:more_related_work}
\subsection{Reasoning Abilities in LLMs and MLLMs}\label{subsec:rel_1}
The rapid rise of multimodal large language models (MLLMs)~\cite{bai2025qwen3vltechnicalreport,bai2025qwen2,zhu2025internvl3,abdin2024phi,hurst2024gpt,team2023gemini} has fundamentally reshaped vision-language interaction, which have demonstrated strong capabilities in visual grounding~\cite{peng2023kosmos,shang2025bridging,guo2025enhanced}, reasoning-based segmentation~\cite{huang2025mllm,lu2025rsvp,zhang2025openmaskdino3d}, and text-image alignment~\cite{yue2025instruction,wang2024multi,yarom2023you}. 

Recently, the integration of reinforcement learning (RL) fine-tuning paradigms~\cite{zhang2025thyme,zheng2025deepeyes,feng2025efficient} has significantly expanded the reasoning capabilities of large language models (LLMs)~\cite{guo2025deepseekr1}. Methods such as Group Relative Policy Optimization (GRPO)~\cite{shao2024deepseekmath} have been particularly effective in activating and enhancing the latent logical deduction abilities within these models. These advancements have enabled LLMs to achieve improved performance in complex reasoning tasks, including mathematical reasoning~\cite{yuan2025gsm8k,yang2024mathglm,wang2024measuring}, fine-grained spatial understanding~\cite{wu2025spatialscore,yang2025thinking,cai2025holistic}, and logical reasoning~\cite{xu2025visulogic,xiao2024logicvista,chollet2025arc}.

Furthermore, the introduction of visual chain-of-thought and multi-dimensional perception has significantly enhanced multimodal reasoning in LLMs~\cite{openaio1,openai2025o3,doubao,team2025kimi} and multimodal large language models (MLLMs), allowing them to integrate and reason across both visual and textual modalities. These models' improved reasoning capabilities have broad implications for real-world applications, including embodied intelligence~\cite{hong2025embodied, qi2025bear, yang2025embodiedbench,shridhar2020alfworld}, game agents~\cite{tan2025lumine, wang2025game, tan2025stardojo}, and autonomous driving~\cite{cui2024survey, hu2025vision, lu2025uniugp}, where sophisticated reasoning is required to make complex decisions in dynamic, multimodal environments.

Against this backdrop, a series of reasoning-oriented multimodal large language models (MLLMs) that inherit advantages from earlier architectural designs have emerged~\cite{bai2025qwen3vltechnicalreport,bai2025qwen2,zhu2025internvl3,abdin2024phi,achiam2023gpt,team2023gemini}. Among them, models such as OpenAI 4o~\cite{hurst2024gpt}, Gemini~\cite{team2023gemini}, and Qwen~\cite{bai2025qwen3vltechnicalreport} have become key reference points for measuring the current level of cross-modal intelligence.

\subsection{Multimodal Reasoning Benchmarks}\label{subsec:rel_2}
Along with the rapid evolution of model capabilities, constructing multidimensional and interpretable evaluation frameworks to systematically characterize the performance of MLLMs across different reasoning levels has become a central challenge in the field~\cite{lu2023mathvista,qiao2025we,li2024mmcode,lu2021inter,li2025eee}. Early benchmarks, including V*Bench~\cite{wu2024v}, VisualPuzzles~\cite{song2025visualpuzzles}, VisuLogic~\cite{xu2025visulogic}, R-Bench~\cite{guo2025r}, and VGRP-Bench~\cite{ren2025vgrp}, primarily relied on synthetic tasks to assess abstract reasoning abilities related to logical structure construction and pattern recognition. Subsequently, benchmarks such as MathVQA~\cite{wang2024measuring}, MMMU~\cite{yue2024mmmu}, and MathVerse~\cite{zhang2024mathverse} further introduced cross-modal mathematical reasoning to examine model performance under scenarios requiring deep integration of symbolic and visual information. Moreover, CityBench~\cite{feng2025citybench} and DriveBench~\cite{xie2025vlms} \textit{et al.} further extended evaluation to spatial reasoning in urban scenarios. 

As an important carrier of high-density geospatial information, map-based visual question answering has gradually evolved into a key research direction for evaluating the depth of multimodal cognition. In this context, MapEval~\cite{dihan2024mapeval}, MapQA~\cite{li2025mapqa}, MapIQ~\cite{srivastava2025mapiq}, and CartoMark~\cite{zhou2024cartomark} have systematically revealed, from perspectives ranging from holistic evaluation and basic perception to textual element parsing, the substantial gap between multimodal models and human performance in geospatial cognition, as well as shared limitations in perception-reasoning decoupling and fine-grained semantic understanding.  

Furthermore, at the level of higher-order spatial reasoning, FRIEDA~\cite{pyo2025frieda} and GeoX-Bench~\cite{zheng2025geox} expose structural deficiencies in complex spatial orientation and compositional reasoning, while CartoMapQA~\cite{ung2025cartomapqa}, through hierarchical evaluation in real-world map service scenarios, characterizes reasoning bottlenecks in practical tasks such as scale understanding and route navigation.

However, although existing benchmarks are able to uncover deficiencies in local perception and specific spatial tasks, their overall coverage of model capabilities remains limited. Most evaluations focus on visual question answering and local perceptual skills, while lacking systematic characterization of global spatial modeling, long-range dependency reasoning, and sequential decision-making. 

\subsection{Map-Based Spatial Reasoning and Planning}\label{subsec:rel_3}

In multimodal reasoning research, map-based spatial reasoning and planning have become core problems in navigation~\cite{wong2025survey,venu2025comprehensive,xu2025flame,ranga2025urbandrivepathway}, intelligent transportation~\cite{shoaib2023survey,yan2024survey,chen2025path}, and autonomous driving~\cite{fu2024drive,wen2023dilu,sima2024drivelm}. Compared with traditional perception-driven map understanding tasks, these studies place greater emphasis on a model's ability to jointly model spatial structures, topological relationships, and long-tracing decision-making.

Recent works such as CityBench~\cite{feng2024citybench} and MapLM~\cite{cao2024maplm} explore map-driven spatial reasoning capabilities from different application scenarios, particularly examining model performance in transportation environments and urban contexts. Meanwhile, for embodied navigation tasks, researchers have proposed evaluation frameworks such as PlanAgent~\cite{zheng2024planagent}, PReP~\cite{zeng2024perceive}, GeoNav~\cite{xu2025geonav}, and NavBench~\cite{qiao2025navbench}. These frameworks assess the spatial reasoning capabilities of multimodal models from two complementary aspects: navigation understanding and step-by-step execution, thereby promoting a transition from static comprehension to dynamic action generation.

Although several benchmarks have focused on evaluating models' reasoning and tool-use capabilities in multi-step spatial tasks, for example, GeoBenchX~\cite{krechetova2025geobenchx}, which examines collaborative reasoning through the integration of multiple geographic functions, and MapBench~\cite{xing2025can}, which investigates hierarchical reading and attention mechanisms for multi-scale map information from a cognitive science perspective. Other benchmarks such as ReasonMap~\cite{feng2025can}, RewardMap~\cite{feng2025rewardmap}, and TraveLLM~\cite{fang2024travellm} further evaluate autonomous reasoning and decision-making in complex scenarios by incorporating long-tracing logical reasoning, reinforcement learning feedback, and dynamic perturbation settings.

However, despite substantial progress in benchmarking spatial reasoning, existing image-based benchmarks are typically confined to single-image settings and thus fail to capture the multi-factor considerations that characterize real-world scenarios, such as time, cost and comfort. Many also evaluate tool-use ability or rely on external tools and symbolic verifiers to compensate for models' weaknesses in certain sub-skills, which in turn obscures the assessment of their native capabilities. Furthermore, their scale is often limited by the difficulty of obtaining high-quality map annotations.

In summary, existing RP-based benchmarks remain insufficient for comprehensively evaluating MLLMs in realistic decision-making settings. Table~\ref{tab:benchmark_comparison} compares MapTab with related benchmarks in terms of input modality, graph--table integration, route planning, multi-criteria reasoning, tool assumptions, and scale. To address the identified gaps, MapTab incorporates realistic multi-criteria considerations, evaluates the native capabilities of MLLMs without external tools, and introduces a dedicated pipeline for large-scale benchmark construction.

\begin{table*}[t]
    \centering

    \caption{Comparison of MapTab with existing multimodal reasoning and planning benchmarks. \textbf{G+T}, \textbf{RP}, and \textbf{MC} denote Graph+Table, Route Planning, and Multi-Criteria, respectively.}
    \label{tab:benchmark_comparison}
    
    \renewcommand{\arraystretch}{1.15}
    \setlength{\tabcolsep}{2.5pt}
    \footnotesize
    
    \begin{tabular*}{\textwidth}{@{\extracolsep{\fill}}>{\raggedright\arraybackslash}m{5cm}>
    {\centering\arraybackslash}m{2.0cm}>{\centering\arraybackslash}m{0.80cm}>{\centering\arraybackslash}m{0.95cm}>{\centering\arraybackslash}m{0.85cm}>{\centering\arraybackslash}m{0.6cm}>{\centering\arraybackslash}m{1.15cm}>{\raggedright\arraybackslash}m{5.0cm}@{}}
    \toprule
    
    \multicolumn{1}{>{\centering\arraybackslash}m{5cm}}{\textbf{Category / Representative Works}} & \textbf{Input} & \textbf{G+T} & \textbf{RP} & \textbf{MC} & \textbf{Tools} & \textbf{Scale} & \multicolumn{1}{>{\centering\arraybackslash}m{5.0cm}}{\textbf{Main Gap}} \\
    \midrule
    \textbf{Visual Reasoning}\newline V*Bench, VisualPuzzles, VisuLogic, R-Bench, VGRP-Bench & Images & $\times$ & $\times$ & $\times$ & $\times$ & Large & Focus on abstract visual and logical reasoning rather than route planning. \\
    \midrule[0.3pt]
    \textbf{General Multimodal Reasoning}\newline MathVQA, MMMU, MathVerse & Images + Text & $\times$ & $\times$ & $\times$ & $\times$ & Large & Evaluate general multimodal reasoning rather than planning tasks. \\
    \midrule[0.3pt]
    \textbf{Map Understanding}\newline MapEval, MapQA, MapIQ, CartoMark, MapBench & Maps & $\times$ & Limited & $\times$ & $\times$ & Medium & Focus on map perception, OCR, and QA rather than constrained route planning. \\
    \midrule[0.3pt]
    \textbf{Spatial Reasoning}\newline CityBench, DriveBench, FRIEDA, GeoX-Bench, CartoMapQA, MapLM & Maps & Partial & Partial & $\times$ & $\times$ & Medium & Evaluate spatial reasoning without explicit multi-criteria optimization. \\
    \midrule[0.3pt]
    \textbf{Planning / Navigation}\newline PlanAgent, PReP, GeoNav, NavBench, GeoBenchX, ReasonMap, RewardMap, TraveLLM & Maps / Environment & Partial & \checkmark & Partial & \checkmark & Small--Medium & Focus on navigation, agent planning, or embodied execution, often with external tools. \\
    \midrule[0.3pt]
    \textbf{MapTab (Ours)} & \textbf{Map + Table} & \checkmark & \checkmark & \checkmark & $\times$ & \textbf{Large} & \textbf{The first benchmark that evaluates native multimodal route planning over map images and structured tables under explicit multi-criteria constraints, requiring models to solve the task without external tools or tool-use assumptions.} \\
    \bottomrule
    \end{tabular*}
\end{table*}

\section{More Details for Dataset Construction}\label{sec:data_construct}
\subsection{Multi-Criteria Tabular Construction}\label{sec:struct_data}
Compared with the ``image + unstructured text'' paradigm, the ``image + table'' paradigm provides a stronger structural prior. In route-planning tasks, tables represent node- and edge-level attributes in a low-entropy and high signal-to-noise form, reducing the semantic ambiguity and redundancy of natural-language descriptions. Accordingly, each map is paired with an \textbf{Edge\_tab} and a \textbf{Vertex\_tab} to complement its visual topology with multi-criteria information. The map images and topological structures are derived from real-world resources, whereas the numerical attributes are synthesized according to predefined schemas, value ranges, and generation rules, making MapTab a semi-synthetic benchmark. Gemini-3-Flash is used only to reduce repetitive human annotation costs, and all generated entries are subsequently processed by automated validation and manual correction.

Metromap is designed to simulate real-world urban commuting. Its Edge\_tab contains six fields: \textit{Edge}, \textit{Line}, \textit{Time}, \textit{Price}, \textit{Comfort Level}, and \textit{Reliability}. The \textit{Edge} field follows the topological annotation format of FarPlanning~\cite{cao2023representation}, while the remaining attributes describe metro operations and user experience (see Appendix~\ref{subsubsec:metromap_Edge_tab}). The corresponding Vertex\_tab additionally includes \textit{Transfer Time} to model delays caused by transfers between metro lines (see Appendix~\ref{subsubsec:metromap_Vertex_tab}).

Travelmap focuses on attraction-level semantics and user travel experience. Because metro lines and transfer times are not applicable to tourist-route planning, the \textit{Line} and \textit{Transfer Time} fields are removed. Its Edge\_tab and Vertex\_tab each retain an entity identifier together with four quantitative attributes: \textit{Time}, \textit{Price}, \textit{Comfort Level}, and \textit{Reliability}. Their values are synthesized from plausible real-world travel ranges, ratings, and the corresponding map structure (see Appendix~\ref{subsubsec:travelmap_Ed_tab} and~\ref{subsubsec:travelmap_ve_tab}). MapTab further provides both CSV and JSON representations for analyzing the influence of tabular formats (see Appendix~\ref{subsec:abl_tabular}).

\subsection{RP Queries Construction}\label{subsec:RP_query_construct}
MapTab contains 12 route-planning query categories, including 3 criteria-free and 9 criteria-based settings. All queries and reference labels are generated from Edge\_tab and Vertex\_tab using deterministic Python programs.

\subsubsection{Criteria-Free Queries}
Criteria-free queries evaluate route planning without numerical attributes under three input settings: \textit{Map-only}, \textit{Edge\_tab-only}, and \textit{Map+Edge\_tab}. In this setting, Edge\_tab retains only the \textit{Edge} column and provides graph connectivity without decision criteria.

Origin--destination pairs are sampled from the topology in Edge\_tab, and shortest paths are computed using Dijkstra's algorithm. Each route contains at least three stations to avoid trivial cases, and sampling continues until the quota for each map and difficulty level is reached.

Each input setting uses five semantically equivalent instruction templates: \textit{Direct Optimization}, \textit{Graph-theoretic}, \textit{Instruction-seeking}, \textit{User-oriented}, and \textit{Constraint-explicit}. These templates are cyclically assigned to reduce instruction bias.

\subsubsection{Criteria-Based Queries}
Criteria-based queries include four single-criterion settings—\textit{Time}, \textit{Price}, \textit{Comfort Level}, and \textit{Reliability}—and five multi-criteria combinations. For \textit{Map+Edge\_tab+Vertex\_tab}, the combinations simulate different user preferences: younger users prioritize \textit{Time+Price+Reliability}, middle-aged users prioritize \textit{Time+Comfort Level+Reliability}, older users prioritize \textit{Price+Comfort Level+Reliability}, and families consider all four criteria.

MapTab retains Edge\_tab and Vertex\_tab and additionally constructs Mix\_tab by integrating their numerical attributes. Under the \textit{Map+Mix\_tab} setting, all four criteria are considered. Only columns relevant to each query are provided during evaluation to reduce context length.

\subsubsection{Reference Routes and Labels}
Reference routes are derived from the topology in Edge\_tab. Criteria-free routes are computed using Dijkstra's algorithm. For criteria-based queries, values from different dimensions are normalized to a common scale and combined through weighted aggregation to identify the optimal route.

Transfer stations are marked by appending (transfer)'' to the station name, producing routes such as A-B(transfer)-C''. Each label also records Total\_Time, Total\_Price, Average\_Comfort Level, and Average\_Reliability for multi-criteria analysis and ablation studies.

\subsection{QA Queries Construction}\label{subsec:QA_query_construct}
MapTab constructs 24 QA task types across Metromap and Travelmap, with 12 task types for each scenario. These tasks cover three dimensions: \textit{Global Perception-based Reasoning}, \textit{Local Perception-based Reasoning}, and \textit{Spatial Relationship Judgment}. They are designed to diagnose eight capabilities required for route planning: \textbf{Visual Perception}, \textbf{Table Understanding}, \textbf{Cross-modal Alignment}, \textbf{Graph Topology Reasoning}, \textbf{Spatial Localization}, \textbf{Numerical Reasoning}, \textbf{Path Planning}, and \textbf{Global Reasoning}.

\subsubsection{Input Settings}
QA queries use four input settings: \textit{Map-only}, \textit{Edge\_tab-only}, \textit{Vertex\_tab-only}, and \textit{Map+Mix\_tab}. The table-only settings evaluate models' understanding and reasoning over structured data, while \textit{Map+Mix\_tab} evaluates their ability to integrate visual topology with tabular attributes.

For Metromap under the \textit{Map+Mix\_tab} setting, the \textit{Line} column is removed to establish a strict \textit{Cross-modal Necessity} setting. \textit{Mix\_tab} retains station-level attributes and merged numerical information but does not provide explicit line topology. Therefore, models must recover the topological structure from the map and align it with the corresponding station information in the table.

\subsubsection{Instruction Templates}
Each QA task type is expressed using five semantically equivalent instruction styles: \textit{Direct / Neutral}, \textit{Simple \& Conversational}, \textit{Slightly Formal}, \textit{Technical / Academic Style}, and \textit{Concise / Dataset-Friendly}. These templates are cyclically assigned to reduce sensitivity to variations in instruction wording.

\subsubsection{Answer Generation and Verification}
All QA tasks are formulated as binary judgment or multiple-choice questions. Ground-truth answers that can be derived from \textit{Edge\_tab} and \textit{Vertex\_tab} are generated using deterministic Python programs. For tasks that cannot be fully covered by rule-based procedures, Gemini-3-Flash~\cite{google2025gemini3flash} generates preliminary answers, which are subsequently verified by human annotators.

During inference, model predictions are extracted from the content enclosed by $<\text{answer\_begin}>$ and $<\text{answer\_end}>$ to ensure consistent automatic evaluation.

\subsection{RP Query Distribution Based on Difficulty Classification Metrics}\label{subsec:PR_Query_distribution}

To achieve fair and balanced allocation of query numbers for Metromap and Travelmap across training and test sets, we propose the following distribution strategy:

\begin{enumerate}
\item \textbf{Input and target definition:}  
Given a set of cities/attractions $\mathcal{C}$, total query number $Q$, weighting exponent $\alpha$, and minimum query threshold $Q_{\min}$, the algorithm takes these parameters as input and outputs the query number $q_c$ for each city/attraction.

\item \textbf{Difficulty grouping and target initialization (Phase 1):}  
Based on the Map Difficulty labels (Hard, Medium, Easy), cities/attractions are divided into three groups $G_{\text{Easy}}$, $G_{\text{Medium}}$, and $G_{\text{Hard}}$.  
The base query number is set as $\text{base} = \lfloor Q/3 \rfloor$. The target query numbers for the Medium and Hard groups, $T_{\text{Medium}}$ and $T_{\text{Hard}}$, are both initialized to this base value, while the remaining queries are assigned to the Easy group as $T_{\text{Easy}} = Q - 2 \times \text{base}$, ensuring conservation of the total query count.

\item \textbf{Training and test target allocation:}  
Within each difficulty group, cities/attractions are first separated according to image set type labels (training\_set or test\_set). Target query numbers $target_{\text{train}}$ and $target_{\text{test}}$ are then assigned to ensure a 4:1 ratio between training and test queries.  
This ratio applies to query counts rather than image counts (the image ratio is 2:1).

\item \textbf{Intra-group weighted allocation (Phase 2):}  
Within each difficulty group and split (six groups in total), intra-group weights are computed based on the $\alpha$-th power of each city's/attraction's Vertex Numbers. The total group weight is  
\[
W = \sum_{c \in S} (c.\text{Vertex Numbers})^\alpha .
\]
Queries $q_c$ are then allocated proportionally to ensure consistency with these weights.

\item \textbf{Rounding and minimum threshold criteria:}  
Within each group, all cities/attractions except the last one apply rounding and enforce the minimum threshold $Q_{\min}$.  
The last city/attraction absorbs rounding errors to ensure that the group's total query count exactly matches $target_s$. This city is recorded as $LastCity_{d,s}$ for later compensation.

\item \textbf{Global ratio compensation (Phase 3):}  
After intra-group allocation, the total number of test queries is checked against the global target of 20\% (i.e., $0.2Q$).  
If an excess $E>0$ exists, queries are preferentially deducted from the last city/attraction in the Easy-test group ($LastCity_{\text{Easy,test}}$) and evenly compensated to the last cities/attractions in the Medium-training and Hard-training groups ($LastCity_{d,\text{train}}$).

\item \textbf{Minimum threshold consistency check:}  
After compensation, the adjusted last cities/attractions are checked to ensure their query counts do not fall below $Q_{\min}$.  
If violations occur, the second-to-last city/attraction is introduced into the compensation process, guaranteeing strict enforcement of the minimum threshold while maintaining the 4:1 training-test query ratio.
\end{enumerate}

This algorithm is applicable to both Metromap and Travelmap. Taking Metromap as an example, the full algorithmic workflow is illustrated as Algorithm \ref{alg:query_dist}.

\begin{algorithm*}[!htt]
   \caption{Balanced City Query Distribution Strategy}
   \label{alg:query_dist}
\begin{algorithmic}
   \STATE {\bfseries Input:} City Data $\mathcal{C}$, Total Queries $Q$, Exponent $\alpha$, Min Threshold $Q_{\min}$
   \STATE {\bfseries Output:} Query Numbers $q_c$ for each city $c$
   
   \STATE \textit{// Phase 1: Grouping and Target Initialization}
   \STATE Partition $\mathcal{C}$ into $G_{\text{Easy}}, G_{\text{Medium}}, G_{\text{Hard}}$ based on Map Difficulty.
   \STATE Set $base = \lfloor Q/3 \rfloor$.
   \STATE Set $T_{\text{Medium}} = base, \quad T_{\text{Hard}} = base, \quad T_{\text{Easy}} = Q - 2 \times base$.

   \STATE \textit{// Phase 2: Weighted Distribution per Group}
   \FOR{each difficulty $d \in \{\text{Easy, Medium, Hard}\}$}
      \STATE Set $target_{\text{train}} = \lfloor T_{d} \times 0.8 \rfloor$ and $target_{\text{test}} = T_{d} - target_{\text{train}}$.
      
      \FOR{each subset type $s \in \{\text{train, test}\}$}
         \STATE Let $S$ be the set of cities in $G_d$ of type $s$.
         \STATE Calculate weight sum $W = \sum_{c \in S} (c.\text{Vertex Numbers})^\alpha$.
         \STATE Initialize $assigned = 0$.
         \FOR{each city $c$ in $S$}
            \IF{$c$ is not the last city}
               \STATE $raw\_q = (c.\text{Vertex Numbers})^\alpha / W \times target_s$
               \STATE $q_c = \max(\text{round}(raw\_q), Q_{\min})$
            \ELSE
               \STATE $q_c = target_s - assigned$ \COMMENT{Handle rounding remainder}
               \STATE Record $c$ as $LastCity_{d, s}$ for potential compensation.
            \ENDIF
            \STATE $assigned = assigned + q_c$
         \ENDFOR
      \ENDFOR
   \ENDFOR

   \STATE \textit{// Phase 3: Global Ratio Compensation (Strict 4:1)}
   \STATE Calculate test excess $E = \sum_{c \in \text{Test}} q_c - (Q \times 0.2)$.
   
   \IF{$E > 0$}
      \STATE \textit{// Step A: Reduce excess from Easy-Test}
      \STATE Let $u = LastCity_{\text{Easy, test}}$.
      \STATE Update $q_u = q_u - E$.
      
      \STATE \textit{// Step B: Distribute excess to Medium/Hard Training}
      \STATE Set $add\_amt = E / 2$.
      \FOR{$d \in \{\text{Medium, Hard}\}$}
         \STATE Let $v = LastCity_{d, \text{train}}$.
         \IF{$q_v + add\_amt < Q_{\min}$}
            \STATE Let $v$ be the penultimate city in $G_{d}.\text{train}$.
         \ENDIF
         \STATE Update $q_v = q_v + add\_amt$.
      \ENDFOR
   \ENDIF
\end{algorithmic}
\end{algorithm*}

In the concrete setup, the total query numbers are set to $Q=8000$ for Metromap and $Q=8400$ for Travelmap.  
The intra-group weighting exponent is fixed to $\alpha=1.5$, and the minimum query threshold is uniformly set to $Q_{\min}=5$.

The significance of this algorithm lies in achieving fair and balanced query allocation across multi-city/multi-attraction and multi-difficulty experimental settings. It prevents large cities/attractions from monopolizing query resources while ensuring sufficient data support for smaller ones, strictly maintains training-test ratios, and preserves reasonable data distributions, thereby guaranteeing sufficient model training and reliable evaluation.  
Moreover, the allocation results are traceable and controllable, ensuring that each city/attraction meets the minimum query threshold and providing a stable, interpretable, and reliable data foundation for multi-city/multi-attraction experiments.

\subsection{Quality Control: Metromap Validation Algorithm}\label{subsec:Metromap_validation_alg}

This section introduces a Metromap-specific validation algorithm designed to ensure the accuracy and consistency of metro line data in Edge\_tab and Vertex\_tab.  
The algorithm includes duplicate detection, station category comparison, vertex \& edge cross-validation, transfer station annotation verification, and manual inspection, comprehensively identifying potential annotation errors and data inconsistencies:

\begin{enumerate}
\item \textbf{Duplicate detection:}  
First, edges in Edge\_tab and stations in Vertex\_tab are checked for duplication within the same metro line (Line).  
Specifically, if station A-B and B-A both appear, they are treated as the same edge. This step ensures the absence of redundant or duplicated metro information and verifies graph correctness.

\item \textbf{Station category set comparison:}  
Station categories in Edge\_tab and Vertex\_tab are collected and compared. This process checks for category consistency across the two files and helps identify annotation errors, especially when vertex and edge errors occur in different files.

\item \textbf{Vertex \& edge cross-validation:}  
Based on the relationship between Edge\_tab and Vertex\_tab, the validation computes:  
$(\text{total number of edges in Edge\_tab}) + (\text{number of metro line categories})-(\text{number of loop lines})$,  
which should equal the total number of station occurrences in Vertex\_tab.  
Ordinary stations are counted once, while transfer stations are counted according to the number of lines they belong to.  
This method detects cases where both Edge\_tab and Vertex\_tab contain errors that escape conventional validation.

\item \textbf{Transfer station annotation verification:}  
Transfer stations are checked separately in Edge\_tab and Vertex\_tab by expanding the set of lines each transfer station belongs to and comparing consistency across the two files.  
Transfer station identification criteria are defined as follows:
\begin{itemize}
\item \textbf{Edge\_tab:} if the same station name appears on multiple metro lines.
\item \textbf{Vertex\_tab:} if a station's Line column contains multiple lines or the Transfer Time value is greater than zero.
\item \textbf{Loop line identification:} in Edge\_tab, if the first and last stations of the same line are identical, the line is considered a loop line. This step ensures accurate transfer station annotation and avoids errors arising from loop line special cases.
\end{itemize}

\item \textbf{Manual verification:}  
Finally, Edge\_tab and Vertex\_tab are manually checked against real metro maps to identify errors not detectable through automated validation.
\end{enumerate}

\subsection{Quality Control: Travelmap Validation Algorithm}\label{subsec:travelmap_val_alg}

The Travelmap-specific validation algorithm retains steps 1, 2, and 5 from the Metromap validation algorithm, with appropriate modifications to step 2.  
For Travelmap, Vertex\_tab removes points that are completely disconnected from all others, retaining only connected vertices.  
In addition, Edge\_tab is required to contain a single fully connected graph, preventing the existence of multiple disconnected subgraphs.

\section{Metrics}\label{sec:metrics}
This section proposes a unified evaluation framework designed to systematically assess model performance, map structural complexity, and query difficulty. 

\subsection{Performance Metrics}\label{subsec:performance_metrics}
For model performance evaluation, the RP task adopts three core metrics Exact Match Accuracy (EMA), Partial Match Accuracy (PMA), and Difficulty-aware Score (DS) to comprehensively measure model performance in terms of path correctness, linguistic consistency, and format compliance. Given that the output format is strictly constrained by prompts, any result that does not conform to the required format is regarded as incorrect. Such format generation failures themselves indicate deficiencies in the model's instruction understanding and execution capabilities. For QA tasks, evaluation is conducted solely based on Accuracy (Acc). 

We first describe the performance metrics for the RP task in detail:

\textbf{1. Exact Match Accuracy (EMA)}

A score of 1 is assigned if the path generated by the model exactly matches the reference path in both station order and content; otherwise, a score of 0 is assigned. To tolerate minor spelling errors, a generated station name is considered identical to the reference station if the string similarity exceeds 50\%. The similarity is computed based on character sequence matching, ignoring case differences and leading or trailing whitespaces. For Metromap, stations involving transfers must be explicitly annotated with the suffix ``transfer''; otherwise, the station is considered incorrect. This transfer annotation requirement does not apply to Travelmap. Notably, when multiple equally optimal routes exist, matching any one of them is considered correct.
    
\textbf{2. Partial Match Accuracy (PMA)} 
   
PMA measures the length of the longest contiguous correct prefix in the generated path, starting from the origin. Evaluation is conducted at the station level, counting only the continuously matched prefix until the first mismatched station appears (excluding the mismatched station). The PMA score is computed as the ratio of the length of this correct prefix to the total length of the reference path. This metric is preferred over simply measuring the proportion of correctly repeated stations, as we assume that once an error occurs in the path, subsequent planning steps become invalid and no longer reflect meaningful routing decisions. If there are multiple optimal reference paths, the highest PMA score among them is used. 
    
\textbf{3. Difficulty-aware Score (DS)}    
  
DS explicitly incorporates task difficulty into the evaluation. Each sample is assigned a discrete score ranging from 2 to 6 based on the combination of its Map\_difficulty and Query\_difficulty, denoted as Map\_difficulty-Query\_difficulty. Specifically, Easy, Medium, and Hard correspond to scores of 1, 2, and 3, respectively, and the final difficulty score is obtained by summing the two values. The difficulty score is counted only when EMA equals 1. This metric effectively reflects the upper bound of model performance, particularly its capability in handling high-difficulty tasks. 

For QA-query tasks, all questions are either binary (true/false) or fill-in-the-blank. Evaluation is conducted using Accuracy (Acc): a score of 1 is assigned if the model-generated answer exactly matches the reference answer; otherwise, a score of 0 is assigned. 

\subsection{Difficulty Classification Metrics} \label{subsec:difficulty_metrics}
This section introduces the classification methods for Map Difficulty and Query Difficulty. The core idea is to sort each Map and Query based on defined evaluation metrics and divide them into three levels: Hard, Medium, and Easy, with a ratio of 1:1:1. In Map Difficulty Classification, three indicators are considered: Graph Size (GS), Weighted Average Shortest Path (WASP), and Meshedness Coefficient (MC). In Query Difficulty Classification, addressing the specific scenarios of Metromap and Travelmap, the Shortest Path Length Index (SPLI) and Simple Path Complexity Index (SPCI) are adopted as difficulty metrics, respectively. These metrics comprehensively evaluate the complexity of Maps and Queries in RP tasks.

\subsubsection{Map Difficulty Classification}

\textbf{1. Graph Size (GS)}

Graph Size represents the overall ``scale'' of the graph, i.e., the sum of the number of nodes and edges:
\begin{equation}
    GS = |V| + |E|
\end{equation}
where $|V|$ denotes the number of nodes in the graph, and $|E|$ denotes the number of edges. This formula indicates that the size of the graph is equal to the sum of all nodes and edges, used to measure the complexity or overall volume of the graph.

\textbf{2. Weighted Average Shortest Path (WASP)}

WASP measures the average shortest path length between nodes in the graph, while considering the potential existence of multiple connected components. Let graph $G$ have $k$ connected components $C_1, C_2, \dots, C_k$, where the number of nodes in each connected component is $V_i = |C_i|$, and the shortest path length between nodes is $d(u, v)$. The unified expression is:
\begin{equation}
    \begin{aligned}
        WASP &= \frac{\sum_{i=1}^{k} |C_i| \cdot \text{ASPD}(C_i)}
        {\sum_{i=1}^{k} |C_i|} \\
        &= \frac{\sum_{i=1}^{k} |C_i| \cdot \text{ASPD}(C_i)}
        {|V|}
    \end{aligned}
\end{equation}

Here, the inner summation and $\text{ASPD}(C_i)$ are used to calculate the average shortest path length of node pairs within each connected component; the outer summation sums these weighted by the number of nodes. When the graph is connected ($k=1$), WASP is equivalent to the ordinary average shortest path length.

\textbf{3. Meshedness Coefficient (MC)}

Meshedness Coefficient measures the degree of density (tightness) between nodes in the graph, usually defined by the relationship between the number of edges and the number of cycles:
\begin{equation}
    MC = \frac{|E| - |V| + 2}{|E|}
\end{equation}
where $|V|$ represents the number of nodes, and $|E|$ represents the number of edges. When $|E| > |V| - 1$, cycles exist in the graph, and a larger Mesh Coefficient (MC) indicates more cycles in the graph and tighter connections between nodes.

For Metromap, each sample considers only two metrics: Graph Size (GS) and Weighted Average Shortest Path (WASP). Each indicator was first normalized and subsequently aggregated by computing their arithmetic mean. Samples are sorted by this average value and strictly divided into three levels: Hard, Medium, and Easy, with a ratio of 1:1:1. 
For Travelmap, the GS, WASP, and MC metrics are considered simultaneously. To ensure a balanced distribution of Map Difficulty levels in both the training and test sets, the algorithm groups every three consecutive samples after sorting. From each group, one sample is randomly selected for the test set, and the other two for the training set, thereby balancing the train/test ratio and difficulty levels.

\subsubsection{Query Difficulty Classification}

In RP tasks, the structural characteristics of Metromap and Travelmap differ significantly: the former has many intersection points and an extremely large number of simple paths between nodes, making the calculation of the average length of all simple paths overly complex, so it is only suitable to use the Shortest Path Length Index (SPLI) for measurement. The latter has a relatively small number of simple paths, so using the Simple Path Complexity Index (SPCI) to calculate the average length of all simple paths can more objectively reflect RP difficulty.

\textbf{1. Shortest Path Length Index (SPLI)}

SPLI measures the shortest path length from source point $s$ to target point $t$ in the graph, i.e., the number of edges passed by the shortest path between points. This metric is used to evaluate the connection efficiency between nodes. For Metromap, which has many intersections and a massive number of simple paths, using only the shortest path length can effectively characterize query difficulty:
\begin{equation}
    SPLI(s, t) = \min(L_1, L_2, \dots, L_n)
\end{equation}
where $L_i$ represents the length of all paths from $s$ to $t$ in the graph.

\textbf{2. Simple Path Complexity Index (SPCI)}

SPCI measures the average length of all simple paths (paths without repeating nodes) from source point $s$ to target point $t$ in the graph, used to characterize path complexity. A higher SPCI value indicates more selectable paths and a more complex structure. For Travelmap, it can objectively reflect the diversified connections and query difficulty between source and target points:
\begin{equation}
    SPCI(s, t) = \frac{1}{N}\sum_{i=1}^{N} l(L_i)
\end{equation}
where $N$ is the number of all simple paths from $s$ to $t$, and $l(L_i)$ is the length of path $L_i$.

The number of queries per image is determined by the algorithm described in Appendix~\ref{subsec:PR_Query_distribution}. Subsequently, after query generation, Query Difficulty Classification is performed for each image. In Metromap and Travelmap, queries are sorted according to their respective difficulty metrics and strictly divided into Hard, Medium, and Easy levels in equal proportions.

\section{Query Template}\label{sec:query_template}

\subsection{RP Query}\label{subsec:PR_query}
In this study, we design a series of RP queries based on \textit{Metromap} and \textit{Travelmap} to evaluate path optimization algorithms under different input modalities and criteria settings. Taking the \textit{Map-Only} setting in \textit{Metromap} as an example, where \{station\_1\} and \{station\_2\} denote randomly sampled origin-destination pairs, we formulate five types of queries with different instruction styles as follows:
\begin{enumerate}
    \item \textbf{Direct Optimization query}: According to the Subway Map, what is the path with the fewest stations from \{station\_1\} to \{station\_2\}?
    \item \textbf{Graph-theoretic}: According to the Subway Map, what is the shortest path from \{station\_1\} to \{station\_2\}?
    \item \textbf{Instruction-seeking}: According to the Subway Map, how can I reach \{station\_2\} from \{station\_1\} via the shortest path?
    \item \textbf{User-oriented Optimization}: According to the Subway Map, how can I travel from \{station\_1\} to \{station\_2\} with the fewest number of stations?
    \item \textbf{Constraint-explicit Optimization}: According to the Subway Map, what is the shortest route in terms of station count from \{station\_1\} to \{station\_2\}?
\end{enumerate}

All remaining queries are also constructed with these five instruction styles. For brevity, we list only the \textit{Direct Optimization query} style below:
\begin{itemize}
    \item \textbf{Criteria-free - Edge\_tab-Only}: According to Edge Table and Vertex Table, what is the path with the fewest stations from \{station\_1\} to \{station\_2\}?
    \item \textbf{Criteria-free - Map+Edge\_tab}: According to the Subway Map, Edge Table and Vertex Table, what is the path with the fewest stations from \{station\_1\} to \{station\_2\}?
    \item \textbf{Criteria-based - Map+Edge\_tab+Vertex\_tab}:
    \begin{enumerate}
        \item \textbf{Time-Only}: According to the Subway Map, Edge Table and Vertex Table, what is the fastest route from \{station\_1\} to \{station\_2\}?
        \item \textbf{Price-Only}: According to the Subway Map, Edge Table and Vertex Table, what is the cheapest route from \{station\_1\} to \{station\_2\}?
        \item \textbf{Comfort Level-Only}: According to the Subway Map, Edge Table and Vertex Table, what is the smoothest ride between \{station\_1\} and \{station\_2\}?
        \item \textbf{Reliability-Only}: According to the Subway Map, Edge Table and Vertex Table, what is the most stable subway route from \{station\_1\} to \{station\_2\}?
        \item \textbf{Time+Price+Reliability}: According to the Subway Map, Edge Table and Vertex Table, what is the optimal route considering time, price, and reliability from \{station\_1\} to \{station\_2\}?
        \item \textbf{Time+Comfort Level+Reliability}: According to the Subway Map, Edge Table and Vertex Table, what is the fastest, most comfortable, and reliable route from \{station\_1\} to \{station\_2\}?
        \item \textbf{Price+Comfort Level+Reliability}: According to the Subway Map, Edge Table and Vertex Table, what is the best route optimizing price, comfort, and reliability from \{station\_1\} to \{station\_2\}?
        \item \textbf{Time+Price+Comfort Level+Reliability}: According to the Subway Map, Edge Table and Vertex Table, what is the optimal route considering time, price, comfort, and reliability from \{station\_1\} to \{station\_2\}?
    \end{enumerate}
    \item \textbf{Multi-criteria - Map+Vertex\_tab}: According to the Subway Map and the Vertex Table, what is the optimal route considering time, price, comfort, and reliability from \{station\_1\} to \{station\_2\}?
\end{itemize}

For \textit{Travelmap}, the term ``Subway Map'' is replaced with ``Scenic Area Planning Map'' in all queries.

\subsection{QA Query}\label{subsec:qa_query}

In addition to RP queries, we design a series of QA queries based on \textit{Metromap} and \textit{Travelmap}. The experimental setup systematically considers four input settings: \textit{Map-Only}, \textit{Edge\_tab-Only}, \textit{Vertex\_tab-Only}, and \textit{Map+Mix\_tab}. For \textit{Metromap}, \textit{Mix\_tab} excludes the \textit{Line} column, creating a strict cross-modal setting in which the model must recover line topology from the map and associate it with the station attributes provided in the table.

Under each input setting, we construct three categories of tasks: \textit{Global Perception-based Reasoning Tasks (GP)}, \textit{Local Perception-based Reasoning Tasks (LP)}, and \textit{Spatial Relationship Judgment Tasks (SR)}. Each QA category is instantiated using five semantically equivalent but stylistically distinct instruction templates: \textit{Direct / Neutral}, \textit{Simple \& Conversational}, \textit{Slightly Formal}, \textit{Technical / Academic Style}, and \textit{Concise / Dataset-Friendly}.

Taking the \textit{Map-Only} \textit{Global Perception-based Reasoning Task (GP)} as an example, the five instruction styles are as follows:
\begin{enumerate}
    \item \textbf{Direct / Neutral}: How many metro lines are there in total on the Subway Map?
    \item \textbf{Simple \& Conversational}: How many subway lines are shown on the metro map?
    \item \textbf{Slightly Formal}: What is the total number of rail lines in the Subway Map?
    \item \textbf{Technical / Academic Style}: How many distinct transit lines does the metro map contain?
    \item \textbf{Concise / Dataset-Friendly}: What is the total count of metro lines in the map?
\end{enumerate}

All QA tasks follow the same five-style formulation. Below, we list a subset of queries using only the \textit{Direct / Neutral} style:
\begin{enumerate}
    \item \textbf{Metromap-Map - GP}: How many metro lines are there in total on the Subway Map?
    \item \textbf{Metromap-Map - LP}: How many stations are there between \{Station A\} and \{Station B\} on the same line?
    \item \textbf{Metromap-Map - SR}: Is \{Station A\} on Line X?
    \item \textbf{Metromap-Edge\_tab - GP}: What is the number of stations on the longest subway line?
    \item \textbf{Metromap-Edge\_tab - LP}: What is the Time/Price/Comfort Level value for the edge \{Station A\}-\{Station B\}?
    \item \textbf{Metromap-Edge\_tab - SR}: How many edges are there between the edge \{Station A\}-\{Station B\} and the edge \{Station C\}-\{Station D\}?
    \item \textbf{Metromap-Vertex\_tab - GP}: How many transfer stations are there in the entire table?
    \item \textbf{Metromap-Vertex\_tab - LP}: How many transfer stations are there on \{Line X\}?
    \item \textbf{Metromap-Vertex\_tab - SR}: Are \{Station A\} and \{Station B\} on the same line?
    \item \textbf{Metromap-Map+Mix\_tab - GP}: What is the total stop time of the subway line with the longest stop time?
    \item \textbf{Metromap-Map+Mix\_tab - LP}: What is the average reliability from \{Station A\} to \{Station B\} on the same line?
    \item \textbf{Metromap-Map+Mix\_tab - SR}: Please check whether there are transfer stations between \{Line X\} and \{Line Y\}. If so, return the shortest transfer time; otherwise, return 0. If there is only one subway line, return 1.
    \item \textbf{Travelmap-Map - GP}: How many tourist attractions are there in total?
    \item \textbf{Travelmap-Map - LP}: Is scenic spot \{Spot A\} on the circular route?
    \item \textbf{Travelmap-Map - SR}: How many scenic spots are adjacent to \{Spot A\}?
    \item \textbf{Travelmap-Edge\_tab - GP}: How many times does the most frequently appearing scenic spot occur in the table?
    \item \textbf{Travelmap-Edge\_tab - LP}: What is the Time/Price value for \{Spot A\}-\{Spot B\}?
    \item \textbf{Travelmap-Edge\_tab - SR}: Is the shortest distance between \{Spot A\} and \{Spot B\} less than 5?
    \item \textbf{Travelmap-Vertex\_tab - GP}: How many scenic spots in the table have a Time value below 30?
    \item \textbf{Travelmap-Vertex\_tab - LP}: Given \{Spot A\}, \{Spot B\}, \{Spot C\}, and \{Spot D\}, what is the price of the cheapest one?
    \item \textbf{Travelmap-Vertex\_tab - SR}: Is \{Spot A\} located above \{Spot B\} in the table?
    \item \textbf{Travelmap-Map+Mix\_tab - GP}: Is the number of scenic spots in the table the same as the actual number of scenic spots in the map?
    \item \textbf{Travelmap-Map+Mix\_tab - LP}: Are there any locations with Time $<$ 30 along all shortest paths from \{Spot A\} to \{Spot B\}?
    \item \textbf{Travelmap-Map+Mix\_tab - SR}: What is the cost of the lowest-priced location adjacent to \{Spot A\}?
\end{enumerate}

This task suite is designed to reveal the capabilities required for subsequent RP tasks, including map perception, table understanding, cross-modal alignment, topology reasoning, numerical reasoning, and path planning. By covering different input settings, reasoning categories, and instruction styles, it supports a comprehensive evaluation of model performance across both Metromap and Travelmap.

\section{Prompt Template}\label{sec:prompt_template}

In this study, the prompt templates are divided into two categories: one for RP tasks and the other for QA tasks. Since the prompt construction methods for \textit{Metromap} and \textit{Travelmap} are identical, we take the prompts for \textit{Metromap} as an illustrative example below.

\subsection{Prompts for Generating Edge\_tab and Vertex\_tab}\label{subsec:prompt_template_1}

In this work, we design the \textit{Edge\_tab} and \textit{Vertex\_tab} attributes for \textit{Metromap} and \textit{Travelmap} based on real-world transportation networks and connectivity characteristics between metro systems and scenic areas. Specifically, the attributes in \textit{Edge\_tab} capture key operational characteristics of metro systems and scenic routes, including time, price, comfort level, and reliability, which are used to simulate the operational properties of different paths and the corresponding user experience. In contrast, the attributes in \textit{Vertex\_tab} focus on the characteristics of metro stations and scenic spots, covering dwell time, comfort level, price, reliability, and transfer time, thereby modeling the transfer complexity between stations or attractions as well as the passenger experience at each location. 

Through the design of these attributes, we aim to faithfully reflect the diverse factors that must be considered in RP tasks, ensuring the scientific validity and rationality of model behavior during path selection and optimization. We use a heuristic design instead of directly adopting real-world data because these attributes are difficult to quantify consistently and accurately in practice. For instance, comfort scores may differ across platforms due to variations in rating systems, user groups and evaluation criteria, and even the same object may receive inconsistent assessments. Moreover, such attributes are inherently subjective and cannot be fully represented by any single real-world data source. We therefore employ simulated attribute distributions inspired by real-world scenarios to ensure benchmark controllability and reproducibility, while leaving the integration of more realistic data sources as an important direction for future work.

The specific attribute designs and their underlying rationales are described below.

\subsubsection{Metromap-Edge\_tab}\label{subsubsec:metromap_Edge_tab}
Based on real-world characteristics between adjacent metro stations, we design the attributes in the \textit{Metromap Edge\_tab} as follows:
\begin{enumerate}
    \item \textbf{Line}: This attribute indicates the metro line to which a station belongs, represented in the form of \textit{x Line}, where \textit{x} denotes the line number or name.  
    
    \textbf{Design Rationale}: This attribute is used solely to annotate the metro line associated with each segment, facilitating the distinction between different lines and clarifying the connectivity relationships between stations.
    
    \item \textbf{Time}: This attribute simulates the travel time between Station A and Station B, with values ranging from 2.00 to 3.00 minutes, rounded to two decimal places.  
    
    \textbf{Design Rationale}: This range reflects typical travel durations between adjacent stations in urban metro systems.
    
    \item \textbf{Price}: This attribute takes values ranging from 0.00 to 1.50 RMB, rounded to two decimal places.  
    
    \textbf{Design Rationale}: This range is intended to simulate a broad spectrum of metro fare levels, which commonly fluctuate within this interval across many cities.
    
    \item \textbf{Comfort Level}: This attribute models passengers' perceived comfort, with values ranging from 0.00 to 1.00, rounded to two decimal places.  
    
    \textbf{Design Rationale}: The range from 0.00 to 1.00 represents a spectrum from extremely poor to excellent riding experiences. Considering that real-world metro comfort may be affected by factors such as carriage crowding, air quality, and noise, this range reasonably captures variations in passenger comfort perception.
    
    \item \textbf{Reliability}: This attribute is set to 0.  
    
    \textbf{Design Rationale}: Since metro services would not operate if they were unreliable, all routes in this model are assumed to be reliable by default, and reliability is therefore not explicitly differentiated.
\end{enumerate}

The generation prompt for \textit{Edge\_tab} in \textit{Metromap} as implemented in the Gemini-3-Flash model is shown in \textbf{Prompt: Metromap Edge\_tab Generation}.

\subsubsection{Metromap-Vertex\_tab}\label{subsubsec:metromap_Vertex_tab}
Based on real-world characteristics of metro stations, we design the attributes in the \textit{Metromap Vertex\_tab} as follows:
\begin{enumerate}
    \item \textbf{Line}: This attribute indicates the metro line to which a station belongs, represented in the form of \textit{x Line}, where \textit{x} denotes the line number or name.  
    
    \textbf{Design Rationale}: This attribute is used to annotate the line associated with each metro station, enabling the model to distinguish stations belonging to different lines.
    
    \item \textbf{Time}: This attribute simulates the dwell time at each metro station, with values ranging from 0.50 to 2.00 minutes, rounded to two decimal places.  
    
    \textbf{Design Rationale}: This attribute models the time a train stops at a station, with values determined by simulated passenger flow and station-specific characteristics.
    
    \item \textbf{Price}: This attribute is set to 0.  
    
    \textbf{Design Rationale}: Since no additional cost is incurred during station dwell time, pricing is only considered during metro travel segments.
    
    \item \textbf{Comfort Level}: This attribute models passengers' perceived comfort, with values ranging from 0.00 to 1.00, rounded to two decimal places.  
    
    \textbf{Design Rationale}: The range from 0.00 to 1.00 is designed to simulate variations in boarding and alighting volumes as well as station crowding levels.
    
    \item \textbf{Reliability}: This attribute takes values ranging from 0.50 to 1.00, rounded to two decimal places.  
    
    \textbf{Design Rationale}: This range simulates the probability of station disruptions and downtime caused by unexpected events (e.g., security incidents or equipment failures), reflecting reliability differences across stations.
    
    \item \textbf{Transfer Time}: This attribute represents the time required for transfers, with values ranging from 5 to 15 minutes, specified as integers.  
    
    \textbf{Design Rationale}: This range reflects the time cost of transfers between stations, varying according to transfer complexity and the physical distance between metro lines.
\end{enumerate}

The generation prompt for \textit{Vertex\_tab} in \textit{Metromap} as implemented in the Gemini-3-Flash model is shown in \textbf{Prompt: Metromap Vertex\_tab Generation}.

\subsubsection{Travelmap-Edge\_tab}\label{subsubsec:travelmap_Ed_tab}
Based on real-world transportation networks and road connectivity characteristics between tourist attractions, we design the attributes in the \textit{Travelmap Edge\_tab} as follows:
\begin{enumerate}
    \item \textbf{Time}: This attribute simulates the commuting time from one attraction to another. If the map explicitly provides a time annotation, that value is directly used. If only distance information is available and the distance exceeds 1 km, the time is computed assuming a speed of 60 km/h. If the distance is less than 1 km, a speed of 4 m/s is assumed and the resulting time is converted into minutes. Any computed time shorter than 1 minute is rounded up to 1 minute. If neither time nor distance information is provided, a random integer between 10 and 60 minutes is generated.  
    
    \textbf{Design Rationale}: Commuting time is inferred from the time or distance annotations provided on the map, or otherwise simulated using reasonable assumptions based on common transportation knowledge.
    
    \item \textbf{Price}: This attribute is a randomly generated integer ranging from 10 to 80, with smaller values occurring more frequently.  
    
    \textbf{Design Rationale}: This attribute simulates commuting costs, reflecting price variations across different routes or transportation modes. The biased random generation toward lower prices captures realistic cost distribution patterns observed in daily travel.
    
    \item \textbf{Comfort Level}: This attribute models passengers' perceived comfort using a travel-review-style rating scheme (e.g., from tourism applications such as Ctrip or Dianping). A single-decimal value is randomly generated within the range of 1.0 to 5.0, with ratings $\geq$ 3.0 accounting for the majority.  
    
    \textbf{Design Rationale}: This attribute simulates comfort experiences across different routes or transportation modes, where higher comfort scores are typically associated with better vehicles, routes, or environmental conditions.
    
    \item \textbf{Reliability}: This attribute is defined as $a/365$, where $a$ is an integer ranging from 1 to 365, with values greater than 180 occurring more frequently. The final value is reported with six decimal places.  
    \textbf{Design Rationale}: This attribute simulates route reliability by accounting for potential disruptions such as road construction, security controls, or closures of high-risk areas. Here, $a$ represents the number of days a route is effectively accessible within a year; larger values indicate higher reliability, reflecting lower frequencies of closures or access restrictions due to special events.
\end{enumerate}
The generation prompt for \textit{Edge\_tab} in \textit{Travelmap} as implemented in the Gemini-3-Flash model is shown in \textbf{Prompt: Travelmap Edge\_tab Generation}.

\subsubsection{Travelmap-Vertex\_tab}\label{subsubsec:travelmap_ve_tab}
Based on real-world characteristics of tourist attractions and transportation facilities, we design the attributes in the \textit{Travelmap Vertex\_tab} as follows:
\begin{enumerate}
    \item \textbf{Time}: This attribute is generated as a random integer in the range of 30-180 minutes. For locations such as airports and railway stations, the time is directly set to 0.  
    
    \textbf{Design Rationale}: This attribute is intended to simulate the duration of stay at each attraction. Locations such as airports and stations typically do not involve sightseeing activities, and thus their time values are set to 0.
    
    \item \textbf{Price}: This attribute is a random integer ranging from 10 to 200 and is constrained to be a multiple of 5. For locations such as airports and railway stations, the price is randomly set to 100, 200, or 300.  
    
    \textbf{Design Rationale}: This attribute simulates entrance fees for attractions and service costs at transportation hubs. In RP tasks, transportation station costs are counted only at the final destination, while prices at intermediate stations are set to 0, reflecting the cost structures of different attractions and transportation facilities.
    
    \item \textbf{Comfort Level}: This attribute models perceived comfort using a travel-review-style rating scheme. A single-decimal value is randomly generated within the range of 1.0-5.0, with ratings $\geq$ 3.0 accounting for the majority.  
    
    \textbf{Design Rationale}: This design simulates tourists' comfort experiences when visiting attractions or using services at transportation hubs.
    
    \item \textbf{Reliability}: This attribute is defined as $a/365$, where $a$ is an integer ranging from 60 to 365, with values greater than 180 occurring more frequently. The final value is reported with six decimal places.  
    
    \textbf{Design Rationale}: This attribute simulates the reliability of different attractions or transportation facilities, reflecting whether attractions are open and whether transportation hubs experience service suspensions or closures. Here, $a$ denotes the number of days per year during which the location is operational; larger values indicate higher reliability.
\end{enumerate}

The generation prompt for \textit{Vertex\_tab} in \textit{Travelmap} as implemented in the Gemini-3-Flash model is shown in \textbf{Prompt: Travelmap Vertex\_tab Generation}.

\subsection{Prompts for RP Queries}\label{subsec:prompt_for_PR}

For RP tasks, the prompts in the Metromap dataset are divided into a total of 12 categories. Among them, three categories correspond to criteria-free prompts under different input modalities, namely Map-only, Edge\_tab-only, and Map+Edge\_tab. The remaining nine categories involve prompts with single or multiple multi-criteria.

Furthermore, in criteria-based Metromap scenarios, in order to enable a comprehensive consideration of different evaluation dimensions, normalization or transformation operations are applied to the indicators Time, Price, Comfort Level, Reliability, and Transfer Time. The specific preprocessing rules are as follows:
\begin{enumerate}
    \item \textbf{Time}: The original value range $[0, 3]$ is linearly scaled to $[0, 1]$.
    \item \textbf{Price}: The original value range $[0, 1.5]$ is linearly scaled to $[0, 1]$.
    \item \textbf{Transfer Time}: The original value range $[5, 15]$ is scaled according to the same benchmark ratio as $[0, 3]$, so as to ensure that transfer behaviors are assigned a significant penalty weight.
    \item \textbf{Comfort Level and Reliability}: Comfort and reliability (originally in the range $[0, 1]$) are subjected to an inverse transformation (i.e., $1 - x$) when jointly considered with other factors, and zero values are removed.
\end{enumerate}

Similarly, for multi-criteria Travelmap datasets, the four indicators, Time, Price, Comfort Level, and Reliability, also require corresponding scaling or inversion operations. The specific rules are as follows:

\begin{enumerate}
    \item \textbf{Time}: The original value range $[0, 180]$ is linearly scaled to $[0, 1]$. If the range is larger, scaling is still performed based on the $[0, 180]$ interval.
    \item \textbf{Price}: The original value range $[0, 200]$ is linearly scaled to $[0, 1]$. If the range is larger, scaling is still performed based on the $[0, 200]$ interval.
    \item \textbf{Comfort Level}: The original value range $[0, 5]$ is linearly scaled to $[0, 1]$.
    \item \textbf{Comfort Level and Reliability}: Comfort and reliability (in the range $[0, 1]$) are subjected to an inverse transformation (i.e., $1 - x$) when jointly considered with other factors, and zero values are removed.
\end{enumerate}

In addition, we construct \textit{Mix\_tab} by preserving all station entries in the original \textit{Vertex\_tab} and merging the numerical attributes from \textit{Edge\_tab} into the corresponding station records. For both Metromap and Travelmap, the value of each merged numerical attribute is computed as the mean value of all edges connected to the corresponding station in \textit{Edge\_tab}. Attributes that are originally defined in \textit{Vertex\_tab} and are not involved in the merging process remain unchanged. Therefore, \textit{Mix\_tab} retains the node-based structure of \textit{Vertex\_tab} while incorporating edge-level numerical information through station-wise aggregation.

The prompts adopted for the five experimental configurations are shown in \textbf{Prompt: Criteria-free Query (Map-Only)}, \textbf{Prompt: Criteria-free Query (Edge\_tab-Only)}, \textbf{Prompt: Criteria-free Query (Map+Edge\_tab)}, \textbf{Prompt: Criteria-based Query (Map+Edge\_tab+Vertex\_tab)}, and \textbf{Prompt: Multi-Conditional Query (Map+Mix\_tab)}, respectively.

\clearpage

\begin{figure*}[ht]
\centering
\begin{minipage}{0.95\textwidth}
    \begin{promptbox}[Prompt: Metromap Edge\_tab Generation]
    # Role
    You are a **table-generation agent** that generates metro edge data.
    
    # Rules
    - **Only output a metro Edge table. No explanations, no extra text, no code block markers.**
    - **Generate strictly based on the provided JSON file. Do not hallucinate.**
    - **Do not duplicate edges** (`A-B` and `B-A` count as one).
    - For each **Line**, list all edges consecutively, then move to the next Line.
    - **Loop lines** must form a closed loop; **non-loop lines** only connect adjacent stations.
    
    # Table Format (exact)
    Edge, Line, Time, Price, Comfort Level, Reliability
    
    # Edge & Name Rules
    - Edge format: `StationA-StationB`
    - If a station name contains `-`, replace it with `_` and remove spaces around it
      (e.g., `King Street-Old Town` -> `King Street_Old Town`)
    - Preserve original spaces inside names (e.g., `King Street`)
    - Preserve other symbols but remove spaces around them
      (e.g., `U Street / African` -> `U Street/African`)
    
    # Value constrained
    - Line: `x Line`
    - Time: random, 2 decimals, **2.00-3.00**
    - Price: random, 2 decimals, **0.00-1.50**
    - Comfort Level: random, 2 decimals, **0.00-1.00**
    - Reliability: only **0**
    
    # Example (format only)
    Edge, Line, Time, Price, Comfort Level, Reliability
    Shady Grove-Rockville, Red Line, 2.20, 1.38, 0.45, 0.91
    Fort Totten-Brookland, Line 1, 2.33, 1.32, 0.55, 0.77
    
    Now start generating the Edge CSV. Wrap the CSV content strictly between `<|edge_csv_begin|>` and `<|edge_csv_end|>`, and ensure that the CSV syntax between these two tags is absolutely correct.
    \end{promptbox}
\end{minipage}
\end{figure*}
\newpage

\begin{figure*}[ht]
\centering
\begin{minipage}{0.95\textwidth}
    \begin{promptbox}[Prompt: Metromap Vertex\_tab Generation]
    # Role
    You are a **table-generation agent** that generates metro **Vertex** data.
    
    # Rules
    - **Strictly generate a metro Vertex table only.**
    - **Output table content only**: no explanations, no extra text, no notes, no comments, no code block markers.
    - **Generate strictly based on the given JSON file. Do not hallucinate.**
    - **Remove duplicate vertices**: if a vertex appears multiple times, keep only the first occurrence.
    - For each **Line**, list all vertices **in order**, then move on to the next Line.
    
    # Table Format (exact)
    Vertex, Line, Time, Price, Comfort Level, Reliability, Transfer Time
    
    # Vertex Rules
    - Vertex format: `StationName` (e.g., `Shady Grove`, `Rockville`, `Twinbrook`)
    
    # Special Character Handling Rules
    1. Replace `-` with `_` and remove spaces if the station name originally contains `-`
       (e.g., `King Street-Old Town` -> `King Street_Old Town`)
    2. Preserve original spaces inside names (e.g., `King Street`)
    3. Preserve other symbols but remove spaces around them
       (e.g., `U Street / African` -> `U Street/African`)
    
    # Column Value Constrained
    - Line: `x Line`
    - Time: random, 2 decimals, **0.50-2.00**
    - Price: only **0**
    - Comfort Level: random, 2 decimals, **0.00-1.00**
    - Reliability: random, 2 decimals, **0.50-1.00**
    - Transfer Time: random, integers, **5-15**
    
    # Example (format only)
    Vertex, Line, Time, Price, Comfort Level, Reliability, Transfer Time
    Fort Totten, Line 1|Line 2, 1.38, 0, 0.38, 0.75, 6
    Chow, Red Line|Green Line, 1.55, 0, 0.92, 0.66, 9
    
    Now start generating the Vertex CSV. Wrap the CSV content strictly between `<|vertex_csv_begin|>` and `<|vertex_csv_end|>`, and ensure that the CSV syntax between these two tags is absolutely correct.
    \end{promptbox}
\end{minipage}
\end{figure*}
\newpage

\begin{figure*}[ht]
\centering
\begin{minipage}{0.95\textwidth}
    \begin{promptbox}[Prompt: Travelmap Edge\_tab Generation]
    # Role
    You are a **table-generation agent** that generates **tourist attraction (scenic area) edge data** based on a map.
    
    # Rules
    - **Strictly generate an Edge table only.**
    - **Output table content only**: no explanations, no extra text, no notes, no comments, no code block markers.
    - **Generate strictly based on the given scenic map image. Do not hallucinate.**
    - **Only list valid Edges**: an Edge exists only if two attractions are directly connected on the map.
    - **Do not duplicate Edges**: `A-B` and `B-A` are the same and must appear only once.
    - List Edges **top to bottom first, then left to right**, following the spatial order of the map.
    
    # Table Format (exact)
    Edge, Time, Price, Comfort Level, Reliability
    
    # Edge Rules
    - Edge format: `AttractionA-AttractionB`
    (e.g., `MUJI YOUSELF-Rockefeller Center`, `Rockefeller Center-Nintendo`)
    - Preserve original spaces inside names (e.g., `Rockefeller Center`)
    - Preserve other symbols but remove spaces around them
      (e.g., `St. Patrick' s Cathedral` -> `St.Patrick's Cathedral`)
    
    # Value constrained
    - **Time**: minutes only; if shown on the map convert directly (`24 minutes` -> `24`, `2 hours` -> `120`); if distance is provided and **> 1 km**, compute time using **60 km/h**; if distance is **< 1 km**, compute time using **4 m/s** and convert to minutes; any result **< 1 minute is rounded up to 1 minute**; if neither time nor distance is shown, generate a random integer **10-60**
    - **Price**: random integer **10-80**, with smaller values appearing more frequently
    - **Comfort Level**: travel review style rating, random one-decimal **1.0-5.0**, values **>= 3.0** should be the majority
    - **Reliability**: defined as `a/365`, where `a` is an integer **1-365**, values with **a > 180** should be the majority, output as a decimal with **six digits**
    
    # Example (format only)
    Edge, Time, Price, Comfort Level, Reliability
    MUJI YOUSELF-Rockefeller Center, 24, 20, 2.3, 0.928767
    Rockefeller Center-Nintendo, 18, 15, 2.4, 0.805479
    
    Now start generating the Vertex CSV. Wrap the CSV content strictly between `<|vertex_csv_begin|>` and `<|vertex_csv_end|>`, and ensure that the CSV syntax between these two tags is absolutely correct.
    \end{promptbox}
\end{minipage}
\end{figure*}
\clearpage
\newpage

\begin{figure*}[ht]
\centering
\begin{minipage}{0.95\textwidth}
    \begin{promptbox}[Prompt: Travelmap Vertex\_tab Generation]
    # Role
    You are a **table-generation agent** that generates **tourist attraction (scenic area) vertex data** based on a map.
    
    # Rules
    - **Strictly generate a Vertex table only.**
    - **Output table content only**: no explanations, no extra text, no notes, no comments, no code block markers.
    - **Generate strictly based on the given scenic map image. Do not hallucinate.**
    - **Only list valid Vertices**: a Vertex must appear as a scenic attraction on the map.
    - **Do not duplicate Vertices**: if the same attraction appears multiple times, list it only once.
    - List Vertices **top to bottom first, then left to right**, following the spatial order of the map.
    
    # Table Format (exact)
    Vertex, Time, Price, Comfort Level, Reliability
    
    # Vertex Rules
    - Vertex format: `AttractionName`
      (e.g., `MUJI YOUSELF`, `Rockefeller Center`, `Nintendo`)
    - Preserve original spaces inside names.
      (e.g., `Rockefeller Center`)
    - Preserve other symbols but remove spaces around them
      (e.g., `St. Patrick' s Cathedral` -> `St.Patrick's Cathedral`)
    
    # Value constrained
    - **Time**: random integer **30-180**, set to **0** for places like airports, train stations, etc.
    - **Price**: random **integer multiple of 5** in **10-200**
    - **Comfort Level**: travel review style rating, random value in **1.0-5.0** with **one decimal place**, values **>= 3.0** should be the majority
    - **Reliability**: defined as `a/365`, where `a` is an integer **60-365**, values with **a > 180** should be the majority, output as a decimal with **six digits**
    
    # Example (format only)
    Vertex, Time, Price, Comfort Level, Reliability
    MUJI YOUSELF, 41, 45, 4.9, 0.975342
    Rockefeller Center, 43, 100, 4.6, 0.745205
    Nintendo, 21, 55, 4.3, 0.380822
    
    Now start generating the Vertex CSV. Wrap the CSV content strictly between `<|vertex_csv_begin|>` and `<|vertex_csv_end|>`, and ensure that the CSV syntax between these two tags is absolutely correct.
    \end{promptbox}
\end{minipage}
\end{figure*}
\clearpage
\newpage

\begin{promptbox}[Prompt: Criteria-free Query (Map-Only)]
# Role
You are a Subway Pathfinding Algorithm. You strictly analyze the provided metro map, and compute the shortest-stop route.

# Input Task
{question}

# Objective
Return only one final route - the one with the absolute minimlum number of stops.
A station cannot appear more than once in the route. Cannot reselect the path that has already been taken!
If multiple routes have the same number of stops, randomly select one of them.

# Output Format Constrained
- **Separator**: Use hyphens `-` strictly, with **no spaces** around the hyphens.
- **Transfer Marking**: If and only if a line transfer occurs at a station, append `(transfer)` immediately after its name.
- **Content**: Output only the final route string. No markdown, no explanation, no filler.

# Few-Shot Output Examples
Notice that this is only the output format example. In real question we will give you an image as map for you to figure out the shortest path.

## Example 1
User: Path from Summerhill to Woodbine
Assistant: Summerhill-Rosedale-Bloor_Yonge(transfer)-Sherbourne-Castle Frank-Broadview-Chester-Pape-Donlands-Greenwood-Coxwell-Woodbine

## Example 2
User: Path from Viktoria_Luise_Platz to Jakob_Kaiser_Platz
Assistant: Viktoria_Luise_Platz-Nollendorfplatz(transfer)-Wittenbergplatz(transfer)-Augsburger Strabe-Spichernstrabe(transfer)-Berliner Strabe(transfer)-Blissestrabe-Konstanzer Strabe-Adenauerplatz-Wilmersdorfer Strabe-Richard_Wagner_Platz-Mierendorffplatz-Jungfernheide-Jakob_Kaiser_Platz
\end{promptbox}
\clearpage
\newpage

\begin{promptbox}[Prompt: Criteria-free Query (Edge\_tab-Only)]
# Role
You are a Subway Pathfinding Algorithm. You strictly analyze the tabular (Edge Table), and compute the shortest-stop route.

# Input Task
{question}

# Objective
Return only one final route - the one with the absolute minimlum number of stops.
A station cannot appear more than once in the route. Cannot reselect the path that has already been taken!
If multiple routes have the same number of stops, randomly select one of them.

# Output Format Constrained
- **Separator**: Use hyphens `-` strictly, with **no spaces** around the hyphens.
- **Transfer Marking**: If and only if a line transfer occurs at a station, append `(transfer)` immediately after its name.
- **Content**: Output only the final route string. No markdown, no explanation, no filler.

# Few-Shot Output Examples
Notice that this is only the output format example. In real question we will give you a tabular (Edge Table) for you to figure out the shortest path.

## Example 1
User: Path from Summerhill to Woodbine
Assistant: Summerhill-Rosedale-Bloor_Yonge(transfer)-Sherbourne-Castle Frank-Broadview-Chester-Pape-Donlands-Greenwood-Coxwell-Woodbine

## Example 2
User: Path from Viktoria_Luise_Platz to Jakob_Kaiser_Platz
Assistant: Viktoria_Luise_Platz-Nollendorfplatz(transfer)-Wittenbergplatz(transfer)-Augsburger Strabe-Spichernstrabe(transfer)-Berliner Strabe(transfer)-Blissestrabe-Konstanzer Strabe-Adenauerplatz-Wilmersdorfer Strabe-Richard_Wagner_Platz-Mierendorffplatz-Jungfernheide-Jakob_Kaiser_Platz
\end{promptbox}
\clearpage
\newpage

\begin{promptbox}[Prompt: Criteria-free Query (Map+Edge\_tab)]
# Role
You are a Subway Pathfinding Algorithm. You strictly analyze the provided metro map, a tabular (Edge Table), and compute the shortest-stop route.

# Input Task
{question}

# Objective
Return only one final route - the one with the absolute minimlum number of stops.
A station cannot appear more than once in the route. Cannot reselect the path that has already been taken!
If multiple routes have the same number of stops, randomly select one of them.

# Output Format Constrained
- **Separator**: Use hyphens `-` strictly, with **no spaces** around the hyphens.
- **Transfer Marking**: If and only if a line transfer occurs at a station, append `(transfer)` immediately after its name.
- **Content**: Output only the final route string. No markdown, no explanation, no filler.

# Few-Shot Output Examples
Notice that this is only the output format example. In real question we will give you an image as map and a tabular (Edge Table) for you to figure out the shortest path.

## Example 1
User: Path from Summerhill to Woodbine
Assistant: Summerhill-Rosedale-Bloor_Yonge(transfer)-Sherbourne-Castle Frank-Broadview-Chester-Pape-Donlands-Greenwood-Coxwell-Woodbine

## Example 2
User: Path from Viktoria_Luise_Platz to Jakob_Kaiser_Platz
Assistant: Viktoria_Luise_Platz-Nollendorfplatz(transfer)-Wittenbergplatz(transfer)-Augsburger Strabe-Spichernstrabe(transfer)-Berliner Strabe(transfer)-Blissestrabe-Konstanzer Strabe-Adenauerplatz-Wilmersdorfer Strabe-Richard_Wagner_Platz-Mierendorffplatz-Jungfernheide-Jakob_Kaiser_Platz
\end{promptbox}
\newpage

\begin{promptbox}[Prompt: Criteria-based Query (Map+Edge\_tab+Vertex\_tab)]
# Role
You are a Subway Pathfinding Algorithm. You strictly analyze the provided metro map, two tabulars (an Edge Table and a Vertex Table), and compute the best overall route based on weighted Time, Price, Comfort Level, and Reliability.

# Input Task
{question}

# Objective
The route should be planned for a family trip. Time, Price, Comfort Level, and Reliability must be considered together according to the weight proportions {w1}, {w2}, {w3}, {w4}. Return only one final route - the one you believe is the best overall choice based on these weighted factors.
A station cannot appear more than once in the route. Cannot reselect the path that has already been taken!
**Note:** Total Time includes Commuting Time of each edge, Standing Time of each vertex and Transfer Time from stations. Transfer Time is only added when a line transfer occurs; stations passed without transferring do not contribute Transfer Time.

# Input Data Description
The input consists of: 1. One metro map image
2. An **Edge Table**: each edge contains `Time`, `Price`, `Comfort Level`, and `Reliability` attributes representing travel time, monetary cost, ride comfort, and route reliability
3. A **Vertex Table**: each station contains `Time`, `Price`, `Comfort Level`, `Reliability`, and `Transfer Time`
  - `Time`: time spent at the station
  - `Price`: monetary cost at the station
  - `Comfort Level`: comfort at the station
  - `Reliability`: reliability at the station
  - `Transfer Time`: time penalty added only when a line transfer occurs at that station

# Attributes Scaling Rules
1. All `Time` values, originally within 0-3, must be proportionally scaled into 0-1
2. All `Price` values, originally within 0-1.5, must be proportionally scaled into 0-1
3. `Transfer Time` values, originally 5-15, must be scaled using the same 0-3 proportion, ensuring transferring carries a significant penalty
4. All `Comfort Level` and `Reliability` values, ranging from 0-1, must be converted into one minus their values for scoring, excluding zero values

# Pathfinding Requirements
1. The metro network must be treated as a heterogeneous weighted graph where edges receive weights from the Edge Table and stations receive weights from the Vertex Table
2. Total Time includes Commuting Time of each edge, Standing Time of each vertex, and Transfer Time from stations, applied only when a line transfer occurs
3. All attribute values of the destination station are excluded from evaluation; starting station attributes are fully included
4. When weight proportions {w1}, {w2}, {w3}, {w4} are provided, the final route score is computed by combining normalized Time, normalized Price, converted Comfort Level, and converted Reliability according to these proportions
5. The system must output only the single route with the best overall score

# Output Format Constrained
- **Separator**: Use hyphens `-` strictly, with no spaces around the hyphens
- **Transfer Marking**: If a line transfer occurs at a station, append `(transfer)` immediately after its name
- **Content**: Output only the final route string. No markdown, no explanation, no filler

# Few-Shot Output Examples
Notice that these are only formatting examples. Actual routes depend on map and tables.
\end{promptbox}

\begin{promptbox}[Prompt: Multi-Condiitional Query (Map+Mix\_tab)]
# Role
You are a Subway Pathfinding Algorithm. You strictly analyze the provided metro map and the Vertex Table, and compute the best overall route based on weighted Time, Price, Comfort Level, and Reliability.

# Input Task
{question}

# Objective
The route should be planned for a family trip. Time, Price, Comfort Level, and Reliability must be considered together according to the weight proportions {w1}, {w2}, {w3}, {w4}. Return only one final route - the one you believe is the best overall choice based on these weighted factors.
A station cannot appear more than once in the route. Cannot reselect the path that has already been taken!
**Note:** Total Time includes both Standing Time of each vertex and Transfer Time from stations. Transfer Time is only added when a line transfer occurs; stations passed without transferring do not contribute Transfer Time.

# Input Data Description
The input consists of:
1. One metro map image
2. A **Vertex Table**: each station contains `Time`, `Price`, `Comfort Level`, `Reliability`, and `Transfer Time`
- `Time`: time spent at the station
- `Price`: monetary cost at the station
- `Comfort Level`: comfort at the station
- `Reliability`: reliability at the station
- `Transfer Time`: time penalty added only when a line transfer occurs at that station

# Attributes Scaling Rules
1. All `Time` values, originally within 0-3, must be proportionally scaled into 0-1
2. All `Price` values, originally within 0-1.5, must be proportionally scaled into 0-1
3. `Transfer Time` values, originally 5-15, must be scaled using the same 0-3 proportion, ensuring transferring carries a significant penalty
4. All `Comfort Level` and `Reliability` values, ranging from 0-1, must be converted into one minus their values for scoring, excluding zero values

# Pathfinding Requirements
1. The metro network must be treated as a heterogeneous weighted graph where stations receive weights from the Vertex Table
2. Total Time includes includes both Standing Time of each vertex and Transfer Time from stations, applied only when a line transfer occurs
3. All attribute values of the destination station are excluded from evaluation; starting station attributes are fully included
4. When weight proportions {w1}, {w2}, {w3}, {w4} are provided, the final route score is computed by combining normalized Time, normalized Price, converted Comfort Level, and converted Reliability according to these proportions
5. The system must output only the single route with the best overall score

# Output Format Constrained
- **Separator**: Use hyphens `-` strictly, with no spaces around the hyphens
- **Transfer Marking**: If a line transfer occurs at a station, append `(transfer)` immediately after its name
- **Content**: Output only the final route string. No markdown, no explanation, no filler

# Few-Shot Output Examples
Notice that these are only formatting examples. Actual routes depend on map and tables.
\end{promptbox}
\newpage

\subsection{Prompts for QA Queries}\label{subsec:prompt_for_qa}
For the QA tasks, a total of 24 questions are designed across the Metromap and Travelmap scenarios. In Metromap, four input modalities: \textit{Map-Only, Edge\_Tab-Only, Vertex\_Tab-Only, and Map + Mix\_Tab} are considered, covering three categories of problems: Global Perception-based Reasoning tasks (GP), Local Perception-based Reasoning tasks (LP), and Spatial Relationship Judgment tasks (SR). This paper reports only the prompts corresponding to the Map + Mix\_Tab input setting.

\begin{promptbox}[Prompt: Map+Mix\_Tab-GP]
    # Role
    You are an algorithm that determines **the total stop time of the subway line with the longest total stop time**.
    You must strictly analyze the provided **map image and the Vertex Table** and compute **which subway line has the maximum total stop time and what that total time is**.
    
    # Input Task
    {question}
    
    # Requirements
    ## For the map
    - **Do NOT use any prior knowledge, memory, or real-world assumptions.**
      You must rely **only on what is visually present in the given map**.
    - The Subway Map provides **all station information for each line**, which must be used to sum the stop times of each line.
    - **Branches are counted as separate lines** and must **NOT** be merged into the main line.
    - If a line **changes its name**, and the new name is **completely different** from the previous one,
      then **each distinct name is treated as a separate line**.
      (This is different from branches.)
    - If the **same line is visually disconnected**, it is still considered **one single line**.
    - If a line **splits into multiple paths**, **all stations on all split paths** are included in the total stop time of that line.
    - **Only lines whose stations have valid `time` values** should be counted.
      Lines with **no `time` values at all** must be **excluded entirely**.
    - Lines **without a name** should **NOT** be counted.
    - Lines **without stations** should **NOT** be counted.
    - Be careful **not to misidentify rivers, roads, or other non-transit visual elements** as subway lines.
    
    ## For the Vertex Table
    - Each station's **stop time** must be obtained **only from the `time` column of the Vertex Table**.
    
    # Output Format Constrained
    - Return **one numeric value only**, **rounded to two decimal places**.
    - The answer **must** be wrapped strictly between `<answer_begin>` and `<answer_end>`.
    - Output **only** the final answer delimited by `<answer_begin>` and `<answer_end>`. No markdown, no explanation, no extra text.

    # Few-Shot Output Examples
    > Note: These are **only format examples**.
    > In the real task, you will be given an **image map and a Vertex Table**.

    ## Example 1
    User: What is the total stop time of the subway line with the longest stop time?
    Assistant: <answer_begin>30.55<answer_end>

    ## Example 2
    User: How long is the total stop time for the subway line with the longest stops?
    Assistant: <answer_begin>77.89<answer_end>
\end{promptbox}
\newpage

\begin{promptbox}[Prompt: Map+Mix\_Tab-LP]
# Role
You are an algorithm that determines **the average reliability from station A to station B along the same subway line**.
You must strictly analyze the provided **map image and the Vertex Table** and compute **the average reliability of all stations between A and B on that line**.

# Input Task
{question}

# Requirements
## For the map
- **Do NOT use any prior knowledge, memory, or real-world assumptions.**
  You must rely **only on what is visually present in the given map**.
- Collect **all stations on that line between station A and station B**, **including station A and station B themselves**.
- Be careful **not to misidentify rivers, roads, or other non-transit visual elements** as subway lines.

## For the Vertex Table
- Each station's **reliability value** must be obtained **only from the `reliability` column of the Vertex Table**.

# Output Format Constrained
- Return **one numeric value only**, **rounded to two decimal places**.
- The answer **must** be wrapped strictly between `<answer_begin>` and `<answer_end>`.
- Output **only** the final answer delimited by `<answer_begin>` and `<answer_end>`. No markdown, no explanation, no extra text.

# Few-Shot Output Examples
> Note: These are **only format examples**.
> In the real task, you will be given an **image map and a Vertex Table**.

## Example 1
User: What is the average reliability from station A to station B on the same line?
Assistant: <answer_begin>30.55<answer_end>

## Example 2
User: Could you provide the average reliability for the route between station A and station B along the same line?
Assistant: <answer_begin>77.89<answer_end>
\end{promptbox}
\newpage

\begin{promptbox}[Prompt: Map+Mix\_Tab-SR]
# Role
You are an algorithm that determines **whether there are transfer stations between Line X and Line Y**.
If transfer stations exist, return **the shortest transfer time**; if not, return **0**.
You must strictly analyze the provided **map image and the Vertex Table** to make this determination.

# Input Task
{question}

# Requirements
## For the map
- **Do NOT use any prior knowledge, memory, or real-world assumptions.**
  You must rely **only on what is visually present in the given map**.
- To determine whether **Line X and Line Y intersect**, inspect the map to see **whether they share any common station(s)**.
- Identify **all intersection (transfer) stations** between Line X and Line Y shown on the map.
- If **no intersection stations exist**, return **0** directly.
- Be careful **not to misidentify rivers, roads, spots from other lines, or any unrelated visual elements** as subway lines or stations.

## For the Vertex Table
- For each identified transfer station, obtain its **transfer time** from the **`transfer time` column** of the Vertex Table.
- Compare the transfer times of **all transfer stations** and select **the minimum value**.

# Output Format Constrained
- Return **a positive integer or 0 only**.
  - A **positive integer** means **the minimum transfer time among all transfer stations between Line X and Line Y**.
  - `0` means **Line X and Line Y do not intersect**.
- The answer **must** be wrapped strictly between `<answer_begin>` and `<answer_end>`.
- Output **only** the final answer delimited by `<answer_begin>` and `<answer_end>`. No markdown, no explanation, no extra text.

# Few-Shot Output Examples
> Note: These are **only format examples**.
> In the real task, you will be given an **image map and a Vertex Table**.

## Example 1
User: Please check if there are transfer stations between Line X and Line Y. If there are, return the shortest transfer time; if not, return 0.
Assistant: <answer_begin>0<answer_end>

## Example 2
User: Can you see if there are any transfer stations between Line X and Line Y? If so, let me know the shortest transfer time. If not, just return 0.
Assistant: <answer_begin>6<answer_end>
\end{promptbox}
\newpage

\section{Additional Experimental Analysis}\label{sec:add_exp_analysis}
\subsection{Experimental Settings}\label{subsec:experiment_settings}
\subsubsection{Model Selection Rules}
Model selection is guided by evaluation requirements centered on the RP task, and two filtering rules are applied accordingly:
\begin{enumerate}
    \item Models that cannot meet long-context requirements are excluded (e.g., DeepSeek-VL2~\cite{wu2024deepseek}), as the input length of some samples exceeds 32k tokens.
    \item Models that support long-context inputs but tend to exhibit reasoning loops or excessive deliberation in complex RP tasks, resulting in unstable or poorly structured outputs, are also excluded, such as llava-v1.6-mistral-hf (7B)~\cite{liu2023improved}, Ovis2.5 (9B)~\cite{lu2025ovis2}, and Glyph (10B)~\cite{cheng2025glyph}.
\end{enumerate}

\subsubsection{Evaluated Models.}
We evaluate a diverse set of state-of-the-art MLLMs, including both general instruction-following models and reasoning-enhanced models. The evaluated models cover the Qwen3-VL series (2B, 8B, 30B-A3B, and 32B)~\cite{bai2025qwen3vltechnicalreport}, Qwen2.5-VL-7B~\cite{bai2025qwen2}, Qwen3.5-9B~\cite{qwen2026qwen35}, Qwen3.6-35B-A3B~\cite{qwen2026github}, Phi-3.5-Vision-4B~\cite{abdin2024phi3technicalreporthighly}, Phi-4-Multimodal-6B~\cite{abdin2024phi}, InternVL3-8B~\cite{zhu2025internvl3}, GPT-4o~\cite{hurst2024gpt}, GPT-4.1~\cite{openai2025gpt41}, GPT-5.5-Instant~\cite{openai2026gpt55instantdoc}, Gemini-3-Flash-Preview~\cite{google2025gemini3flash}, Gemini-3.5-Flash~\cite{google2026gemini35flashdoc}, Doubao-Seed-1.6~\cite{doubao2025seed16}, and Qwen3-VL-Plus-Thinking~\cite{bai2025qwen3vltechnicalreport}, together with their corresponding reasoning-enabled variants where available. All models are evaluated under a unified protocol with consistent prompts, inference settings, and multimodal input preprocessing. 

\subsubsection{Model Parameter Settings}
Open-source models are primarily evaluated using the VLLM inference framework on a cluster equipped with four NVIDIA A100 GPUs. To ensure reproducibility, all experiments uniformly set \texttt{temperature} to 0.0, \texttt{maximum\_generation\_length} to 2{,}048 tokens, and extend \texttt{max\_model\_len} to the maximum supported by each model. Some open-source models are accessed via third-party APIs, with other parameters kept at official default settings; closed-source models are evaluated through their official APIs, following the default configurations specified in the corresponding technical documentation. For models equipped with a Thinking mechanism, the \texttt{thinking\_budget} is uniformly set to 4{,}096 tokens.

\subsubsection{Input Preprocessing}
Tabular data are primarily provided in CSV format, with a small number of samples using JSON format. Visual inputs adopt an adaptive preprocessing strategy: for models that support arbitrary resolutions, original images are preserved; when image dimensions exceed model limits, images are resized to the maximum allowable size while maintaining the original aspect ratio; for models that enforce a fixed aspect ratio (e.g., square inputs), non-uniform stretching is applied. Although this operation alters the geometric proportions of the images, it preserves key topological structures and textual forms, thereby ensuring the readability of semantic information.

\subsection{Additional Route Planning Results} \label{subsec:additional_rp_results}
\begin{table*}[!ht]

  \caption{Evaluation results of various Multimodal Large Language Models (MLLMs) on the MapTab route planning task under the \textbf{Metromap} and \textbf{Travelmap} scenarios. EMA, PMA, and DS denote Exact Match Accuracy, Partial Match Accuracy, and Difficulty-aware Score, respectively. Map-only: map information only; Edge\_tab-only: edge data only; Map+Edge\_tab: map and edge data; Map+Edge\_tab+Vertex\_tab: map, edge, and vertex data; Map+Mix\_tab: map and Mix\_tab. We did not include Edge\_tab+Vertex\_tab because the comparison between it and Map+Mix\_tab yielded conclusions consistent with those from the Map-only and Edge\_tab-only control groups, without new findings. Bold values represent the best performance within open-source and closed-source groups, respectively.}

  \label{combined-eval-table}
  \renewcommand{\arraystretch}{0.95}
  \vspace{-0.5em}

  \begin{center}
    \begin{small}
      \resizebox{0.99\textwidth}{!}{
      \begin{tabular}{ll ccc ccc ccc ccc ccc}
        \toprule

        \textbf{Model} & \textbf{Type}
        & \multicolumn{3}{c}{\textbf{Map-only}}
        & \multicolumn{3}{c}{\textbf{E\_tab-only}}
        & \multicolumn{3}{c}{\textbf{Map+E\_tab}}
        & \multicolumn{3}{c}{\textbf{Map+E\_tab+V\_tab}}
        & \multicolumn{3}{c}{\textbf{Map+Mix\_tab}} \\

        \cmidrule(r){3-5}
        \cmidrule(lr){6-8}
        \cmidrule(lr){9-11}
        \cmidrule(lr){12-14}
        \cmidrule(l){15-17}

        & & \textbf{EMA} & \textbf{PMA} & \textbf{DS}
        & \textbf{EMA} & \textbf{PMA} & \textbf{DS}
        & \textbf{EMA} & \textbf{PMA} & \textbf{DS}
        & \textbf{EMA} & \textbf{PMA} & \textbf{DS}
        & \textbf{EMA} & \textbf{PMA} & \textbf{DS} \\

        \midrule

        \multicolumn{17}{c}{\textit{\textbf{Scenario: Metromap}}} \\

        \midrule

        \multicolumn{17}{l}{\textit{\textbf{Open-source Models}}} \\

        Qwen3-VL-8B-Instruct & No-Thinking & 2.75 & 17.58 & 103 & 25.69 & 46.44 & 1153 & 21.25 & 41.30 & 921 & 19.31 & 39.31 & 855 & 4.69 & 21.87 & 182 \\
        Qwen3-VL-8B-Thinking & Thinking & 5.12 & 20.99 & 188 & 31.69 & 49.76 & 1427 & 38.00 & 57.06 & 1771 & 23.75 & 41.69 & 1080 & 6.38 & 22.93 & 270 \\
        Qwen3-VL-2B-Instruct & No-Thinking & 0.94 & 15.14 & 35 & 9.88 & 27.61 & 437 & 6.63 & 23.85 & 282 & 7.00 & 26.91 & 325 & 2.00 & 17.82 & 78 \\
        Qwen2.5-VL-7B-Instruct & No-Thinking & 0.94 & 15.02 & 32 & 14.00 & 31.20 & 621 & 11.69 & 28.32 & 508 & 7.94 & 20.77 & 357 & 3.38 & 18.09 & 131 \\
        Phi-3.5-Vision-Instruct-4B & No-Thinking & 0.06 & 10.40 & 2 & 10.87 & 27.92 & 476 & 6.63 & 22.14 & 272 & 2.75 & 12.27 & 117 & 0.81 & 12.94 & 26 \\
        Phi-4-Multimodal-Instruct-6B & No-Thinking & 0.00 & 9.75 & 0 & 2.13 & 12.52 & 84 & 2.13 & 11.78 & 91 & 1.75 & 9.51 & 70 & 0.44 & 9.02 & 14 \\
        InternVL3-8B-Instruct & No-Thinking & 0.13 & 13.98 & 4 & 10.50 & 29.57 & 460 & 12.81 & 31.83 & 555 & 9.00 & 24.73 & 413 & 1.75 & 17.00 & 76 \\
        Qwen3-VL-30B-A3B-Instruct & No-Thinking & 3.31 & 19.26 & 129 & 23.69 & 44.33 & 1062 & 22.56 & 43.58 & 1017 & 19.00 & 40.03 & 842 & 6.75 & 26.22 & 288 \\
        Qwen3-VL-32B-Instruct & No-Thinking & 6.31 & 22.23 & 250 & 31.87 & 54.45 & 1430 & 32.12 & 54.54 & 1463 & 28.50 & 50.06 & 1303 & 6.56 & 24.43 & 262 \\
        Qwen3-VL-32B-Thinking & Thinking & \textbf{13.31} & \textbf{29.43} & 558 & 31.81 & 54.94 & 1451 & 44.12 & 62.77 & 2148 & 26.56 & 51.48 & 1211 & 9.19 & 28.89 & 381 \\
        Qwen3.5-9B & No-Thinking & 5.69 & 22.44 & 231 & 25.56 & 48.95 & 1174 & 26.50 & 49.33 & 1189 & 20.25 & 44.15 & 903 & 8.75 & 28.15 & 365 \\
        Qwen3.6-35B-A3B & Thinking & 11.56 & 29.16 & 476 & \textbf{64.88} & \textbf{77.06} & 3510 & \textbf{53.87} & \textbf{71.43} & 2811 & \textbf{43.69} & \textbf{64.16} & 2275 & \textbf{14.75} & \textbf{34.76} & 643 \\

        \midrule

        \multicolumn{17}{l}{\textit{\textbf{Closed-source Models}}} \\

        GPT4-o & No-Thinking & 6.63 & 25.61 & 257 & 42.38 & 64.07 & 2098 & 40.69 & 62.40 & 1969 & 35.63 & 55.51 & 1702 & 11.31 & 31.11 & 469 \\
        GPT4.1 & No-Thinking & 7.94 & 25.52 & 306 & 48.56 & 67.07 & 2446 & 46.81 & 65.18 & 2344 & 41.81 & 62.88 & 2059 & 14.06 & 35.98 & 608 \\
        GPT-5.5-Instant & No-Thinking & 43.63 & 64.27 & 2320 & \textbf{88.75} & \textbf{96.24} & 5441 & \textbf{85.31} & \textbf{93.89} & 5244 & \textbf{88.88} & \textbf{92.15} & 5467 & \textbf{69.50} & \textbf{79.76} & 4021 \\
        Doubao-Seed-1.6-w/o-Thinking & No-Thinking & 8.13 & 24.60 & 315 & 46.94 & 66.98 & 2351 & 48.06 & 66.95 & 2434 & 40.56 & 62.11 & 2041 & 13.81 & 35.61 & 579 \\
        Doubao-Seed-1.6-Thinking & Thinking & 12.06 & 30.49 & 512 & 74.38 & 86.23 & 4284 & 74.00 & 85.68 & 4256 & 76.06 & 83.41 & 4356 & 22.03 & 42.48 & 1029 \\
        Qwen-VL-Plus-w/oThinking & No-Thinking & 4.81 & 21.83 & 186 & 36.88 & 58.69 & 1725 & 38.25 & 58.59 & 1804 & 31.62 & 52.92 & 1479 & 6.94 & 27.69 & 288 \\
        Qwen-VL-Plus-Thinking & Thinking & 10.75 & 29.11 & 437 & 61.50 & 76.62 & 3276 & 62.19 & 76.42 & 3331 & 45.75 & 64.46 & 2290 & 16.38 & 37.44 & 714 \\
        Gemini-3-Flash-Preview & No-Thinking & 37.06 & 57.15 & 1881 & 74.75 & 84.99 & 4457 & 73.06 & 83.37 & 4316 & 69.19 & 76.14 & 4060 & 53.87 & 65.84 & 2976 \\
        Gemini-3.5-flash & No-Thinking & \textbf{44.19} & \textbf{65.40} & 2435 & 83.87 & 93.53 & 5125 & 83.75 & 93.04 & 5099 & 82.37 & 85.34 & 4993 & 60.50 & 71.67 & 3433 \\
        
        \specialrule{\heavyrulewidth}{\aboverulesep}{\belowrulesep}

        \multicolumn{17}{c}{\textit{\textbf{Scenario: Travelmap}}} \\
        
        \midrule
        
        \multicolumn{17}{l}{\textit{\textbf{Open-source Models}}} \\
        
        Qwen3-VL-8B-Instruct & No-Thinking & 19.29 & 42.50 & 1040 & 44.05 & 61.66 & 2597 & 43.33 & 61.39 & 2540 & 34.52 & 55.56 & 2002 & 15.65 & 40.97 & 804 \\
        Qwen3-VL-8B-Thinking & Thinking & 22.62 & 45.94 & 1203 & 74.17 & 82.41 & 4511 & 82.68 & 88.54 & 5199 & 33.15 & 55.60 & 1857 & 12.74 & 38.10 & 653 \\
        Qwen3-VL-2B-Instruct & No-Thinking & 8.45 & 34.30 & 450 & 11.25 & 32.35 & 644 & 19.17 & 45.68 & 1066 & 12.14 & 40.47 & 689 & 3.15 & 30.69 & 159 \\
        Qwen2.5-VL-7B-Instruct & No-Thinking & 7.68 & 30.48 & 390 & 21.07 & 38.15 & 1160 & 24.82 & 42.02 & 1337 & 15.60 & 37.47 & 818 & 4.70 & 28.84 & 228 \\
        Phi-3.5-Vision-Instruct-4B & No-Thinking & 0.12 & 20.00 & 6 & 12.20 & 34.81 & 681 & 9.82 & 31.87 & 543 & 4.46 & 23.21 & 234 & 1.31 & 22.68 & 67 \\
        Phi-4-Multimodal-Instruct-6B & No-Thinking & 0.42 & 19.26 & 20 & 7.20 & 17.63 & 410 & 5.30 & 15.93 & 283 & 1.73 & 9.36 & 96 & 1.43 & 18.96 & 67 \\
        InternVL3-8B-Instruct & No-Thinking & 6.61 & 29.21 & 308 & 29.58 & 49.69 & 1636 & 29.40 & 50.16 & 1647 & 13.57 & 36.78 & 799 & 2.50 & 24.28 & 132 \\
        Qwen3-VL-30B-A3B-Instruct & No-Thinking & 17.86 & 44.15 & 976 & 50.95 & 65.36 & 2978 & 53.75 & 67.71 & 3186 & 38.45 & 58.02 & 2318 & 9.70 & 37.93 & 517 \\
        Qwen3-VL-32B-Instruct & No-Thinking & 36.90 & 57.44 & 2091 & 64.52 & 76.16 & 3944 & 68.39 & 78.99 & 4269 & 52.56 & 69.18 & 3169 & 21.67 & 47.34 & 1161 \\
        Qwen3-VL-32B-Thinking & Thinking & 39.17 & 58.84 & 2262 & 69.76 & 79.60 & 4307 & 91.79 & 94.55 & 5949 & 42.32 & 62.99 & 2492 & 19.94 & 46.73 & 1082 \\
        Qwen3.5-9B & No-Thinking & 25.95 & 49.39 & 1424 & 40.83 & 54.74 & 2487 & 50.06 & 66.84 & 2939 & 33.93 & 57.12 & 1923 & 16.49 & 42.32 & 892 \\
        Qwen3.6-35B-A3B & Thinking & \textbf{42.74} & \textbf{61.74} & 2461 & \textbf{95.42} & \textbf{96.93} & 6289 & \textbf{92.26} & \textbf{94.68} & 6017 & \textbf{62.02} & \textbf{74.68} & 3825 & \textbf{27.08} & \textbf{49.28} & 1532 \\
        
        \midrule
        
        \multicolumn{17}{l}{\textit{\textbf{Closed-source Models}}} \\
        
        GPT4-o & No-Thinking & 16.85 & 40.98 & 930 & 65.06 & 75.84 & 4651 & 62.74 & 74.11 & 4467 & 46.07 & 63.07 & 3069 & 12.08 & 38.07 & 675 \\
        GPT4.1 & No-Thinking & 20.30 & 43.24 & 1077 & 74.82 & 82.98 & 4650 & 70.89 & 79.84 & 4364 & 54.70 & 69.59 & 3310 & 15.06 & 40.67 & 793 \\
        GPT-5.5-Instant & No-Thinking & 68.15 & 78.51 & 4242 & 98.51 & 99.06 & 6593 & 98.04 & 98.55 & 6546 & 83.99 & 87.04 & 5510 & 57.98 & 69.87 & 3535 \\
        Doubao-Seed-1.6-w/o-Thinking & No-Thinking & 33.04 & 54.15 & 1880 & 73.51 & 82.16 & 4552 & 76.85 & 84.04 & 4819 & 56.25 & 71.46 & 3390 & 25.48 & 49.54 & 1406 \\
        Doubao-Seed-1.6-Thinking & Thinking & 38.45 & 58.46 & 2295 & 98.39 & 98.87 & 6567 & 97.86 & 98.47 & 6527 & 83.15 & 89.08 & 5423 & 25.30 & 48.90 & 1437 \\
        Qwen-VL-Plus-w/o-Thinking & No-Thinking & 30.60 & 52.64 & 1691 & 64.23 & 76.45 & 3921 & 69.64 & 79.78 & 4289 & 53.99 & 70.07 & 3243 & 22.92 & 47.65 & 1262 \\
        Qwen-VL-Plus-Thinking & Thinking & 38.27 & 58.94 & 2194 & 64.35 & 76.53 & 3929 & 94.23 & 96.04 & 6159 & 56.19 & 70.84 & 3408 & 23.21 & 47.18 & 1289 \\
        Gemini-3-Flash-Preview & No-Thinking & 60.00 & 73.20 & 3693 & 98.27 & 98.38 & 6567 & 94.40 & 94.87 & 6253 & 78.51 & 82.40 & 5194 & 43.51 & 60.11 & 2687 \\
        Gemini-3.5-flash & No-Thinking & \textbf{68.21} & \textbf{79.12} & 4186 & \textbf{99.52} & \textbf{99.69} & 6674 & \textbf{98.57} & \textbf{98.95} & 6599 & \textbf{92.50} & \textbf{94.28} & 6177 & \textbf{58.93} & \textbf{71.30} & 3648 \\
        
        \bottomrule
        \end{tabular}
        }
    \end{small}
\end{center}
\end{table*}

\begin{table*}[!ht]

  \caption{Performance of QA tasks across multiple MLLMs in the Metromap and Travelmap scenarios. In the Metromap scenario, the Mix\_tab paired with the Map input has the Line column removed to prevent excessive table information from affecting the evaluation of map-table coordination. Task types are categorized into three classes: Global Perception-based Reasoning Tasks (GP), Local Perception-based Reasoning Tasks (LP), and Spatial Relationship Judgment Tasks (SR). Bold values in the table indicate the best performance among open-source and closed-source models, respectively.}
  \label{tab:qa-performance}

  \begin{center}
    \begin{small}
      \resizebox{\textwidth}{!}{
        \begin{tabular}{ll ccc ccc ccc ccc}
          \toprule
          \multirow{2}{*}{Model} & \multirow{2}{*}{Type} & \multicolumn{3}{c}{Map} & \multicolumn{3}{c}{Edge\_tab} & \multicolumn{3}{c}{Vertex\_tab} & \multicolumn{3}{c}{Map+Mix\_tab} \\
          \cmidrule(r){3-5} \cmidrule(lr){6-8} \cmidrule(lr){9-11} \cmidrule(l){12-14}
          & & GP & LP & SR & GP & LP & SR & GP & LP & SR & GP & LP & SR \\
          \midrule

          \multicolumn{14}{c}{\textbf{Scenario: Metromap}} \\
          \midrule
          \multicolumn{14}{l}{\textit{Open-source Models}} \\

          Qwen3-VL-8B-Instruct & No-Thinking & 55.00 & 17.50 & 73.12 & 22.50 & \textbf{100.0} & 7.50 & 57.50 & 51.88 & 86.88 & 0.63 & 22.50 & 38.75 \\
          Qwen3-VL-8B-Thinking & Thinking & 53.12 & 28.12 & 51.25 & 56.87 & 99.38 & 56.87 & 79.37 & 77.50 & 98.12 & 7.50 & 9.38 & 35.63 \\
          Qwen3-VL-2B-Instruct & No-Thinking & 8.13 & 5.00 & 63.12 & 3.75 & 87.50 & 3.12 & 26.25 & 11.25 & 64.38 & 0.00 & 8.13 & 26.87 \\
          Qwen2.5-VL-7B-Instruct & No-Thinking & 48.75 & 15.62 & 66.25 & 15.62 & \textbf{100.0} & 10.00 & 44.37 & 60.62 & 87.50 & 1.25 & 25.62 & 33.12 \\
          Phi-3.5-Vision-Instruct-4B & No-Thinking & 58.13 & 18.75 & 78.12 & 22.50 & \textbf{100.0} & 21.88 & 53.75 & 66.87 & 98.12 & 0.63 & 21.25 & 40.00 \\
          Phi-4-Multimodal-Instruct-6B & No-Thinking & \textbf{60.62} & 40.62 & 80.00 & 41.25 & 93.13 & 42.50 & 58.75 & 92.50 & 99.38 & 5.63 & 35.63 & 49.38 \\
          InternVL3-8B-Instruct & No-Thinking & 35.00 & 20.62 & 65.00 & 7.50 & 83.13 & 3.12 & 33.75 & 10.00 & 70.63 & 0.00 & 12.50 & 28.12 \\
          Qwen3-VL-30B-A3B-Instruct & No-Thinking & 20.62 & 16.25 & 60.62 & 0.63 & 68.75 & 1.25 & 0.00 & 1.25 & 68.75 & 0.00 & 11.25 & 18.12 \\
          Qwen3-VL-32B-Instruct & No-Thinking & 5.00 & 0.00 & 38.12 & 10.62 & 78.12 & 3.12 & 16.25 & 4.37 & 73.12 & 0.00 & 7.50 & 23.75 \\
          Qwen3-VL-32B-Thinking & Thinking & 26.87 & 19.38 & 50.00 & 5.00 & 99.38 & 5.00 & 21.25 & 25.00 & 85.62 & 0.00 & 6.88 & 60.62 \\
          Qwen3.5-9B & No-Thinking & 59.38 & 17.50 & 76.25 & 5.63 & 99.38 & 8.75 & 57.50 & 61.25 & 76.25 & 1.25 & 20.00 & 75.62 \\
          Qwen3.6-35B-A3B & Thinking & \textbf{60.62} & \textbf{42.50} & \textbf{82.50} & \textbf{71.88} & 98.75 & \textbf{83.13} & \textbf{85.62} & \textbf{98.12} & \textbf{100.0} & \textbf{15.00} & \textbf{45.62} & \textbf{95.63} \\

          \midrule
          \multicolumn{14}{l}{\textit{Closed-source Models}} \\

          GPT4-o & No-Thinking & 62.50 & 13.75 & 78.75 & 31.87 & \textbf{100.0} & 28.12 & 55.63 & 78.12 & \textbf{100.0} & 3.75 & 29.38 & 45.00 \\
          GPT4.1 & No-Thinking & 61.88 & 26.87 & 76.25 & 50.62 & 99.38 & 38.12 & 64.38 & 83.75 & \textbf{100.0} & 3.12 & 25.62 & 49.38 \\
          GPT-5.5-Instant & No-Thinking & 58.75 & 83.13 & 91.25 & 99.38 & \textbf{100.0} & 96.25 & \textbf{100.0} & \textbf{100.0} & \textbf{100.0} & 68.75 & 81.87 & \textbf{96.88} \\
          Doubao-Seed-1-6-251015-w/o\_Thinking & No-Thinking & 55.63 & 20.62 & 76.25 & 41.88 & \textbf{100.0} & 59.38 & 58.13 & 86.25 & 99.38 & 3.75 & 49.38 & 50.62 \\
          Doubao-Seed-1-6-251015-Thinking & Thinking & 54.37 & 40.62 & 77.50 & 72.50 & \textbf{100.0} & 69.37 & 96.25 & 98.75 & \textbf{100.0} & 27.50 & 50.00 & 53.12 \\
          Qwen-VL-Plus-w/o\_Thinking & No-Thinking & 60.00 & 21.88 & 78.75 & 40.62 & \textbf{100.0} & 40.00 & 64.38 & 77.50 & 97.50 & 1.25 & 25.62 & 40.62 \\
          Qwen-VL-Plus-Thinking & Thinking & 57.50 & 45.00 & 81.87 & 68.75 & \textbf{100.0} & 71.25 & 90.62 & 95.63 & \textbf{100.0} & 13.75 & 46.25 & 55.63 \\
          Gemini-3-Flash-Preview & No-Thinking & 59.38 & 82.50 & \textbf{93.13} & 91.25 & 98.12 & 75.62 & 88.75 & 94.37 & \textbf{100.0} & 48.13 & 80.00 & 94.37 \\
          Gemini-3.5-flash & No-Thinking & \textbf{63.75} & \textbf{86.25} & 88.75 & \textbf{100.0} & 99.38 & \textbf{98.75} & 98.12 & 99.38 & 99.38 & \textbf{88.75} & \textbf{82.50} & \textbf{96.88} \\

          \midrule

          \multicolumn{14}{c}{\textbf{Scenario: Travelmap}} \\
          \midrule
          \multicolumn{14}{l}{\textit{Open-source Models}} \\

          Qwen3-VL-8B-Instruct & No-Thinking & 7.14 & 60.12 & 52.98 & 17.86 & 99.40 & 45.24 & 38.69 & 50.60 & 61.31 & 75.60 & 70.24 & 14.29 \\
          Qwen3-VL-8B-Thinking & Thinking & 39.29 & 70.83 & 52.38 & 87.50 & \textbf{100.0} & 39.88 & \textbf{100.0} & \textbf{100.0} & \textbf{100.0} & 63.10 & 69.05 & 13.69 \\
          Qwen3-VL-2B-Instruct & No-Thinking & 12.50 & 58.93 & 9.52 & 6.00 & 94.64 & 64.88 & 1.19 & 46.43 & 38.10 & 64.29 & 67.86 & 4.17 \\
          Qwen2.5-VL-7B-Instruct & No-Thinking & 4.76 & 65.48 & 54.17 & 38.10 & 99.40 & 47.62 & 17.26 & 87.50 & 79.76 & 33.93 & 68.45 & 16.07 \\
          Phi-3.5-Vision-Instruct-4B & No-Thinking & 13.10 & 48.21 & 59.52 & 39.88 & 99.40 & 41.07 & 50.00 & 75.60 & 77.38 & 72.02 & 74.40 & 10.71 \\
          Phi-4-Multimodal-Instruct-6B & No-Thinking & \textbf{44.05} & 70.24 & 58.93 & 78.57 & 98.81 & 33.93 & 96.43 & 98.21 & \textbf{100.0} & 73.21 & 69.05 & 17.86 \\
          InternVL3-8B-Instruct & No-Thinking & 12.50 & 59.52 & 35.71 & 8.93 & 97.62 & 49.40 & 10.71 & 50.00 & 51.79 & 70.83 & 67.86 & 4.76 \\
          Qwen3-VL-30B-A3B-Instruct & No-Thinking & 5.95 & 50.60 & 6.55 & 7.74 & 86.31 & 52.38 & 8.93 & 44.64 & 45.24 & 54.76 & 35.12 & 2.98 \\
          Qwen3-VL-32B-Instruct & No-Thinking & 0.00 & 42.26 & 14.88 & 11.31 & 63.69 & 38.31 & 18.45 & 45.83 & 47.62 & 27.98 & 63.69 & 5.95 \\
          Qwen3-VL-32B-Thinking & Thinking & 8.33 & 60.12 & 23.21 & 1.19 & 97.02 & 44.64 & 10.12 & 38.69 & 56.55 & 69.05 & 66.67 & 5.95 \\
          Qwen3.5-9B & No-Thinking & 25.00 & 66.07 & 54.76 & 8.33 & 99.40 & 69.64 & 20.24 & 61.31 & 75.60 & \textbf{82.74} & 71.43 & 21.43 \\
          Qwen3.6-35B-A3B & Thinking & 23.21 & \textbf{75.00} & \textbf{64.29} & \textbf{97.62} & \textbf{100.0} & \textbf{77.38} & \textbf{100.0} & \textbf{100.0} & \textbf{100.0} & 79.76 & \textbf{76.19} & \textbf{25.00} \\

          \midrule
          \multicolumn{14}{l}{\textit{Closed-source Models}} \\

          GPT4-o & No-Thinking & 11.31 & 63.69 & 49.40 & 47.02 & \textbf{100.0} & 36.31 & 53.57 & 99.40 & 67.26 & 71.43 & 73.21 & 11.31 \\
          GPT4.1 & No-Thinking & 3.57 & 69.64 & 55.95 & 47.02 & \textbf{100.0} & 42.86 & 66.07 & \textbf{100.0} & 69.64 & 73.21 & 77.38 & 17.86 \\
          GPT-5.5-Instant & No-Thinking & 44.05 & \textbf{91.67} & \textbf{82.14} & \textbf{100.0} & 99.40 & \textbf{98.81} & \textbf{100.0} & \textbf{100.0} & \textbf{100.0} & 80.36 & \textbf{97.62} & \textbf{32.74} \\
          Doubao-Seed-1-6-251015-w/o\_Thinking & No-Thinking & 22.62 & 58.93 & 50.00 & 63.10 & 98.81 & 51.79 & 54.76 & \textbf{100.0} & 98.81 & 66.07 & 76.79 & 25.60 \\
          Doubao-Seed-1-6-251015-Thinking & Thinking & 24.40 & 71.43 & 55.95 & 95.83 & 84.52 & 71.43 & 97.62 & \textbf{100.0} & \textbf{100.0} & 78.57 & 72.02 & 22.02 \\
          Qwen-VL-Plus-w/o\_Thinking & No-Thinking & 19.64 & 72.02 & 52.98 & 48.81 & \textbf{100.0} & 35.71 & 57.74 & 98.81 & 82.74 & 54.17 & 70.24 & 19.64 \\
          Qwen-VL-Plus-Thinking & Thinking & 24.40 & 67.86 & 62.50 & 98.81 & \textbf{100.0} & 34.52 & \textbf{100.0} & \textbf{100.0} & \textbf{100.0} & 69.05 & 72.62 & 24.40 \\
          Gemini-3-Flash-Preview & No-Thinking & 45.83 & 85.12 & 77.98 & 97.62 & 99.40 & 86.31 & \textbf{100.0} & 99.40 & \textbf{100.0} & \textbf{85.71} & 81.55 & 26.19 \\
          Gemini-3.5-flash & No-Thinking & \textbf{60.12} & 89.88 & 76.19 & \textbf{100.0} & 99.40 & 93.45 & 99.40 & \textbf{100.0} & \textbf{100.0} & 82.74 & 95.83 & 29.76 \\

          \bottomrule
        \end{tabular}
      }
    \end{small}
  \end{center}
\end{table*}

Table~\ref{combined-eval-table} highlights four key findings about the strengths and limitations of MLLMs on multi-criteria heterogeneous graph tasks, spanning perceptual grounding, modality robustness, multimodal interaction, and explicit reasoning chains.

\textbf{Observation 1: Symbolic anchors mitigate perceptual errors in multimodal reasoning.}
Introducing multi-criteria information generally increases the reasoning burden. However, in the visually dense Metromap scenario, \textit{Map+Edge\_tab+Vertex\_tab} substantially outperforms \textit{Map-only}. For example, the EMA of GPT-5.5-Instant increases from 43.63\% to 88.88\%, while that of Gemini-3.5-flash increases from 44.19\% to 82.37\%. The structured tables provide reliable symbolic anchors for identifying stations, edges, and attributes, thereby reducing OCR and entity-grounding errors. This result identifies visual perception as an important bottleneck and shows that structured cues can stabilize subsequent reasoning.

\textbf{Observation 2: Tables are more robust than maps under perceptual challenges.}
Across nearly all models, \textit{Edge\_tab-only} consistently outperforms \textit{Map-only}. This pattern remains clear for the newly evaluated models. On Metromap, Qwen3.6-35B-A3B improves from 11.56\% to 64.88\% EMA, while GPT-5.5-Instant improves from 43.63\% to 88.75\%. Similar improvements are observed on Travelmap, where their EMA values increase from 42.74\% to 95.42\% and from 68.15\% to 98.51\%, respectively. Structured tables therefore provide more reliable inputs when visual perception is difficult. Nevertheless, the lower performance under multimodal and multi-criteria settings shows that perception is not the only bottleneck; cross-modal integration and route reasoning also remain challenging.

\textbf{Observation 3: Images become burdensome mainly in perceptually complex settings while remaining indispensable.}
The comparison between \textit{Edge\_tab-only} and \textit{Map+Edge\_tab} shows that the effect of visual inputs depends on perceptual complexity. For many Instruct models, adding maps provides greater benefits in Travelmap but little or negative improvement in Metromap. For example, Qwen3.5-9B improves from 40.83\% to 50.06\% EMA on Travelmap, but decreases from 35.56\% to 26.50\% on Metromap. Stronger models reduce this multimodal penalty: GPT-5.5-Instant and Gemini-3.5-flash show only small differences between the two settings, particularly on Travelmap. Therefore, visual inputs remain important for representing the original topology, but models require stronger visual understanding to use them without introducing additional errors.

\textbf{Observation 4: CoT helps resolve multimodal coordination challenges under perceptual difficulty.}
Thinking models generally achieve larger gains on multimodal inputs, especially when visual perception is difficult. For example, introducing Thinking increases the EMA of Doubao-Seed-1.6 from 48.06\% to 74.00\% on \textit{Map+Edge\_tab} in Metromap and from 76.85\% to 97.86\% in Travelmap. Qwen3-VL-32B shows similar improvements from 32.12\% to 44.12\% and from 68.39\% to 91.79\%, respectively. However, these gains are not universal: on Travelmap, Qwen3-VL-32B-Thinking underperforms its Instruct counterpart under \textit{Map+Edge\_tab+Vertex\_tab} and \textit{Map+Mix\_tab}. Moreover, non-thinking models such as GPT-5.5-Instant and Gemini-3.5-flash still achieve the strongest overall results. These findings indicate that CoT can improve multimodal coordination, but may cause overthinking in simpler or redundant settings and cannot compensate for limitations in the model's intrinsic reasoning capability.

\subsection{Supplementary QA Analysis}\label{subsec:supp_qa_analysis}
Table~\ref{tab:qa-performance} reveals three complementary findings about the QA performance of current MLLMs.

\textbf{Observation 1: Structured inputs largely resolve local retrieval, but not global aggregation.}
Most models achieve close to 100\% accuracy on local table-based tasks, particularly Edge\_tab-LP and Vertex\_tab-LP, indicating that retrieving a specific attribute from a structured table is no longer a major bottleneck. However, performance remains much lower on GP tasks that require counting, comparison, or aggregation over many entries. For example, Qwen3.6-35B-A3B achieves 98.75\% on Metromap Edge\_tab-LP but only 71.88\% on Edge\_tab-GP, while its Travelmap Map-LP accuracy reaches 75.00\% compared with only 23.21\% on Map-GP. Therefore, the main difficulty is not simply locating individual information, but maintaining complete coverage and correctly aggregating information across the entire map or table.

\textbf{Observation 2: Cross-modal difficulty depends on the specific relation that must be reconstructed.}
The Map+Mix\_tab results exhibit markedly different patterns across the two scenarios. In Metromap, models perform well on SR but poorly on GP: Qwen3.6-35B-A3B achieves 95.63\% on SR but only 15.00\% on GP. This is because the GP task requires reconstructing complete line structures from the map and then aggregating the corresponding station attributes from the table. In Travelmap, the strongest models achieve relatively high GP and LP accuracy, but remain weak on SR; GPT-5.5-Instant and Gemini-3.5-flash obtain 32.74\% and 29.76\%, respectively. Here, models must first identify adjacent locations from the map, align them with table entries, and then compare their prices. These results show that cross-modal reasoning is not uniformly difficult: its performance depends on how visual relations and tabular attributes must be aligned and composed.

\textbf{Observation 3: Explicit CoT provides task-dependent gains rather than universal improvements.}
Thinking models often improve tasks that require intermediate structural reasoning. In Metromap, Qwen3-VL-8B improves from 22.50\% to 56.87\% on Edge\_tab-GP and from 57.50\% to 79.37\% on Vertex\_tab-GP, while Doubao-Seed-1.6 improves from 41.88\% to 72.50\% and from 58.13\% to 96.25\% on the same tasks. However, CoT may reduce performance on more direct perceptual or retrieval-oriented tasks: Qwen3-VL-8B drops from 73.12\% to 51.25\% on Metromap Map-SR and from 22.50\% to 9.38\% on Map+Mix\_tab-LP. Moreover, non-thinking models such as GPT-5.5-Instant and Gemini-3.5-flash achieve the strongest results in many settings. This suggests that CoT is beneficial when a task genuinely requires intermediate information composition, but may introduce unnecessary reasoning noise when the answer depends mainly on direct perception or lookup.

\section{More Experiments}\label{sec:more_experiments}
\subsection{Ablation study of RP and QA for resolution}\label{subsec:abl_res_rp_and_qa}
\begin{figure*}[!ht] 
    \centering
    \begin{subfigure}[t]{0.2\textwidth}
        \centering
        \includegraphics[width=\linewidth]{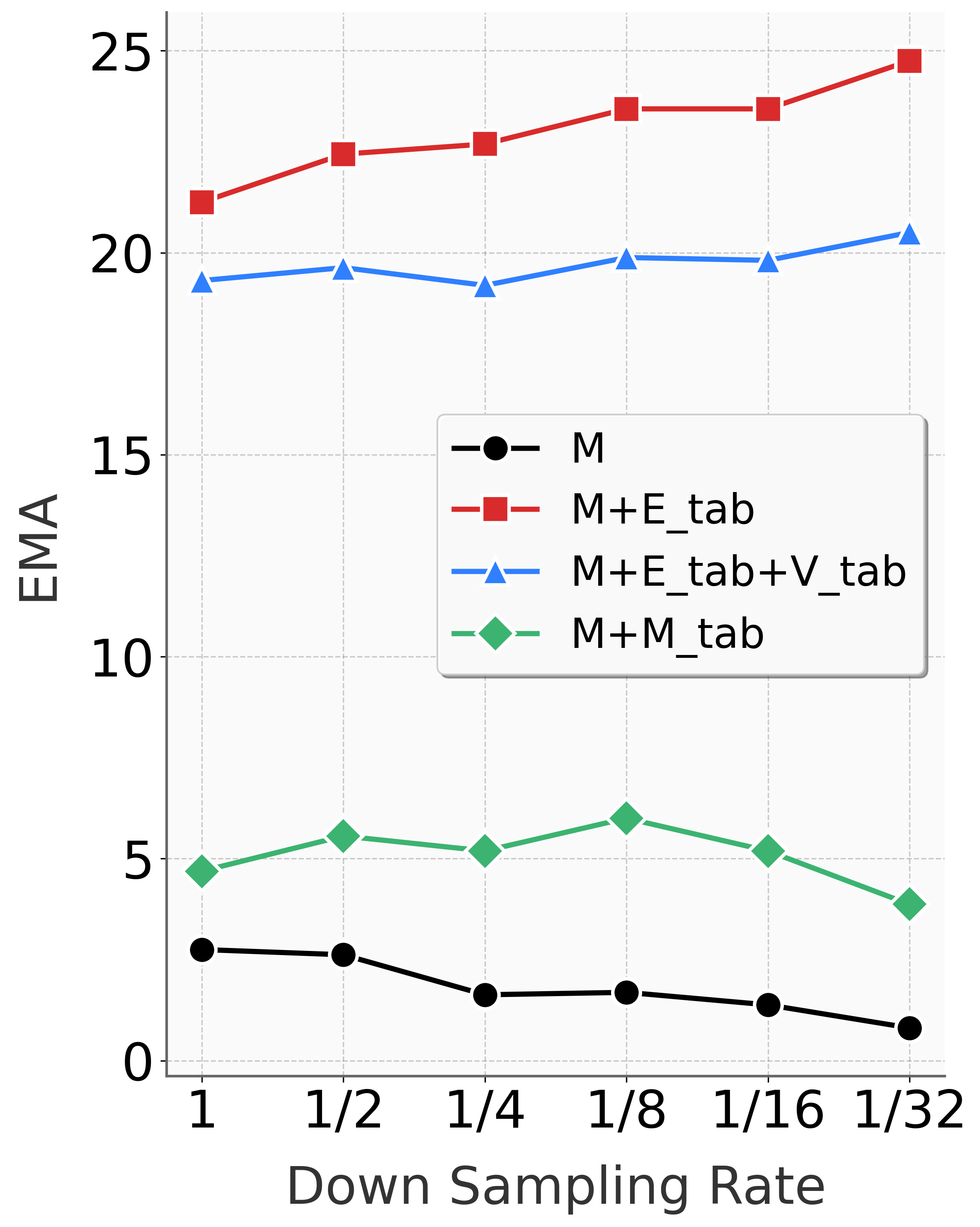} 
        \caption{RP for Metromap}
        \label{fig:metromap_planning_down_sampling}
    \end{subfigure}
    \hfill 
    \begin{subfigure}[t]{0.2\textwidth}
        \centering
        \includegraphics[width=\linewidth]{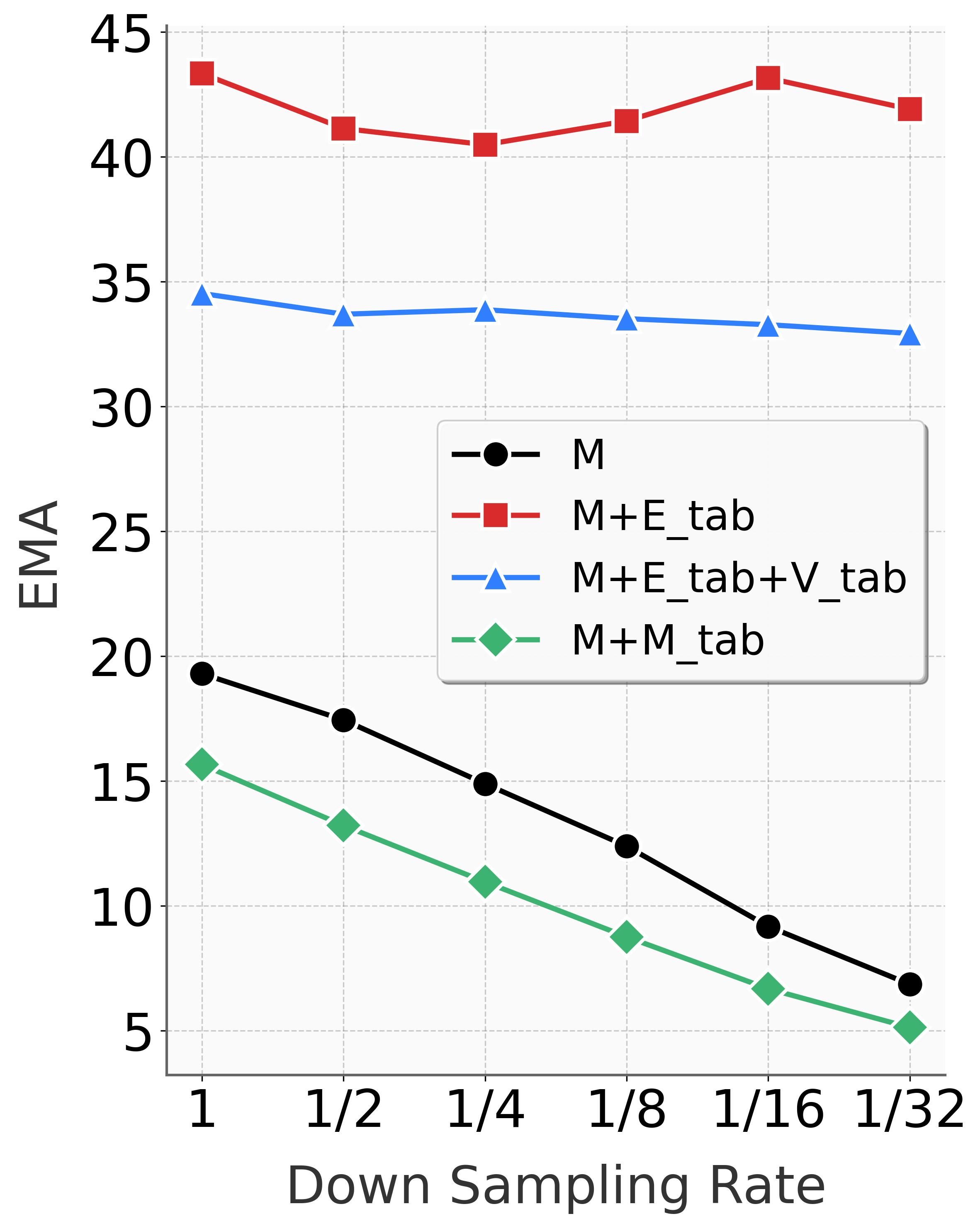}
        \caption{RP for Travelmap}
        \label{fig:travelmap_planning_down_sampling}
    \end{subfigure}
    \hfill 
    \begin{subfigure}[t]{0.2\textwidth}
        \centering
        \includegraphics[width=\linewidth]{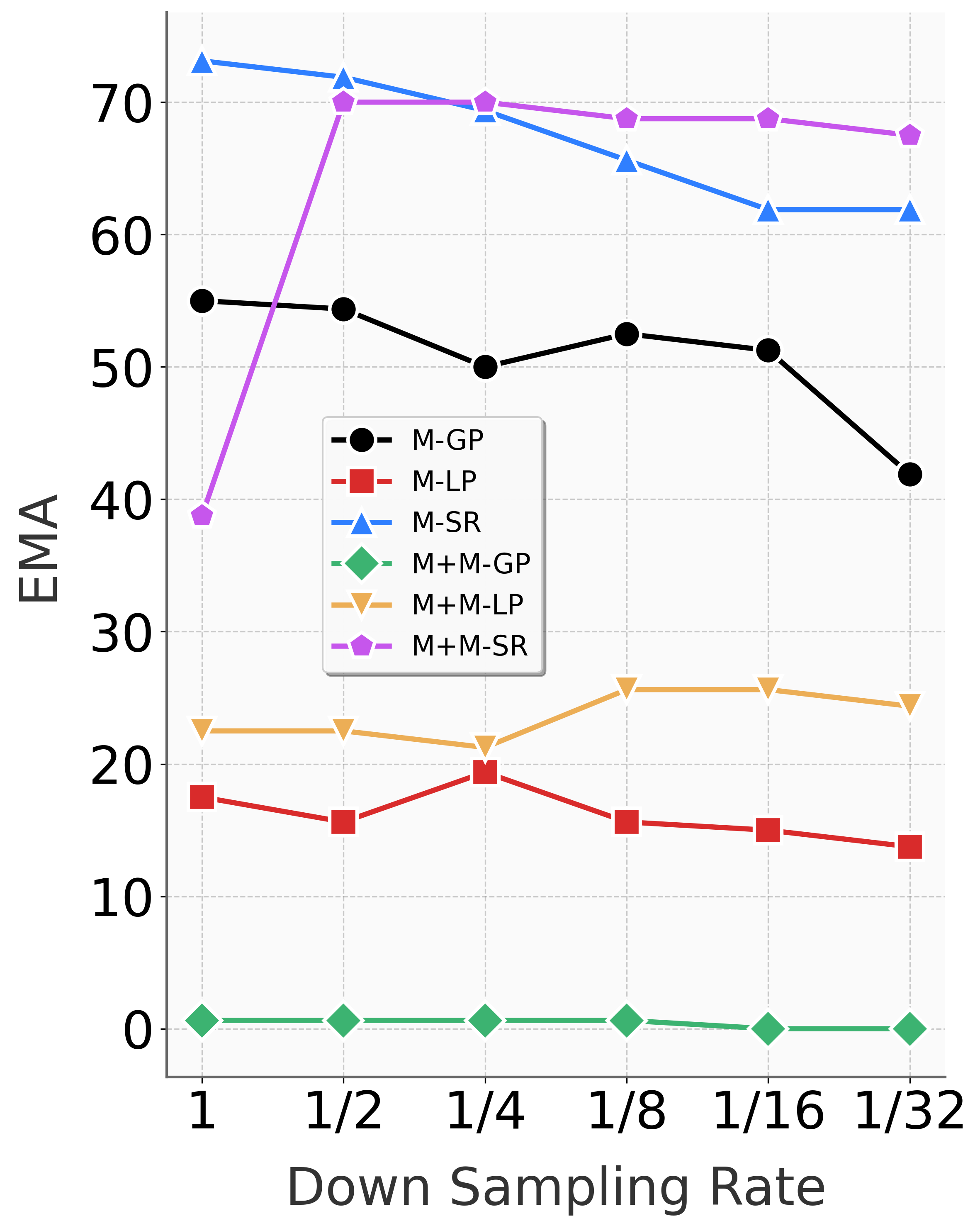} 
        \caption{QA for Metromap}
        \label{fig:metromap_qa_down_sampling}
    \end{subfigure}
    \hfill 
    \begin{subfigure}[t]{0.2\textwidth}
        \centering
        \includegraphics[width=\linewidth]{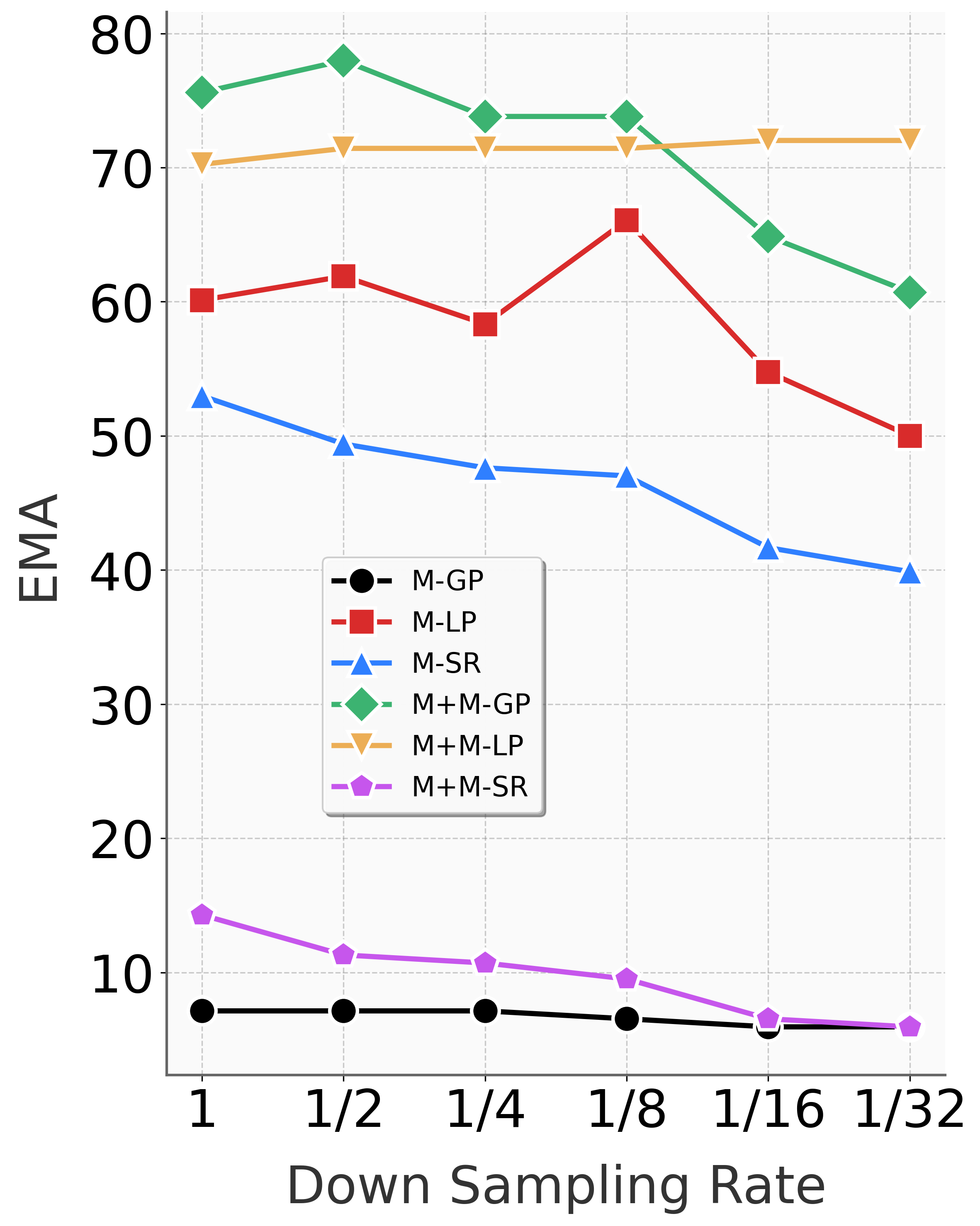}
        \caption{QA for Travelmap}
        \label{fig:travelmap_qa_down_sampling}
    \end{subfigure}
    \caption{Impact of Image Resolution on RP and QA Tasks in Metromap and Travelmap Scenarios. Images are proportionally downsampled to $1/2$, $1/4$, $1/8$, $1/16$, and $1/32$ of the original resolution. 
    For RP task, evaluation is conducted under 4 Map-based input settings: Map-Only (M), Map+Edge\_tab (M+E\_tab), Map+Edge\_tab+Vertex\_tab (M+E\_tab+V\_tab), and Map+Mix\_tab (M+M\_tab).
    For QA task, evaluation is conducted under 6 Map-based input settings: Map-GP (M-GP), Map-LP (M-LP), Map-SR (M-SR), Map+Mix\_tab-GP (M+M-GP), Map+Mix\_tab-LP (M+M-LP), and Map+Mix\_tab-SR (M+M-SR).}
    \label{fig:resolution_impact}
\end{figure*}
To analyze the effects of visual perception degradation across 2 task types, Figure \ref{fig:resolution_impact} illustrates the impact of different image resolutions on RP and QA task under two scenarios.

We first discuss RP results, where resolution effects differ across scenarios and input configurations:

\textbf{Metromap scenario.} Under Map+Edge\_tab and Map+E\_tab+Mix\_tab, image downsampling improves performance despite reduced visual fidelity, indicating that models primarily rely on E\_tab while visual maps may introduce interference. Under Map+Mix\_tab, performance remains largely stable, suggesting dependence on global graph information rather than local visual details. In contrast, in the Map-only setting, performance consistently degrades with downsampling, demonstrating that purely visual reasoning is highly sensitive to visual quality.

\textbf{Travelmap scenario.} Downsampling generally degrades performance, but a rebound under the E\_tab setting indicates that models initially attend to both visual and tabular inputs, but progressively shift their reasoning toward E\_tab as image quality degrades. These results suggest that E\_tab parsing defines the performance floor, whereas visual perception quality determines the performance ceiling. Thus, improving model's performance ceiling is more fundamentally contingent upon the accurate perception and stable representation of crucial information in complex images.

For QA queries, experimental results indicate that improvements in image clarity are not a sufficient condition for enhancing model performance. On the contrary, for multiple categories of questions, moderately downsampled images often lead to better reasoning outcomes. In some tasks, model accuracy even exhibits a stable upward trend as the degree of downsampling increases. This phenomenon suggests that the sampling rate itself is not the key determinant of model performance. Instead, model behavior is more strongly constrained by the perceptual granularity and reasoning paradigm required by the specific task. When a task primarily relies on global structural understanding or high-level semantic abstraction, reducing visual redundancy can help suppress irrelevant details, thereby enabling the model to focus more effectively on information that is critical for reasoning.

Overall, the RP and QA experiments all suggest that performance is jointly shaped by input modality preference, task structure, and perceptual granularity requirements.

\subsection{Ablation study of Map difficulty and Query difficulty}\label{subsec:abl_map_and_query_dfficulty}
\begin{figure*}[!ht] 
    \centering
    
    
    \begin{subfigure}[b]{0.48\linewidth}
        \centering
        \includegraphics[width=\linewidth]{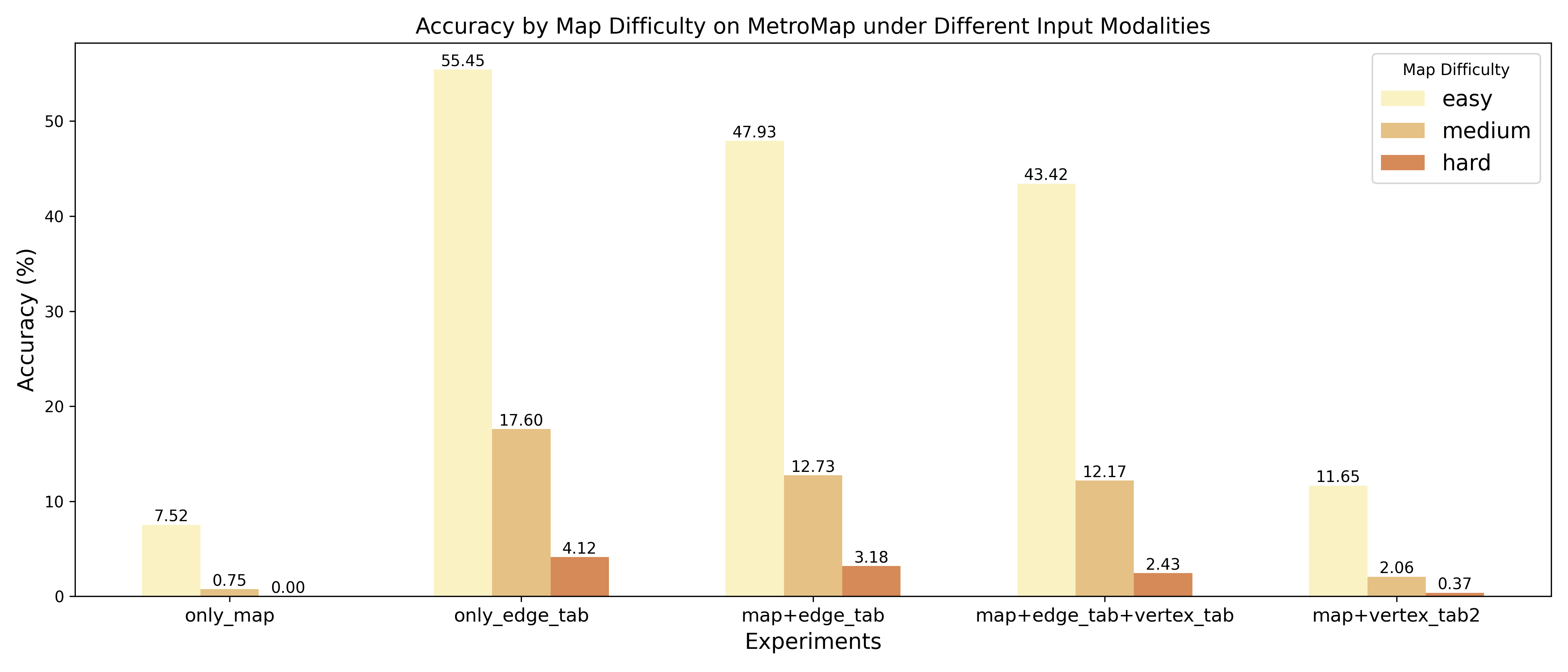} 
        \caption{Metromap-Map difficulty}
    \end{subfigure}
    \hfill 
    \begin{subfigure}[b]{0.48\linewidth}
        \centering
        \includegraphics[width=\linewidth]{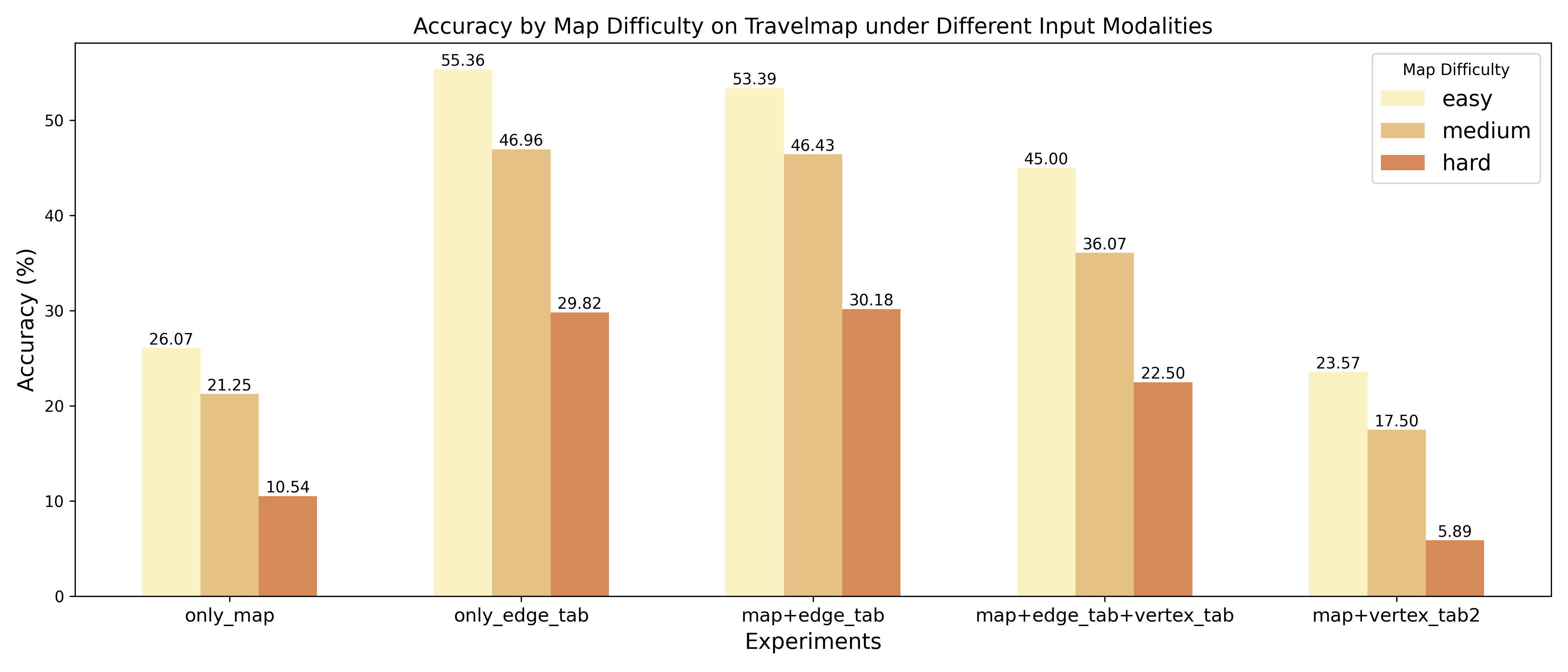}
        \caption{Travelmap-Map difficulty}
    \end{subfigure}
    
    \vspace{0.5em} 
    
    
    \begin{subfigure}[b]{0.48\linewidth}
        \centering
        \includegraphics[width=\linewidth]{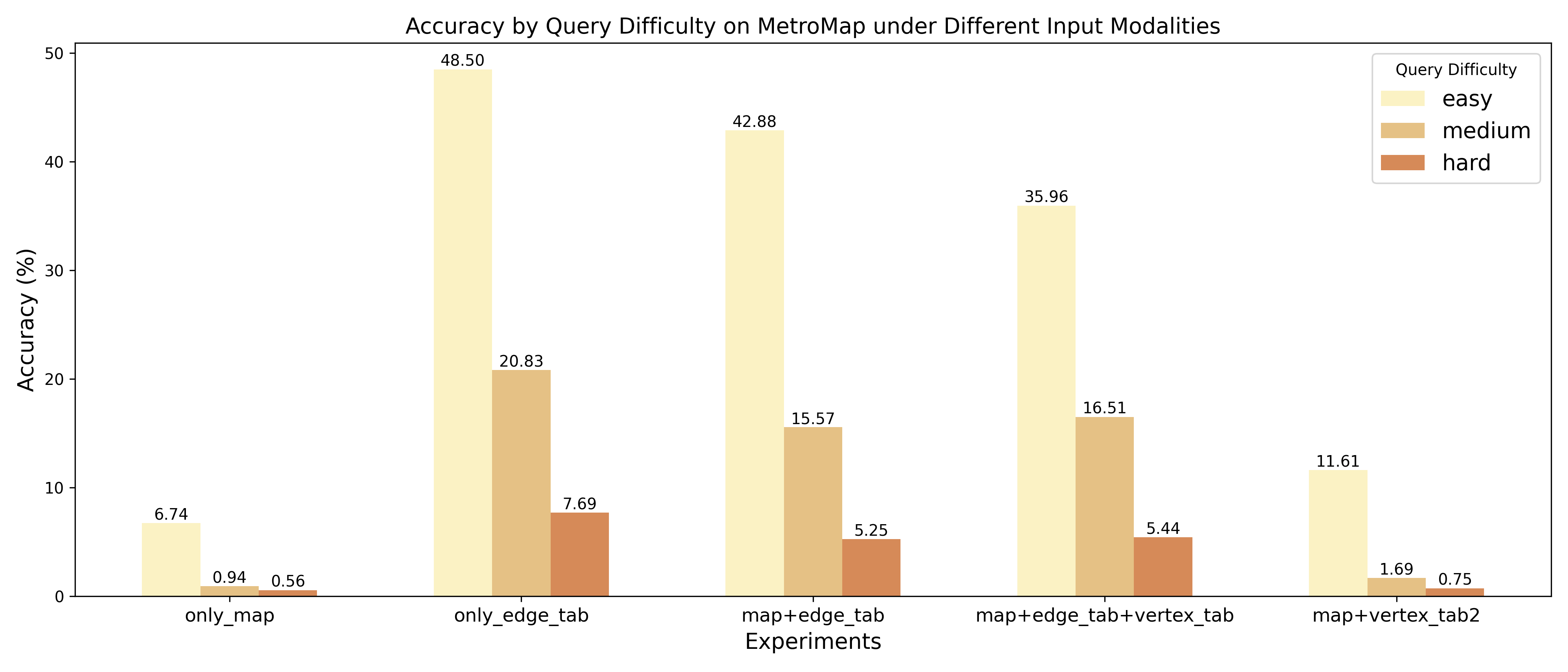}
        \caption{Metromap-Query difficulty}
    \end{subfigure}
    \hfill
    \begin{subfigure}[b]{0.48\linewidth}
        \centering
        \includegraphics[width=\linewidth]{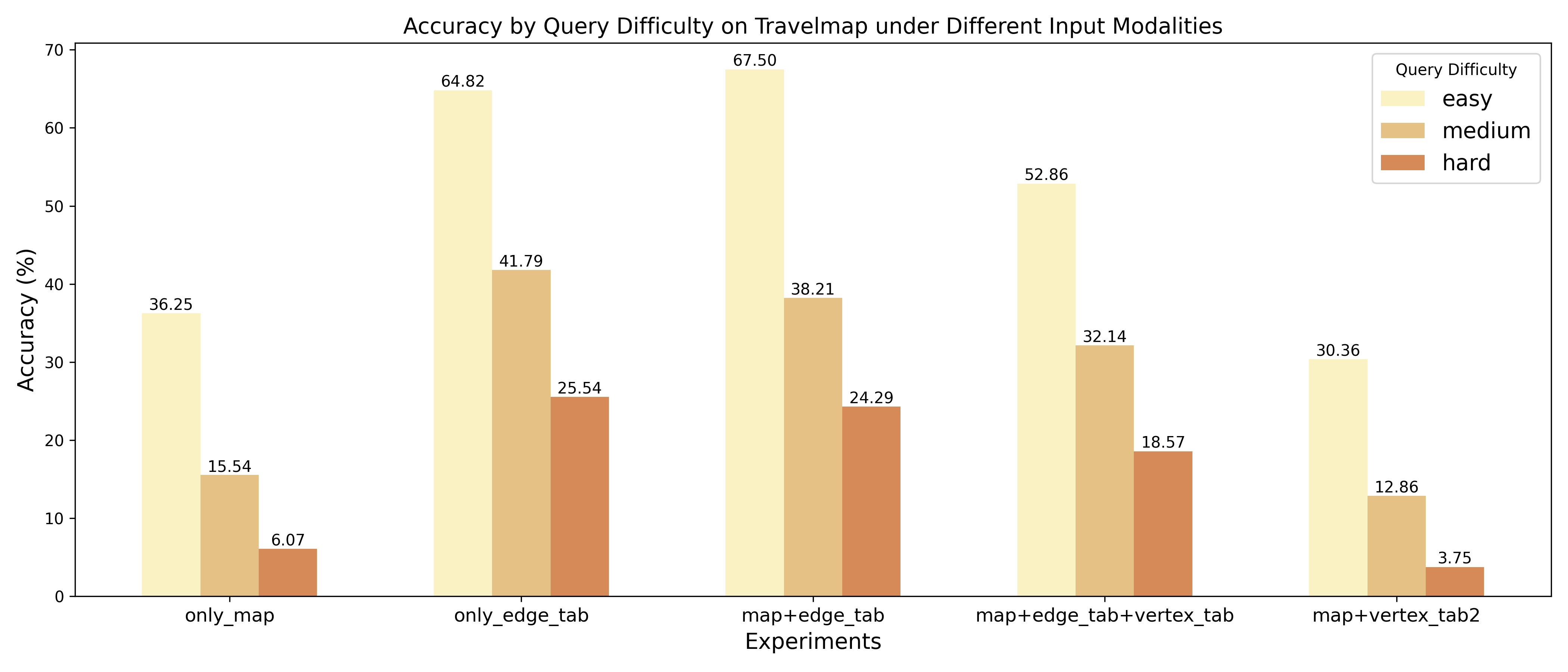}
        \caption{Travelmap-Query difficulty}
    \end{subfigure}
    
    \caption{Distribution of model accuracy across Map Difficulty and Query Difficulty under different input modalities (Metromap and Travelmap scenarios).}
    \label{fig:model_acc_bar}
\end{figure*}

Figure \ref{fig:model_acc_bar} illustrates in detail the distribution of model accuracy across different input modalities as a function of Map Difficulty and Query Difficulty. We observed distinctly different performance decay patterns between the two datasets:

\begin{itemize}
    \item \textbf{Polarized Performance and Reasoning Collapse in Metromap:} In the Metromap scenario, the accuracy distribution exhibits significant \textbf{imbalance}. Models perform reasonably well on ``Easy" samples, but experience a \textbf{precipitous drop} on "Medium" and "Hard" levels, with accuracy for ``Hard" samples approaching zero. This phenomenon indicates that the high-level topological structures in Metromap, such as complex transfer networks and densely connected nodes, have reached or even exceeded the \textbf{cognitive capacity ceiling} of current models. Once the graph complexity surpasses a critical threshold, the reasoning ability collapses systematically rather than declining gradually.
    
    \item \textbf{Linear Decay and Adaptivity in Travelmap:} In contrast, Travelmap shows a more stable performance across difficulty levels, exhibiting a uniform distribution. Although accuracy decreases with increasing difficulty, the decay curve remains gradual without functional failure. This suggests that the topological logic in Travelmap, even in the Hard mode, largely remains within the models' effective reasoning boundary, allowing them to maintain relatively stable generalization capabilities.
\end{itemize}
\begin{figure*}[!ht]
    \centering
    
    
    \begin{subfigure}[b]{0.19\linewidth}
        \centering
        \includegraphics[width=\linewidth]{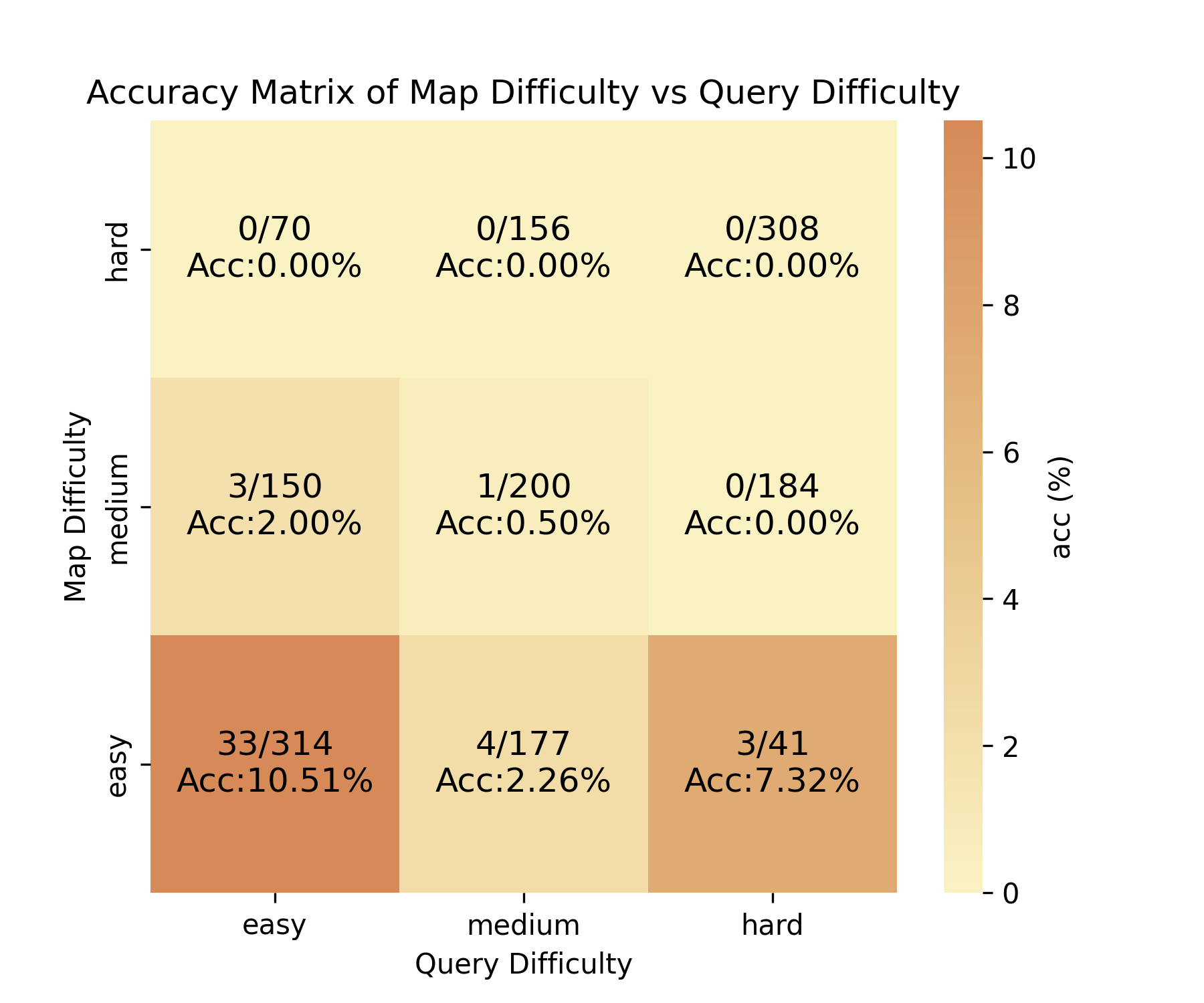}
        \caption{Metro-M}
        \label{fig:row1_col1}
    \end{subfigure}
    \hfill 
    \begin{subfigure}[b]{0.19\linewidth}
        \centering
        \includegraphics[width=\linewidth]{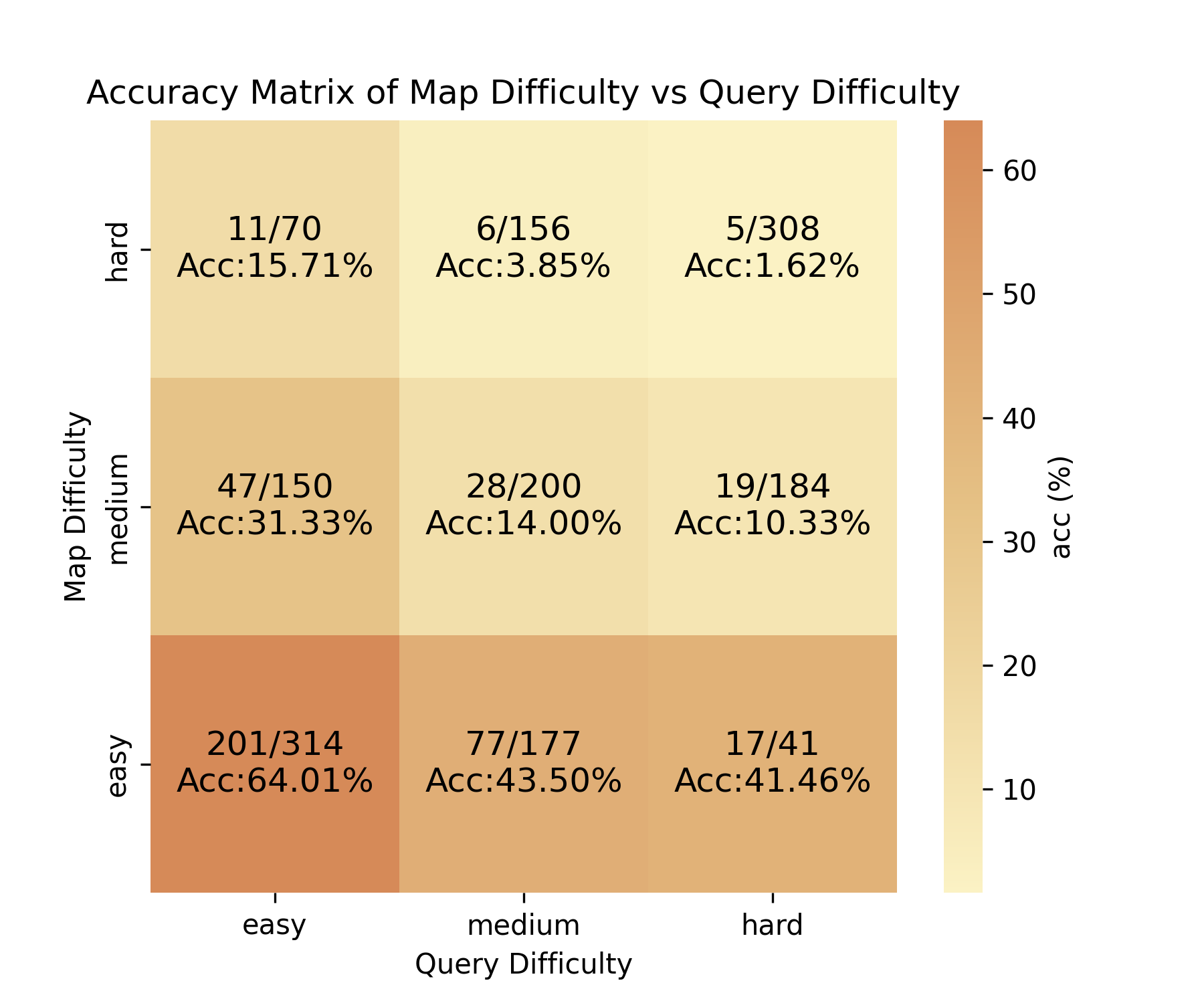}
        \caption{Metro-E}
    \end{subfigure}
    \hfill
    \begin{subfigure}[b]{0.19\linewidth}
        \centering
        \includegraphics[width=\linewidth]{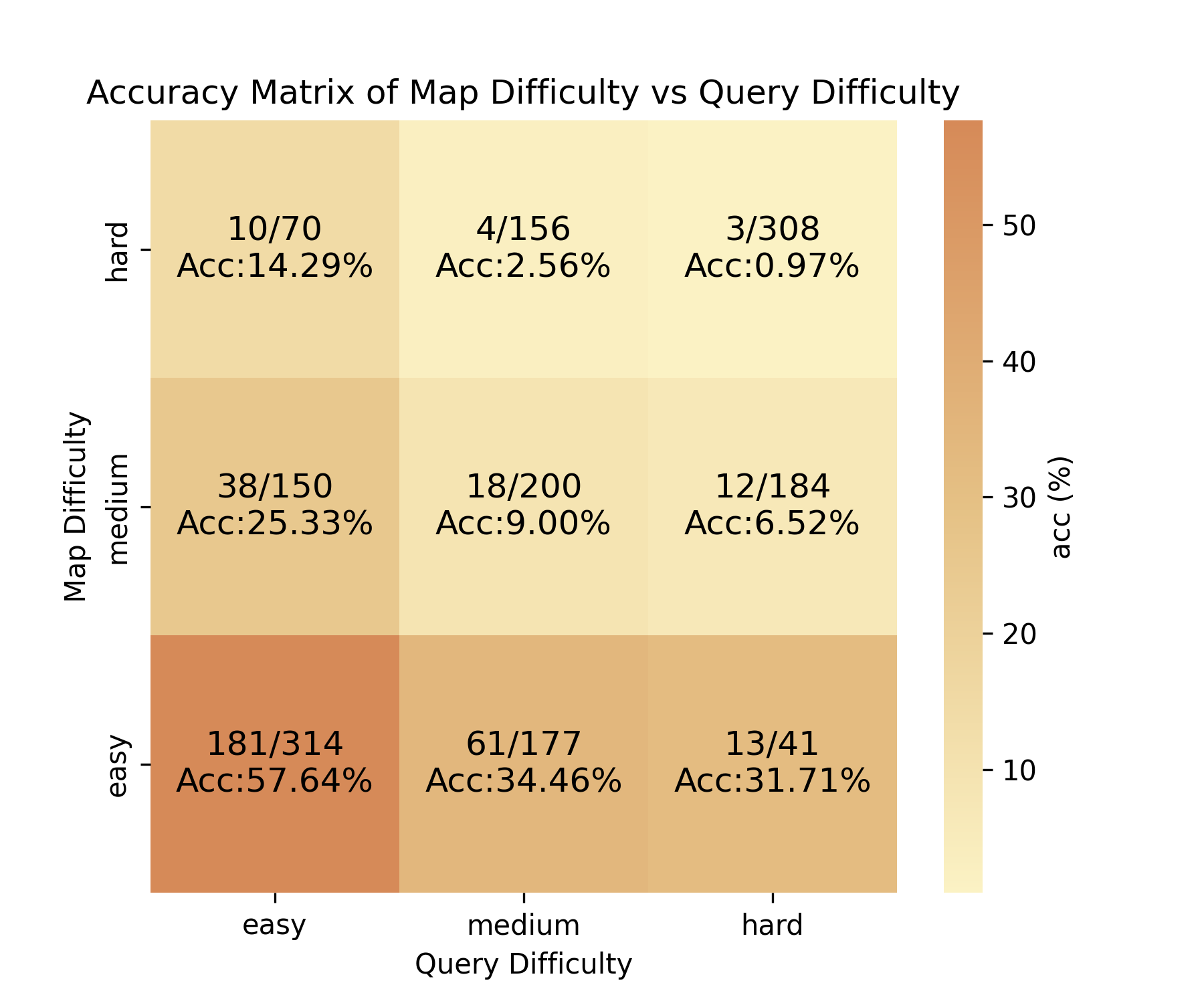}
        \caption{Metro-M+E}
    \end{subfigure}
    \hfill
    \begin{subfigure}[b]{0.19\linewidth}
        \centering
        \includegraphics[width=\linewidth]{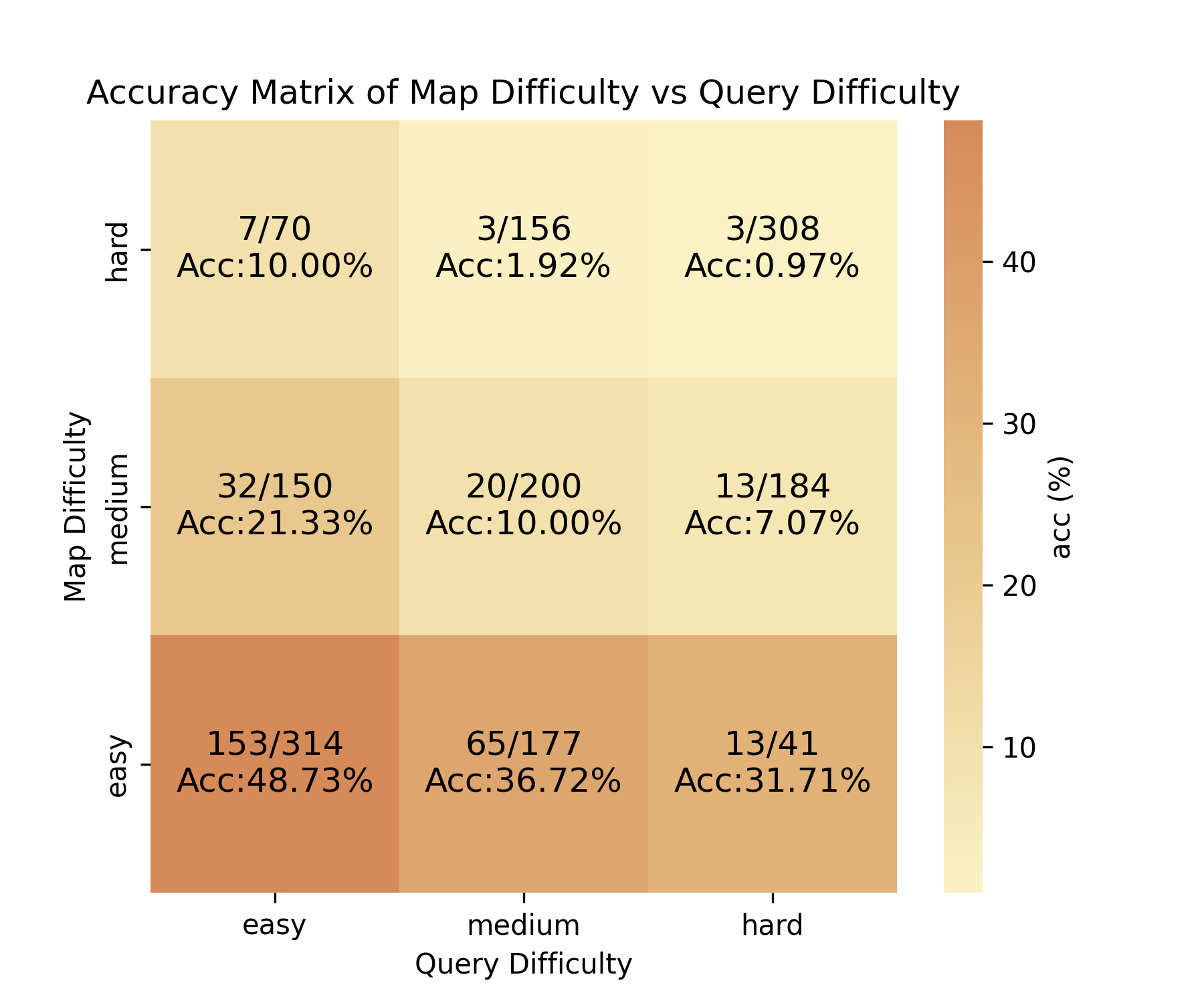}
        \caption{Metro-M+E+V}
    \end{subfigure}
    \hfill
    \begin{subfigure}[b]{0.19\linewidth}
        \centering
        \includegraphics[width=\linewidth]{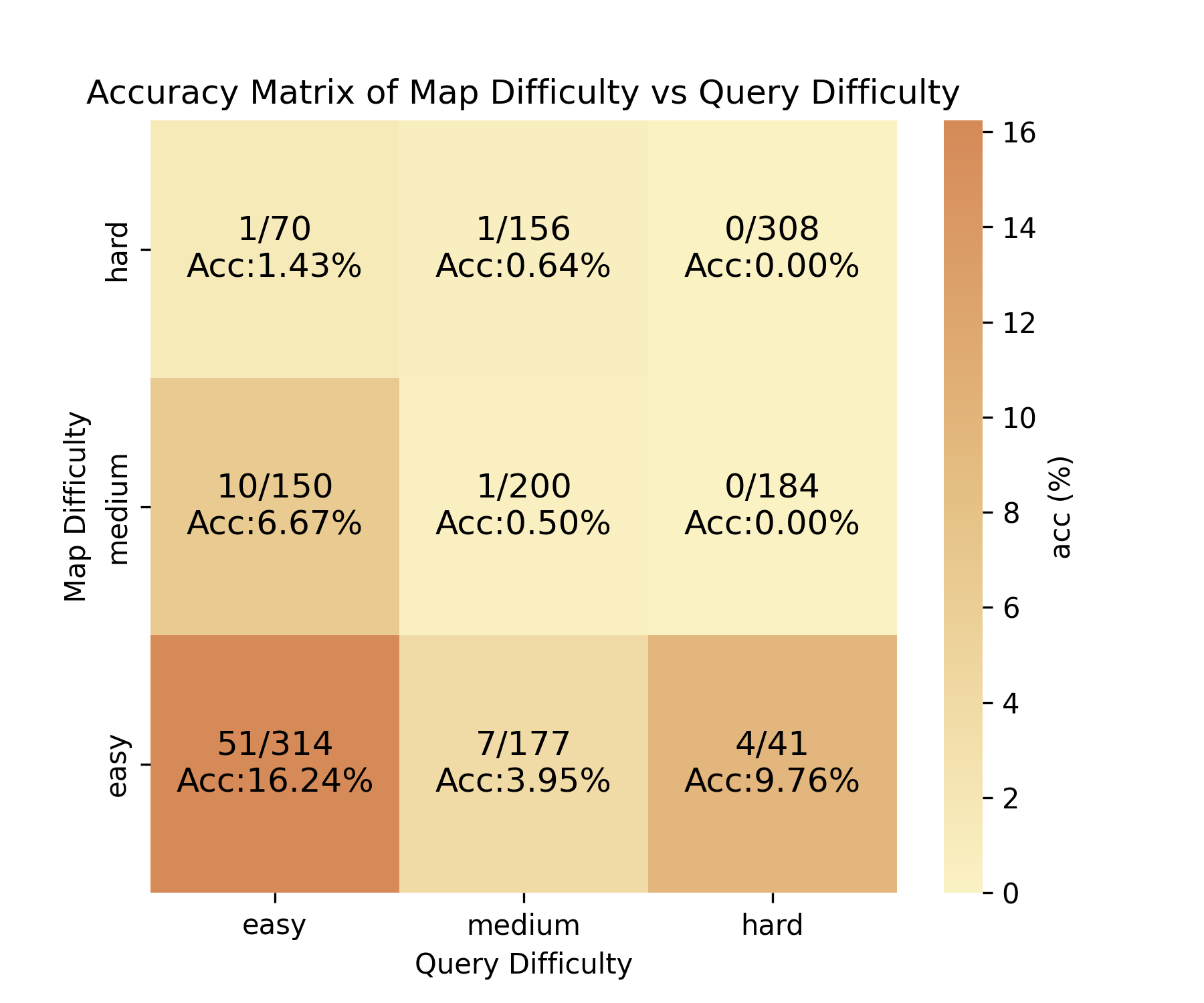}
        \caption{Metro-M+M}
    \end{subfigure}

    \par\vspace{1em} 
    
    \begin{subfigure}[b]{0.19\linewidth}
        \centering
        \includegraphics[width=\linewidth]{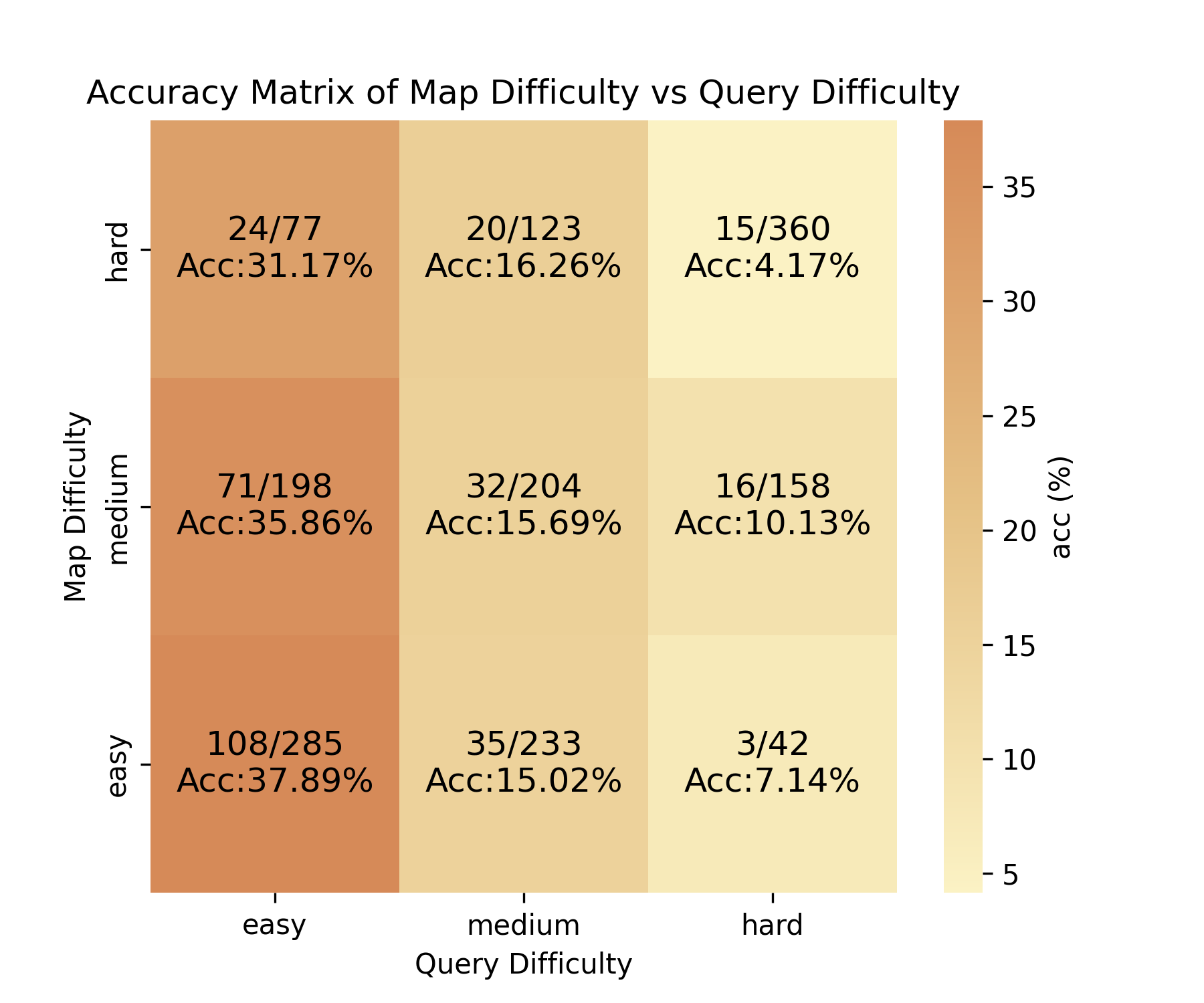}
        \caption{Travel-M}
    \end{subfigure}
    \hfill
    \begin{subfigure}[b]{0.19\linewidth}
        \centering
        \includegraphics[width=\linewidth]{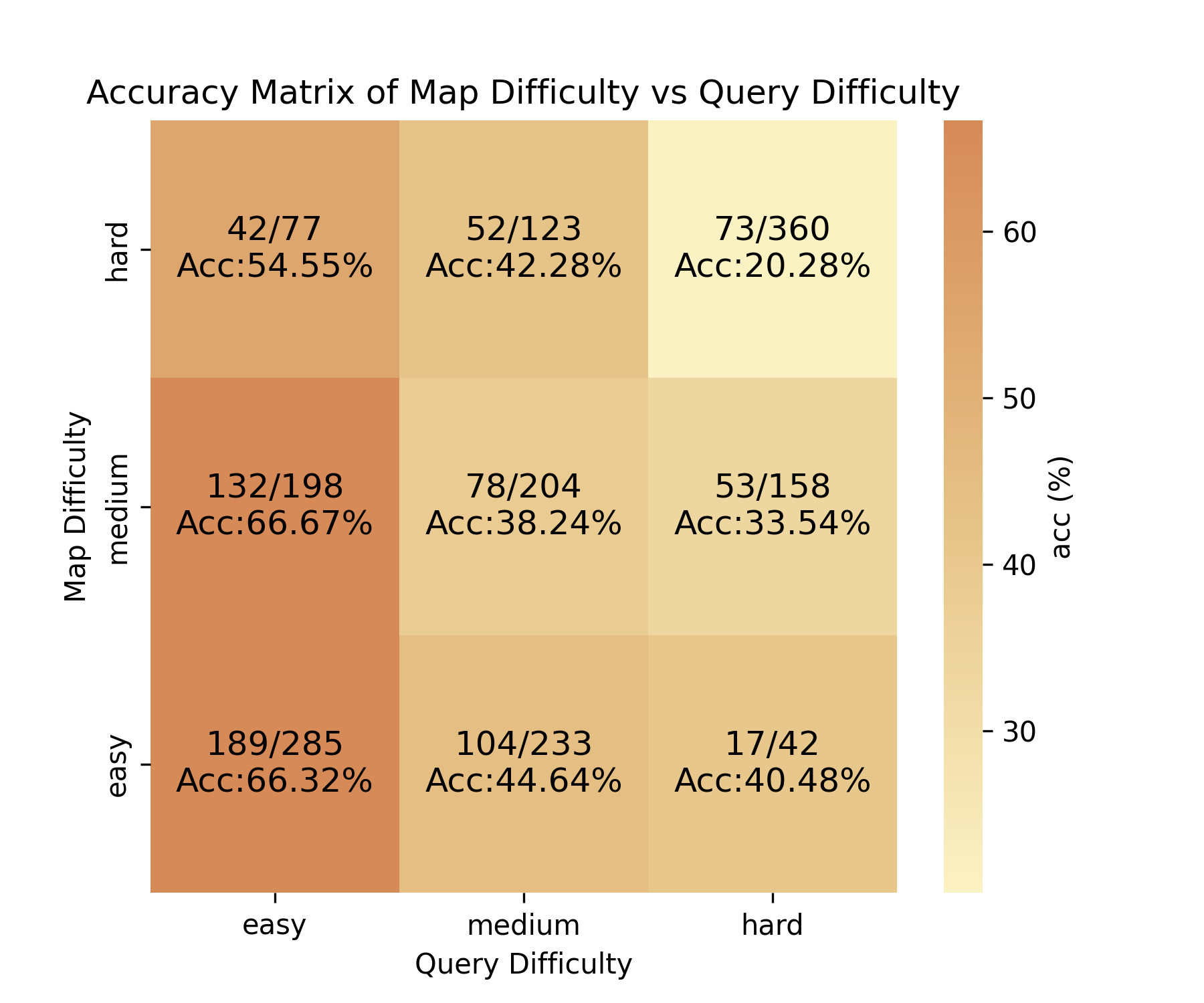}
        \caption{Travel-E}
    \end{subfigure}
    \hfill
    \begin{subfigure}[b]{0.19\linewidth}
        \centering
        \includegraphics[width=\linewidth]{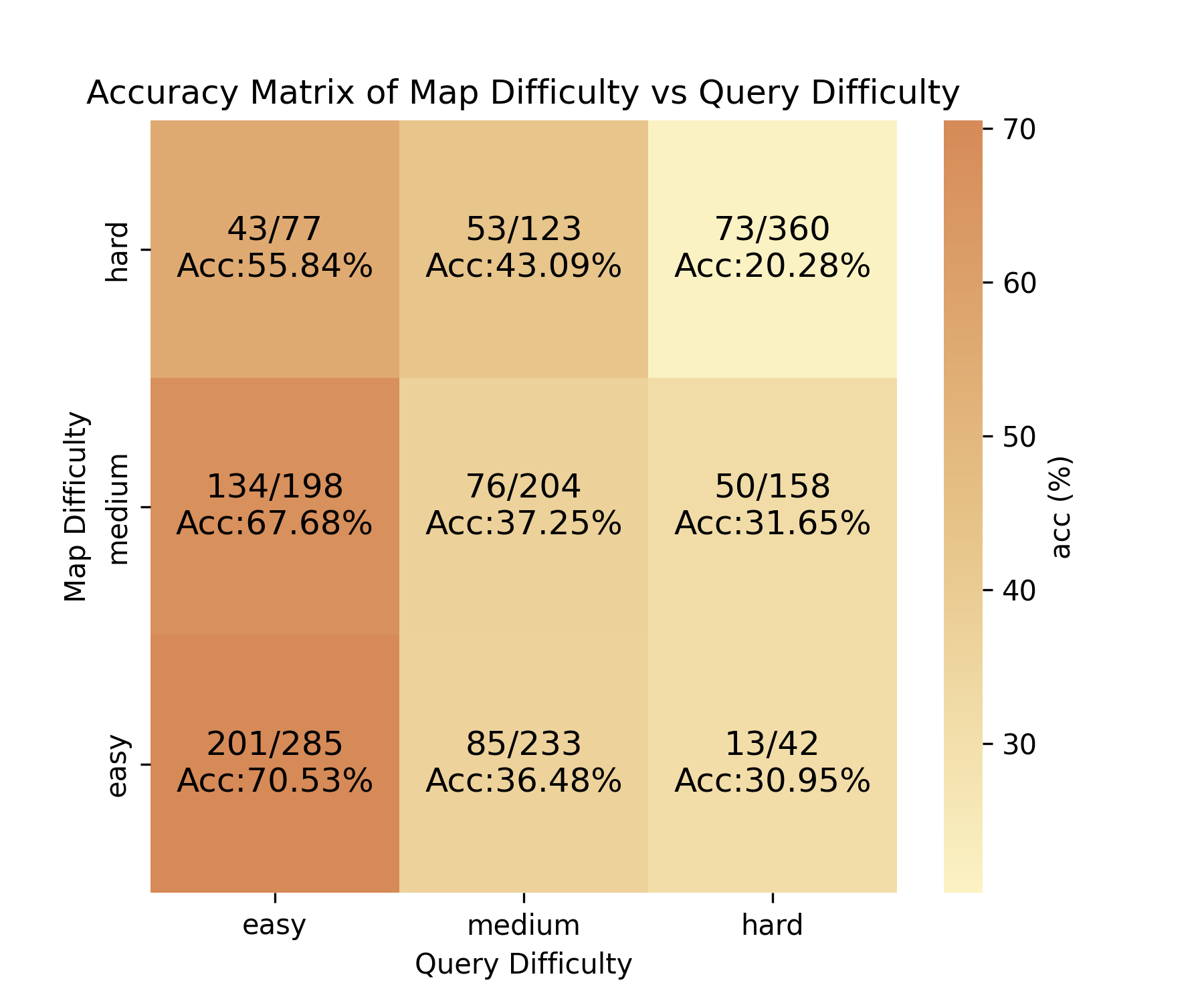}
        \caption{Travel-M+E}
    \end{subfigure}
    \hfill
    \begin{subfigure}[b]{0.19\linewidth}
        \centering
        \includegraphics[width=\linewidth]{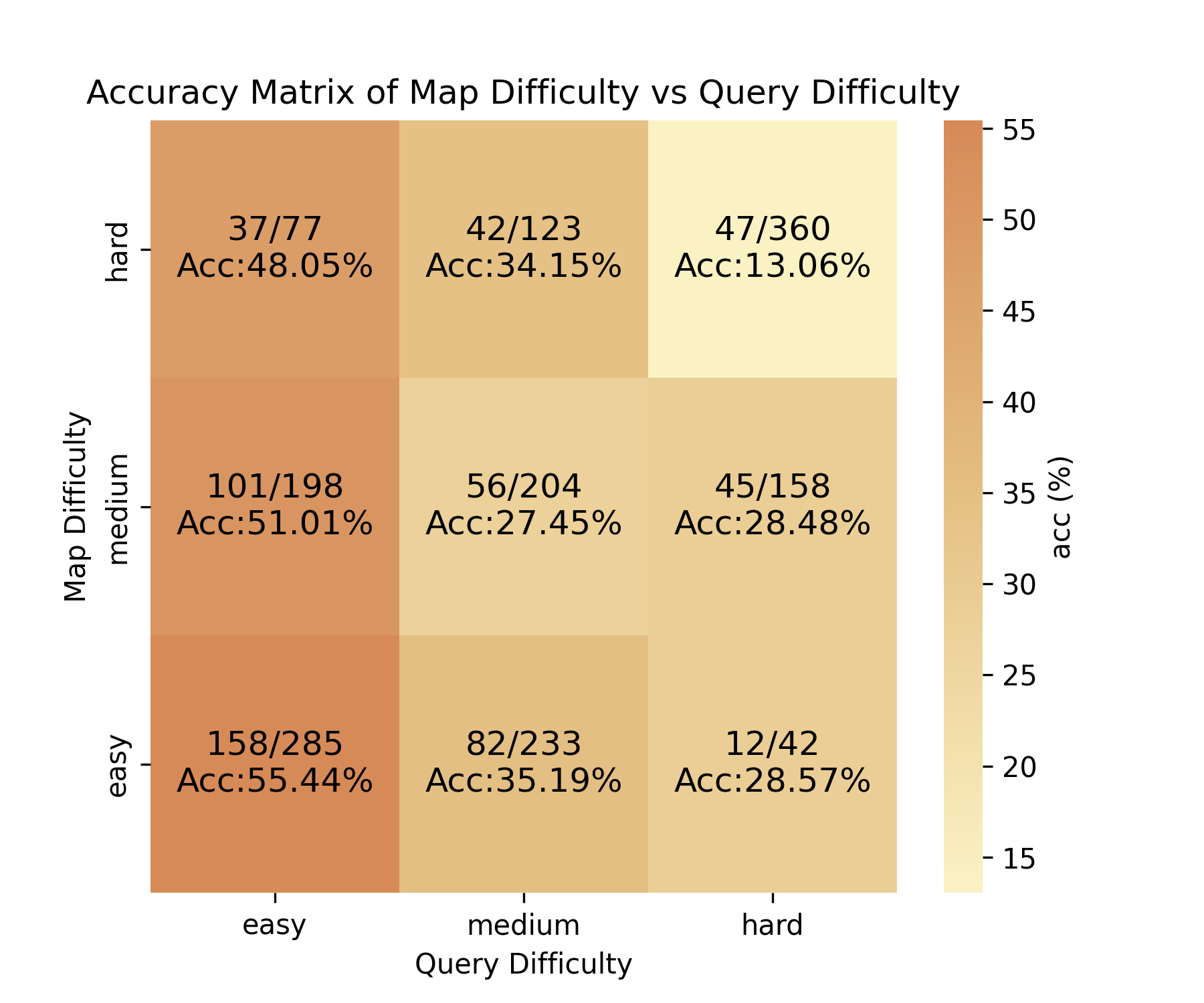}
        \caption{Travel-M+E+V}
    \end{subfigure}
    \hfill
    \begin{subfigure}[b]{0.19\linewidth}
        \centering
        \includegraphics[width=\linewidth]{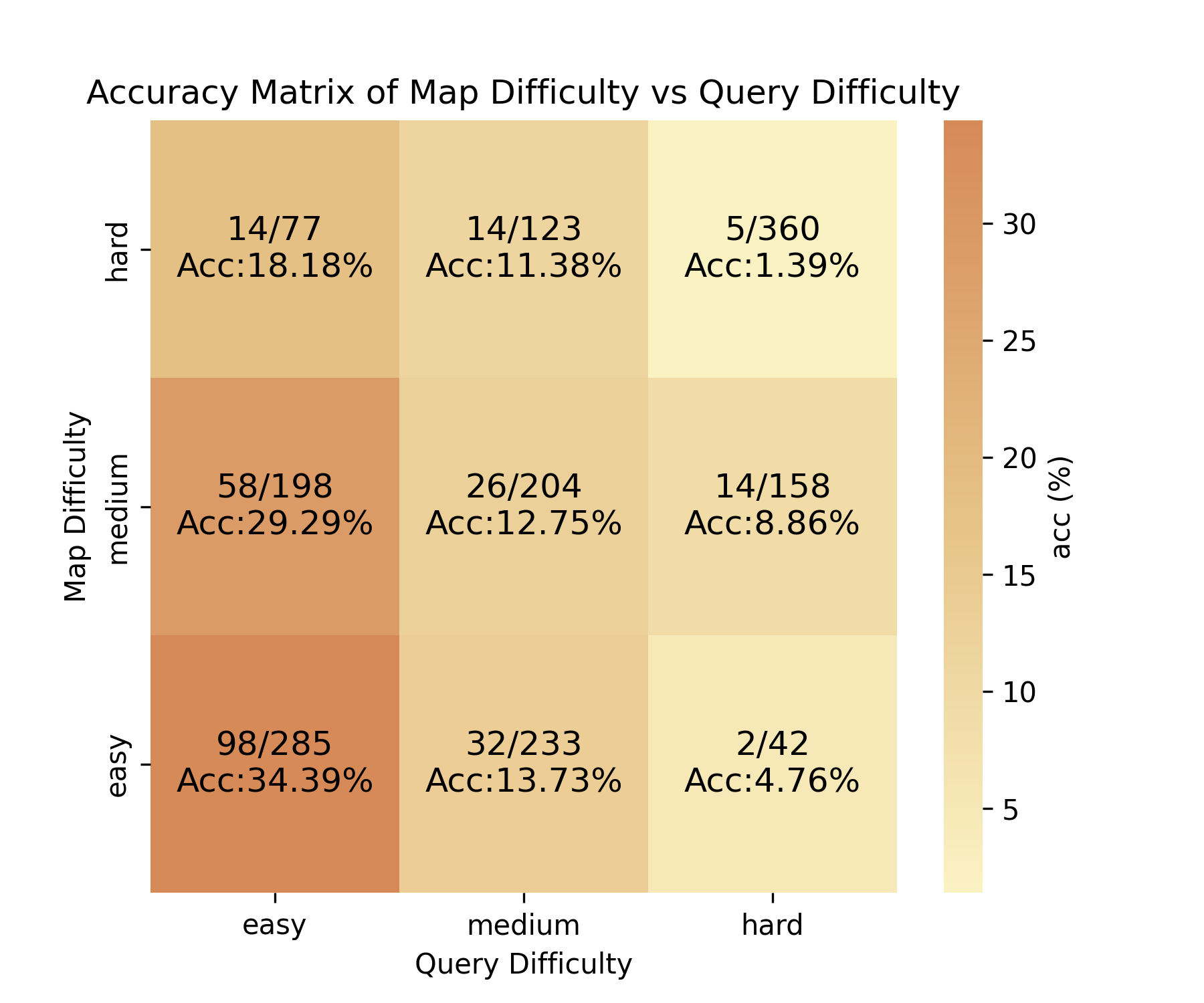}
        \caption{Travel-M+M}
    \end{subfigure}
    
    \caption{Model accuracy matrix under different combinations of Map Difficulty and Query Difficulty. 
The first row shows results in the Metromap scenario, while the second row corresponds to the Travelmap scenario. 
M denotes Map Only; E denotes Edge\_Tab Only; M+E denotes Map + Edge\_Tab; M+E+V denotes Map + Edge\_Tab + Vertex\_Tab; 
and M+M denotes Map + Mix\_tab.}

    \label{fig:model_acc_matrix}
\end{figure*}

Through a cross-analysis of Map Difficulty and Query Difficulty, Figure \ref{fig:model_acc_matrix} presents a more fine-grained distribution of performance gradients. Overall, model accuracy exhibits a pronounced stepwise decline as task complexity increases, progressively degrading from the Easy-Easy combination to the Hard-Hard combination.

Further fine-grained analysis reveals clear differences in sensitivity to the sources of complexity across task scenarios. In the Metromap task, map topological complexity has a significantly greater impact on final performance than query logical complexity, indicating that the overall visual complexity and information density of metro network diagrams constitute the primary challenge. In contrast, in the Travelmap task, query logical complexity exerts a substantially stronger influence on performance than map topological complexity, suggesting that the reasoning structure of the queries and multi-step logical composition are the key limiting factors in this scenario.

These results demonstrate that the dominant factors constraining model performance vary across task scenarios. Consequently, distinct optimization strategies are required for scenarios dominated by perceptual complexity versus those dominated by reasoning complexity.

\subsection{Ablation study of Tabular Modality}\label{subsec:abl_tabular}
We adopt two table formats, namely CSV and JSON, for the tabular modality. Table \ref{tab:modality_token} reports the token counts of three table types, including Edge\_tab, Vertex\_tab, and Mix\_tab, under both formats. In addition, Table \ref{tab:modality_token} investigates the performance of CSV and JSON formats in two multi-criteria RP settings, namely Map+Edge\_tab+Vertex\_tab and Map+Mix\_tab, analyzing whether different table representations affect the model's perception and understanding of tabular content. Table \ref{tab:modality_perf} further presents comparative experimental results of the two formats on QA queries.

We initially hypothesized that the JSON format would provide stronger structural cues and thus improve model understanding and overall performance, whereas the CSV format, despite being more token-efficient, might suffer from weaker interpretability. However, experimental results reveal that although the CSV format uses more than 50\% fewer tokens than the JSON format, its performance is not inferior in either RP or QA tasks, and even outperforms JSON in certain cases. This indicates that while the JSON format offers richer structural information, it does not increase the upper bound of model perception and reasoning capabilities. On the contrary, the longer context induced by JSON representations may sometimes hinder reasoning performance.

\begin{table*}[!ht]
\centering

\caption{Token counts under CSV and JSON table formats and their performance comparison on RP queries.}
\label{tab:modality_token}
\small 
\setlength{\tabcolsep}{4pt} 
\renewcommand{\arraystretch}{1.2}

\begin{tabular}{l ccc ccc ccc}
\toprule
 & \multicolumn{3}{c}{Token Statistics} & \multicolumn{3}{c}{Map+E\_tab+V\_tab} & \multicolumn{3}{c}{Map+M\_tab} \\
\cmidrule(lr){2-4} \cmidrule(lr){5-7} \cmidrule(lr){8-10}

\textbf{Input format} & E\_tab & V\_tab & M\_tab & \textbf{EMA} & \textbf{PMA} & \textbf{DS} & \textbf{EMA} & \textbf{PMA} & \textbf{DS} \\
\midrule

\multicolumn{10}{c}{\textbf{Metromap}} \\
\midrule
csv & 2651 & 2264 & 2496 & \textbf{19.75} & \textbf{40.16} & \textbf{876} & \textbf{6.31} & \textbf{24.48} & \textbf{255} \\
json & \textbf{6465} & \textbf{6193} & \textbf{6397} & 19.31 & 39.31 & 855 & 4.69 & 21.87 & 182 \\
\midrule

\multicolumn{10}{c}{\textbf{Travelmap}} \\
\midrule
csv & 577 & 462 & 462 & 33.39 & 55.10 & 1929 & 15.42 & 40.83 & 800 \\
json & \textbf{1299} & \textbf{1131} & \textbf{1131} & \textbf{34.52} & \textbf{55.56} & \textbf{2002} & \textbf{15.65} & \textbf{40.97} & \textbf{804} \\
\bottomrule
\end{tabular}
\end{table*}

\begin{table*}[!ht]
\centering

\caption{Performance comparison of CSV and JSON table formats on QA queries.}
\label{tab:modality_perf}
\footnotesize 
\setlength{\tabcolsep}{3pt} 
\renewcommand{\arraystretch}{1.2}

\begin{tabular}{l ccccccccc}
\toprule
\textbf{Input format} & E-GP & E-LP & E-SR & V-GP & V-LP & V-SR & M+M-GP & M+M-LP & M+M-SR \\
 & (Acc) & (Acc) & (Acc) & (Acc) & (Acc) & (Acc) & (Acc) & (Acc) & (Acc) \\
\midrule

\multicolumn{10}{c}{\textbf{Metromap}} \\
\midrule
csv & 20.00 & 100 & 3.12 & 47.50 & \textbf{58.75} & 78.75 & \textbf{1.25} & 21.25 & \textbf{70.00} \\
json & \textbf{22.50} & \textbf{100} & \textbf{7.50} & \textbf{57.50} & 51.88 & \textbf{86.88} & 0.63 & \textbf{22.50} & 38.75 \\
\midrule

\multicolumn{10}{c}{\textbf{Travelmap}} \\
\midrule
csv & \textbf{19.64} & \textbf{99.40} & 44.05 & 25.00 & 47.02 & 54.17 & \textbf{76.19} & \textbf{72.02} & \textbf{16.07} \\
json & 17.86 & \textbf{99.40} & \textbf{45.24} & \textbf{38.69} & \textbf{50.60} & \textbf{61.31} & 75.60 & 70.24 & 14.29 \\
\bottomrule
\end{tabular}
\end{table*}

\begin{table*}[!ht]
\centering

\caption{Experimental performance results under different language distributions in the Metromap and Travelmap scenarios.}
\label{tab:language_perf}

\resizebox{\textwidth}{!}{
    \begin{tabular}{l cccccccccccc}
    \toprule

    \multirow{2}{*}{\textbf{Model and Criteria Settings}} & 
    \multicolumn{2}{c}{\textbf{Chinese-Chinese}} & 
    \multicolumn{2}{c}{\textbf{English-English}} & 
    \multicolumn{2}{c}{\textbf{Others-Others}} & 
    \multicolumn{2}{c}{\textbf{Others-Chinese}} & 
    \multicolumn{2}{c}{\textbf{Others-English}} & 
    \multicolumn{2}{c}{\textbf{English-Others}} \\
    
    \cmidrule(lr){2-3} \cmidrule(lr){4-5} \cmidrule(lr){6-7} \cmidrule(lr){8-9} \cmidrule(lr){10-11} \cmidrule(lr){12-13}
    
     & \textbf{EMA} & \textbf{PMA} & \textbf{EMA} & \textbf{PMA} & \textbf{EMA} & \textbf{PMA} & \textbf{EMA} & \textbf{PMA} & \textbf{EMA} & \textbf{PMA} & \textbf{EMA} & \textbf{PMA} \\
    \midrule
    
    
    Qwen3-map & 12.10 & 31.65 & 6.03 & 22.85 & 2.64 & 19.47 & 18.31 & 38.81 & 17.86 & 43.18 & 0 & 13.17 \\
    
    Qwen3-map+e\_tab & 30.78 & 47.77 & 31.03 & 56.41 & 29.92 & 53.61 & 41.36 & 60.79 & 57.14 & 71.25 & 4.17 & 33.70 \\
    
    Qwen3-map+e\_tab+v\_tab & 26.29 & 44.86 & 32.76 & 52.13 & 23.80 & 48.53 & 31.36 & 54.22 & 60.71 & 76.29 & 4.17 & 29.93 \\
    
    Qwen3-map+m\_tab & 10.59 & 31.30 & 7.76 & 27.07 & 5.12 & 26.07 & 15.25 & 39.19 & 17.86 & 48.41 & 0 & 18.23 \\
    \midrule
    
    
    GPT4o-map & 11.42 & 31.23 & 15.52 & 34.51 & 8.26 & 33.63 & 15.42 & 39.05 & 28.57 & 59.34 & 12.50 & 37.30 \\
    
    GPT4o-map+e\_tab & 49.71 & 63.21 & 55.17 & 75.92 & 46.61 & 73.51 & 61.86 & 76.77 & 96.43 & 98.66 & 58.33 & 76.84 \\
    
    GPT4o-map+e\_tab+v\_tab & 37.14 & 53.46 & 61.21 & 74.75 & 42.15 & 66.64 & 47.46 & 66.93 & 57.14 & 75.49 & 41.67 & 70.84 \\
    
    GPT4o-map+m\_tab & 9.49 & 30.23 & 25.00 & 46.14 & 14.05 & 40.25 & 13.56 & 40.51 & 14.29 & 47.60 & 46.67 & 35.21 \\
    \bottomrule
    \end{tabular}
}
\end{table*}

\subsection{Ablation study of Language}\label{subsec:abl_language}
\begin{figure*}[!ht]
    \centering
    \newcommand{\imgheight}{4.5cm} 
    
    \begin{subfigure}[b]{0.32\linewidth}
        \centering
        \includegraphics[height=\imgheight]{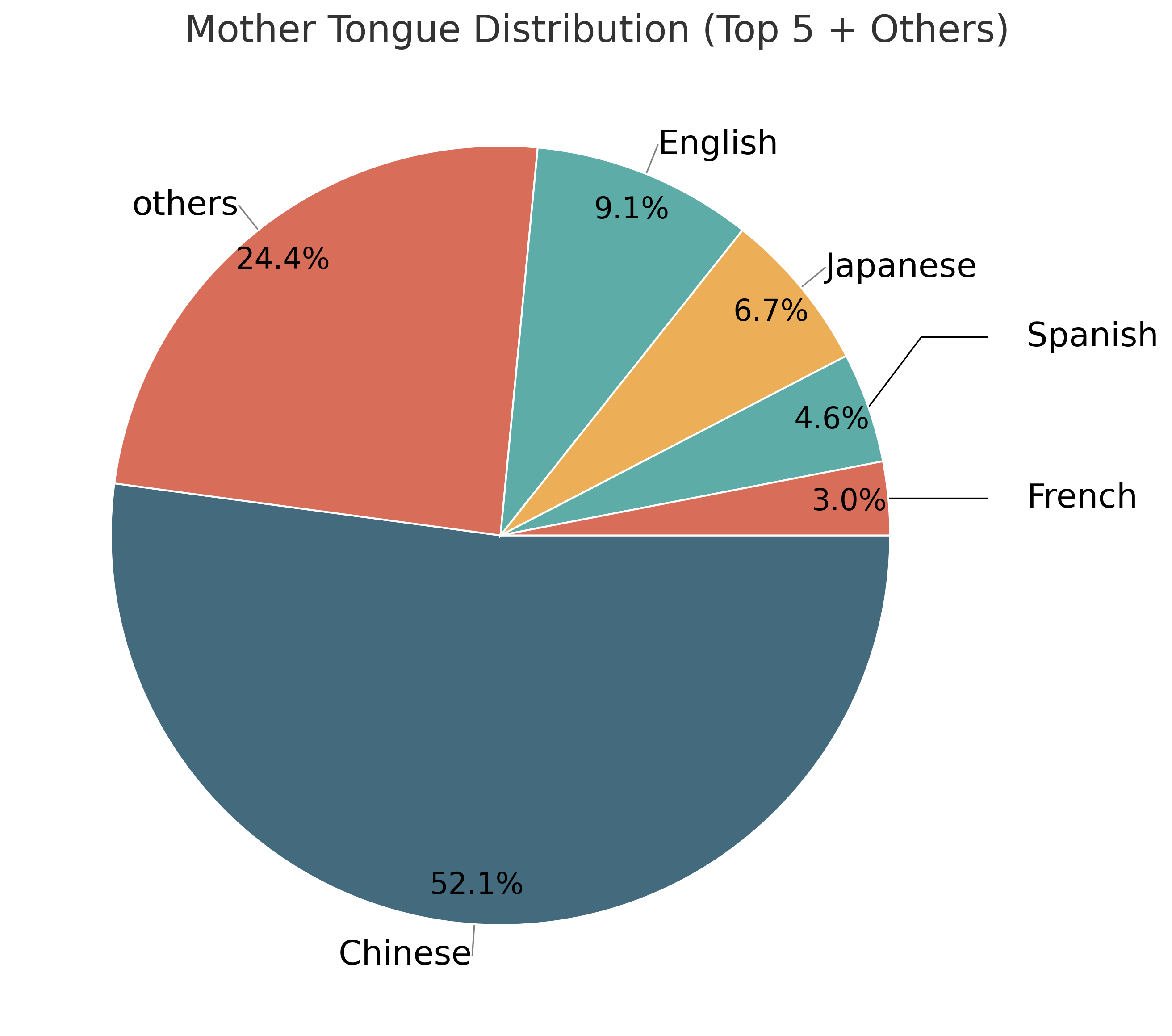}
        \caption{Native language}
        \label{fig:language_sec_a}
    \end{subfigure}
    \hfill 
    \begin{subfigure}[b]{0.32\linewidth}
        \centering
        \includegraphics[height=\imgheight]{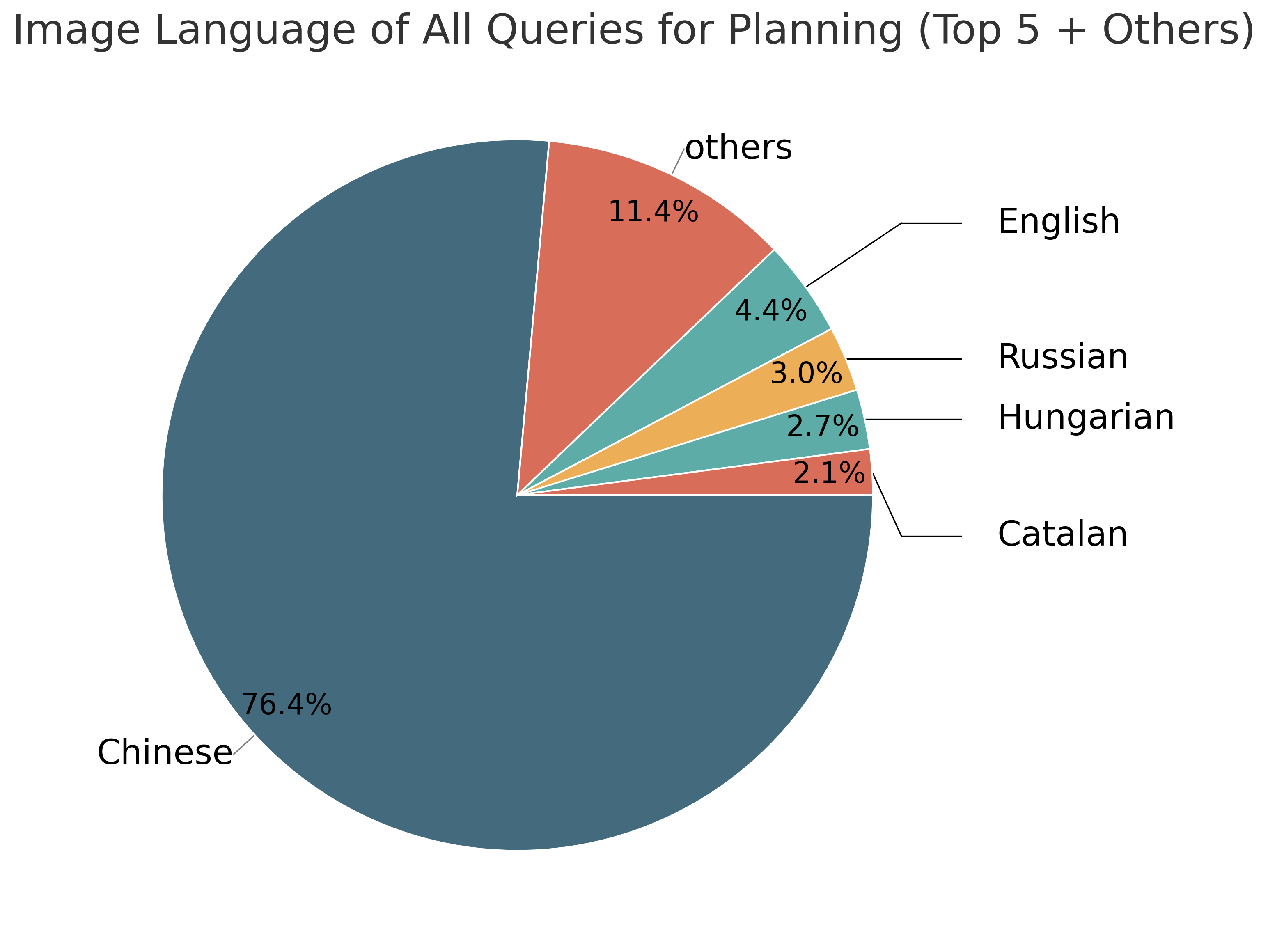}
        \caption{Actual language}
        \label{fig:language_sec_b}
    \end{subfigure}
    \hfill
    \begin{subfigure}[b]{0.32\linewidth}
        \centering
        \includegraphics[height=\imgheight]{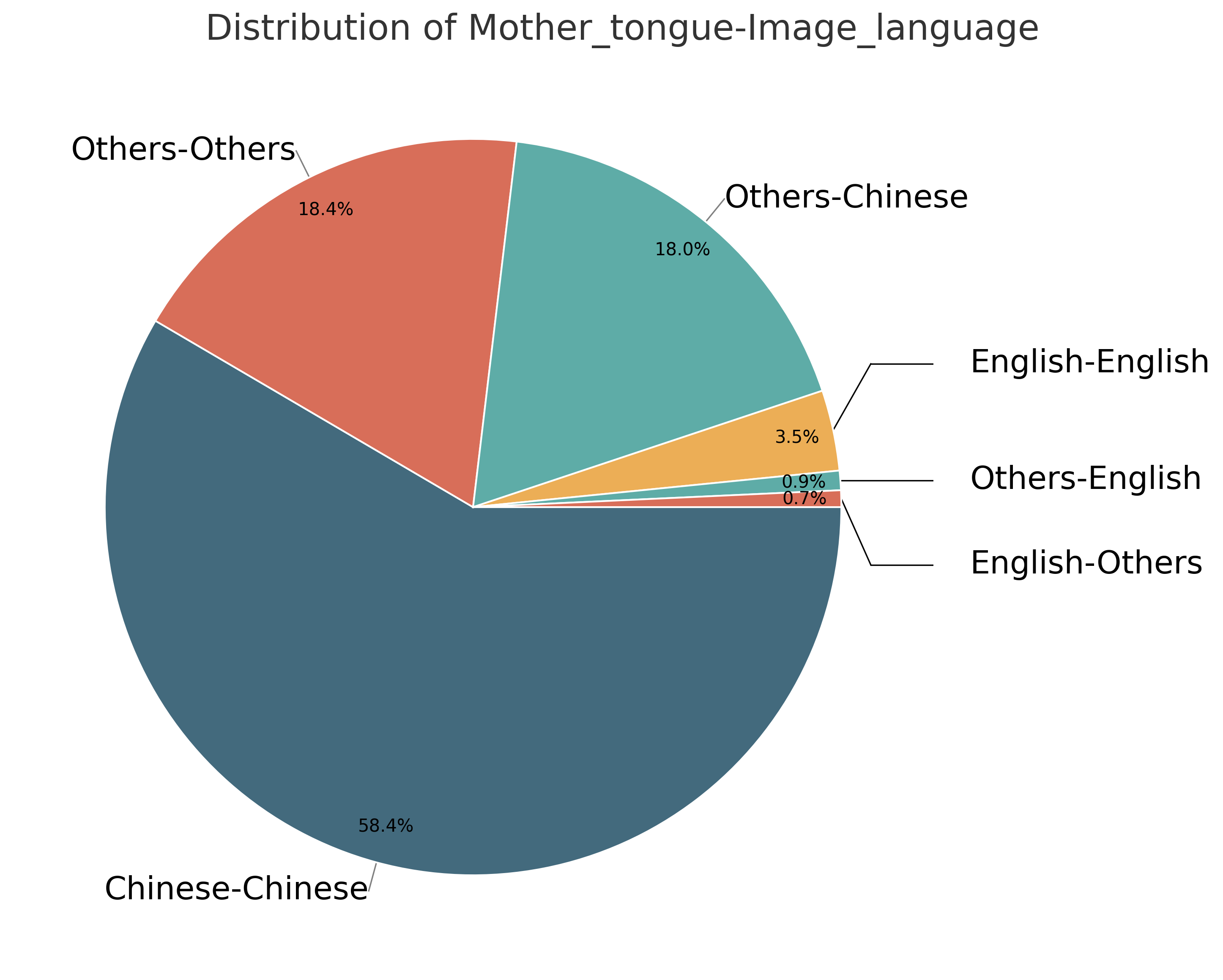}
        \caption{Combinations}
        \label{fig:language_sec_c}
    \end{subfigure}
    
    \caption{All images have the exact same height. Widths are adjusted automatically.}
    \label{fig:language_sec}
\end{figure*}

Figure \ref{fig:language_sec} presents the overall language distribution in the Metromap and Travelmap scenarios using a mixed statistical approach, retaining only the top five categories by proportion and aggregating the remaining categories. Figure \ref{fig:language_sec_a} shows the distribution of the native language of the locations corresponding to the images, covering a total of 328 images. The results indicate that locations with Chinese as the native language account for the highest proportion at 52.1\%, followed by English at 24.4\%. Figure \ref{fig:language_sec_b} further summarizes the actual language used in the images, also based on 328 images, showing a distribution trend similar to that of the native language statistics, with Chinese being the most frequently used and English second. Based on these observations, Figure \ref{fig:language_sec_c} categorizes the languages into Chinese, English, and Others, denoted as Chinese, English, and Others, respectively, and further divides them into nine combinations according to the mother\_tongue-image\_language pairing. Since the combinations Chinese-English, Chinese-Others, and English-Chinese do not appear in the data, the analysis focuses on the remaining six categories. The results show that the Chinese-Chinese combination has the highest proportion at 58.4\%.

Experimental results in Table \ref{tab:language_perf} indicate a significant language bias in model performance: across regional datasets, when the input uses a dominant language (Chinese or English), the model consistently performs better than in cases using other languages. This phenomenon is highly correlated with the frequency distribution of different languages in the training corpus, reflecting the model's stronger adaptation to high-frequency languages. Meanwhile, the model does not exhibit a significant language prior based on geographic or regional knowledge. When the native language of the location corresponding to the image does not match the language actually used in the data, model performance does not show systematic variation, indicating that the key factor affecting model performance is the language used in the sample itself rather than the native language attribute of the location.

\section{Error Case Analysis}\label{sec:err_analysis}
\begin{table*}[t]
    \centering
    

    \caption{
    Diagnostic criteria and corresponding capability gaps for different error types.
    The capability taxonomy includes Visual Perception, Table Understanding,
    Cross-modal Alignment, Graph Topology Reasoning, Spatial Localization,
    Numerical Reasoning, Path Planning, and Global Reasoning.
    }
    \label{tab:error-capability-analysis}
    
    \renewcommand{\arraystretch}{1.15}
    
    \newcolumntype{L}[1]{>{\raggedright\arraybackslash}m{#1}}
    
    \resizebox{\textwidth}{!}{
    \footnotesize
    \begin{tabular}{L{3.3cm}L{10.2cm}L{5.0cm}}
    
    \toprule

    \textbf{Error Type} &
    \textbf{Diagnostic Criterion} &
    \textbf{Capability Gap} \\
    
    \midrule
    
    Perception &
    
    The model fails to correctly recognize, localize, or associate task-critical visual entities (e.g., source and destination stations), resulting in incorrect inputs for subsequent route reasoning. &
    
    Visual Perception; Cross-modal Alignment \\
    \specialrule{0.3pt}{0pt}{0pt}
    
    Transfer Indeterminate &
    
    The predicted transfer annotations are inconsistent with the actual line-switching status along the route. &
    
    Graph Topology Reasoning \\
    \specialrule{0.3pt}{0pt}{0pt}
    
    Path Search &
    
    The model correctly identifies the optimization objective but fails to search for the optimal route, resulting in a path that deviates from the ground-truth solution. &
    
    Path Planning; Global Reasoning; Numerical Reasoning \\
    \specialrule{0.3pt}{0pt}{0pt}
    
    Topology Extraction &
    
    The recognized stations are correct, but the predicted route contains invalid graph connections or violates the underlying map topology. &
    
    Graph Topology Reasoning; Spatial Localization \\
    \specialrule{0.3pt}{0pt}{0pt}
    
    Output Format &
    
    The final response violates the required output specification, either because the model misunderstands the output requirements or because excessive reasoning prevents a valid final response. &
    
    Global Reasoning \\
    \specialrule{0.3pt}{0pt}{0pt}
    
    Table Lookup &
    
    The model fails to correctly interpret or retrieve the required information from \texttt{mix\_tab}, leading to an incorrect understanding of the tabular content. &
    
    Table Understanding \\
    
    \bottomrule
    
    \end{tabular}
    }

\end{table*}

Based on the experimental results, we conduct a systematic analysis of the major failure modes of current vision-language models on MapTab. After excluding anomalous outputs caused by repeated generation, answer collapse, or invalid long-form responses in certain small-scale open-source models~\cite{yao2025understanding}, we categorize all erroneous predictions according to the error taxonomy in Table~\ref{tab:error-capability-analysis} and the corresponding capability gaps summarized in Table~\ref{tab:error-capability-analysis}. For each error category, we further select a representative failure case to analyze its underlying cause and the associated capability limitation. The six representative error types include \textbf{Perception}, \textbf{Transfer Indeterminate}, \textbf{Path Search}, \textbf{Topology Extraction}, \textbf{Output Format}, and \textbf{Table Lookup}. Together, these representative cases provide a comprehensive view of the key challenges faced by current vision-language models in multimodal map reasoning.

\begin{figure*}[!ht] 
    \centering
    \includegraphics[width=0.8\textwidth]{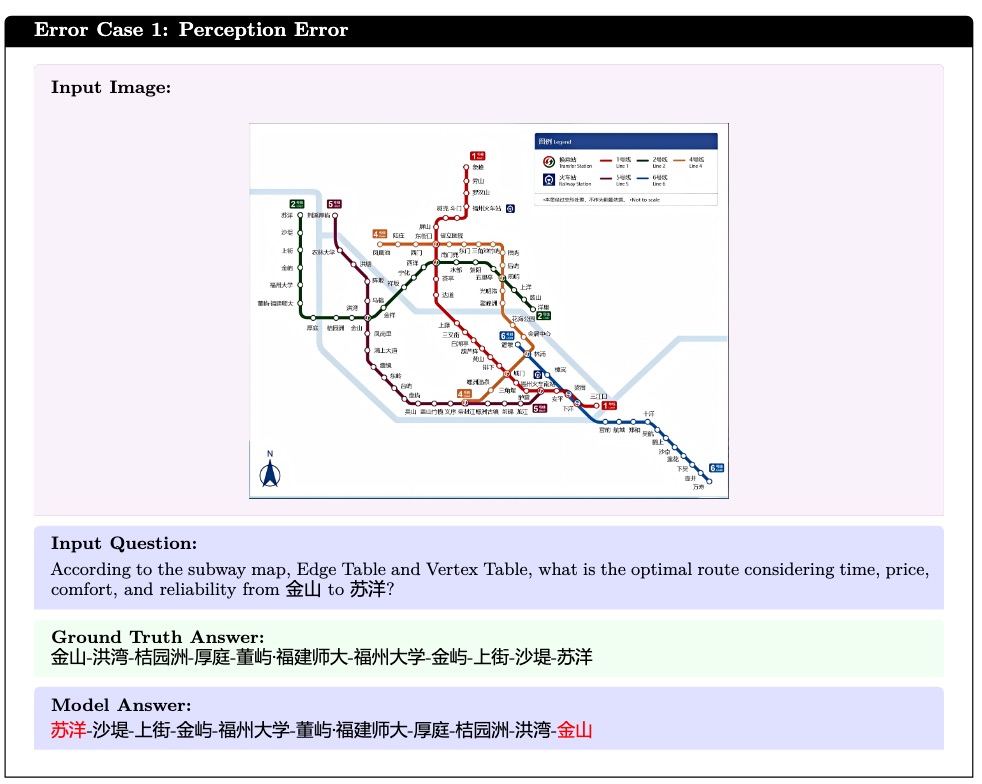} 
    \caption{Error Case 1: Perception Error}
    \label{fig:err_case1}
\end{figure*}

\textbf{Perception.}
Perception errors occur when the model fails to accurately perceive the task-critical visual information required for subsequent reasoning. As shown in Figure~\ref{fig:err_case1}, the task requires planning a route from Jinshan to Suyang, whereas the model instead generates the reverse route from Suyang to Jinshan. Since most intermediate stations remain consistent with the underlying topology, the failure does not originate from graph reasoning or path planning. Instead, the error is introduced at the earliest perception stage, where the model incorrectly identifies or associates the source and destination stations.

Unlike natural-image perception, perception in map reasoning is fundamentally symbolic. The model must simultaneously recognize station names, localize their positions, distinguish visually adjacent entities, and correctly associate the textual query with the corresponding map nodes. Dense layouts, small fonts, overlapping labels, and irregular spatial arrangements substantially increase the difficulty of this process. Consequently, even a small perception error, such as confusing the source and destination, propagates through the entire reasoning pipeline and inevitably leads to an incorrect route.

An important observation is that perception errors in MapTab are not limited to traditional OCR failures. They also include incorrect visual localization, ambiguous entity association, and inaccurate cross-modal alignment between textual queries and graphical symbols. Therefore, the primary capability missing from current vision-language models is not simply stronger text recognition, but more reliable fine-grained symbolic perception that can accurately identify, localize, and verify task-critical entities before higher-level graph reasoning begins.

\begin{figure*}[!ht] 
    \centering
    \includegraphics[width=0.8\textwidth]{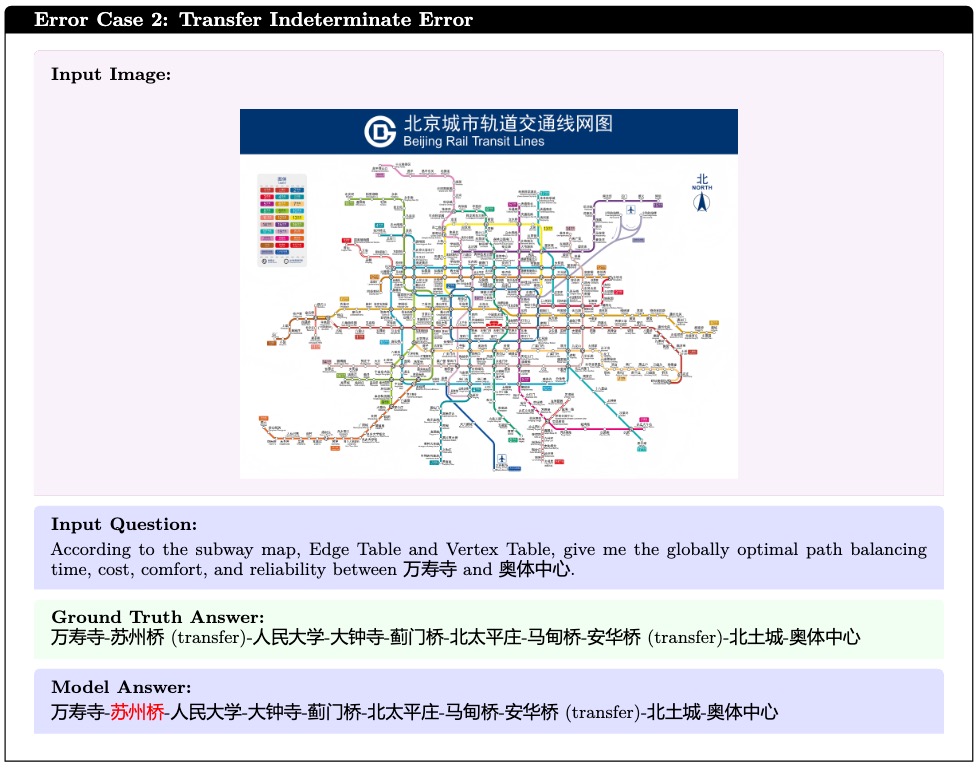} 
    \caption{Error Case 2: Transfer Indeterminate Error}
    \label{fig:err_case2}
\end{figure*}

\textbf{Transfer Indeterminate.}
Transfer Indeterminate errors occur when the model recovers a correct or nearly correct station sequence but fails to identify, localize, or annotate the transfer operation. As shown in Figure~\ref{fig:err_case2}, the predicted route is almost identical to the ground truth, but the model omits the required ``(transfer)'' label after Suzhou Street. Although the route itself is correct, the answer does not satisfy the task specification.

This error demonstrates that a correct station sequence does not necessarily imply a complete understanding of the route. Transfer identification requires the model to jointly maintain at least three types of information: whether adjacent stations belong to the same line, whether the current station performs a line-switching function, and where the output protocol requires the transfer annotation to be inserted. The task therefore goes beyond simple format compliance and requires the integration of \textbf{topological relations, line identity, and instruction constraints}. A model may know which station comes next without explicitly understanding whether a line transition has occurred.

Such errors reveal limitations in state tracking and relational attribute modeling. Current models are generally better at predicting node sequences than at maintaining the changing line state along a route. The main conclusion is that their graph understanding remains biased toward static nodes rather than dynamic path states. Reliable map reasoning requires models to track not only which stations are visited, but also which line is currently active, when the line changes, and how that change should be represented in the final answer.

\begin{figure*}[!ht] 
    \centering
    \includegraphics[width=0.8\textwidth]{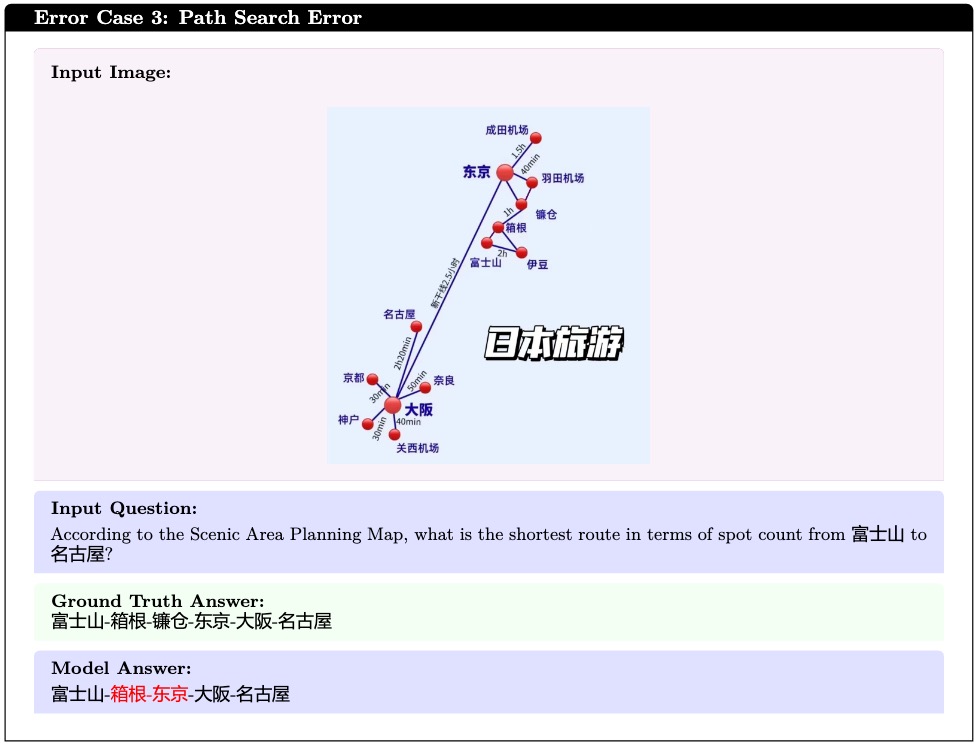} 
    \caption{Error Case 3: Path Search Error}
    \label{fig:err_case3}
\end{figure*}

\textbf{Path Search.}
Path Search errors occur when the model fails to recover the complete station sequence along a valid route. As shown in Figure~\ref{fig:err_case3}, the model predicts a route from Mount Fuji to Nagoya but skips one or more intermediate stations. Although the overall travel direction is correct and the predicted stations belong to a valid route, the generated path does not match the ground-truth station sequence because required intermediate stations are omitted.

This error differs from Topology Extraction. In Topology Extraction errors, the model introduces connections that do not exist in the underlying graph. In contrast, Path Search errors arise when the model does not fully traverse or expand a valid route, resulting in omitted intermediate stations. The model may correctly recognize the source, destination, and major waypoints, but it fails to maintain a complete step-by-step search over the graph, leading to an incomplete station sequence.

This behavior indicates limitations in \textbf{path planning, global reasoning, and numerical reasoning}. Rather than systematically traversing the graph and expanding every intermediate station, vision-language models tend to generate only a coarse route guided by prominent landmarks or local associations. Consequently, they are prone to skipping valid intermediate stations and producing an incomplete path even when the overall route direction is correct.

\begin{figure*}[!ht] 
    \centering
    \includegraphics[width=0.8\textwidth]{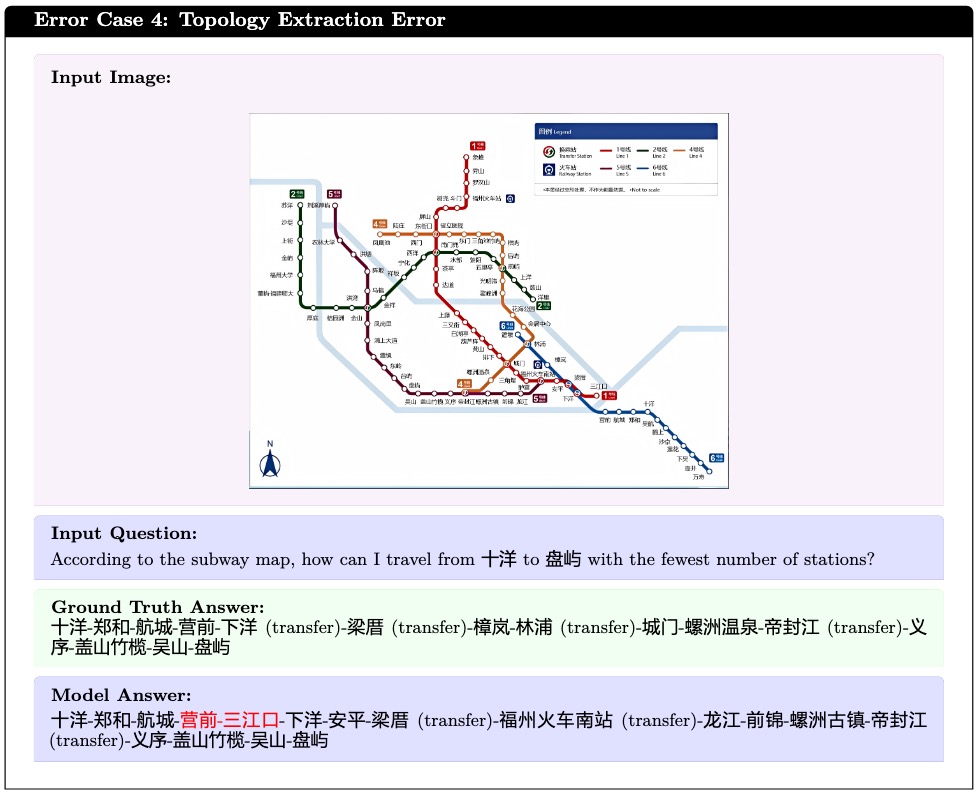} 
    \caption{Error Case 4: Topology Extraction Error}
    \label{fig:err_case4}
\end{figure*}

\textbf{Topology Extraction.}
Topology Extraction errors occur when the model incorrectly reconstructs the connectivity of the visual map and introduces an edge that does not exist in the underlying graph. Figure~\ref{fig:err_case4} illustrates a typical illegal line-jumping case, in which the model directly moves from one line to another line that is not connected at that location. Unlike Path Search errors, which omit one or more intermediate stations along an otherwise valid route, Topology Extraction errors violate the underlying graph structure itself.

Map images organize information very differently from natural images. Natural images represent objects through continuous textures, contours, and regional features, whereas subway and tourist maps encode discrete symbolic relations using thin lines, nodes, colors, intersections, and relative positions. Two lines may visually cross without providing a valid transfer, and a station label may be spatially close to a line without belonging to it. The model must therefore distinguish visual proximity, geometric intersection, and actual topological connectivity.

These errors show that even strong closed-source models cannot always convert a high-resolution map into an accurate graph structure. The missing capability is \textbf{symbol-centered visual parsing}: the model must continuously trace the same line, determine which nodes belong to which edges, and identify whether an intersection is accompanied by an explicit transfer symbol. The main conclusion is that the key visual bottleneck in map reasoning is not general object recognition, but the recovery of high-density discrete topology from visually sparse graphical features.

\begin{figure*}[!ht] 
    \centering
    \includegraphics[width=0.8\textwidth]{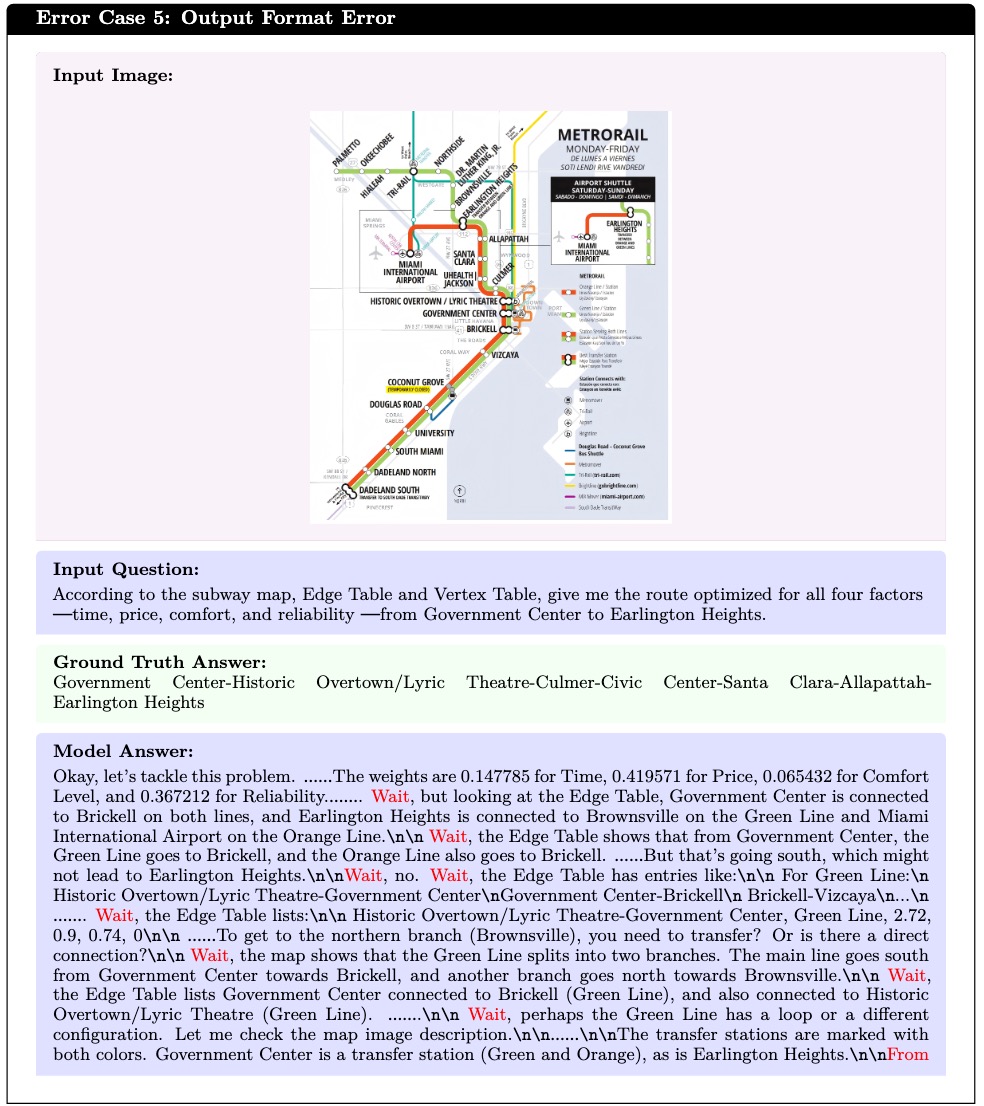} 
    \caption{Error Case 5: Output Format Error}
    \label{fig:err_case5}
\end{figure*}

\textbf{Output Format.}
Output Format errors occur when the model fails to produce the final answer in the required format and instead exposes an incomplete or excessively long reasoning process. As shown in Figure~\ref{fig:err_case5}, Qwen3-VL-Plus can directly return the correct route when explicit reasoning is disabled. After the reasoning mechanism is enabled, however, the model performs up to 25 \texttt{wait} operations, repeatedly questioning previous decisions, revising intermediate conclusions, and switching among different reasoning directions. Rather than terminating with a concise route in the prescribed format, the model outputs a lengthy thinking process that does not satisfy the evaluation protocol.

The direct cause of this error is overthinking. For a relatively simple instance, the model continues reasoning after sufficient evidence has already been obtained. The repeated self-correction not only increases inference cost, but also prevents the model from reliably transitioning from internal reasoning to final-answer generation. Therefore, the failure is not merely a superficial formatting mistake, such as using an incorrect separator. It reflects a deeper deficiency in \textbf{reasoning termination and output control}: the model cannot determine when to stop reasoning and when to convert the current conclusion into the required answer format.

This case shows that explicit reasoning can reduce performance when the model lacks an effective termination mechanism. Longer reasoning does not necessarily produce a better answer; instead, unnecessary continuation may cause the model to expose intermediate thoughts, exceed the expected output structure, or fail to generate a valid final response altogether. The key missing capability is therefore the ability to adapt reasoning depth to task difficulty and to reliably separate the internal thinking process from the final structured answer.

\begin{figure*}[!ht] 
    \centering
    \includegraphics[width=0.8\textwidth]{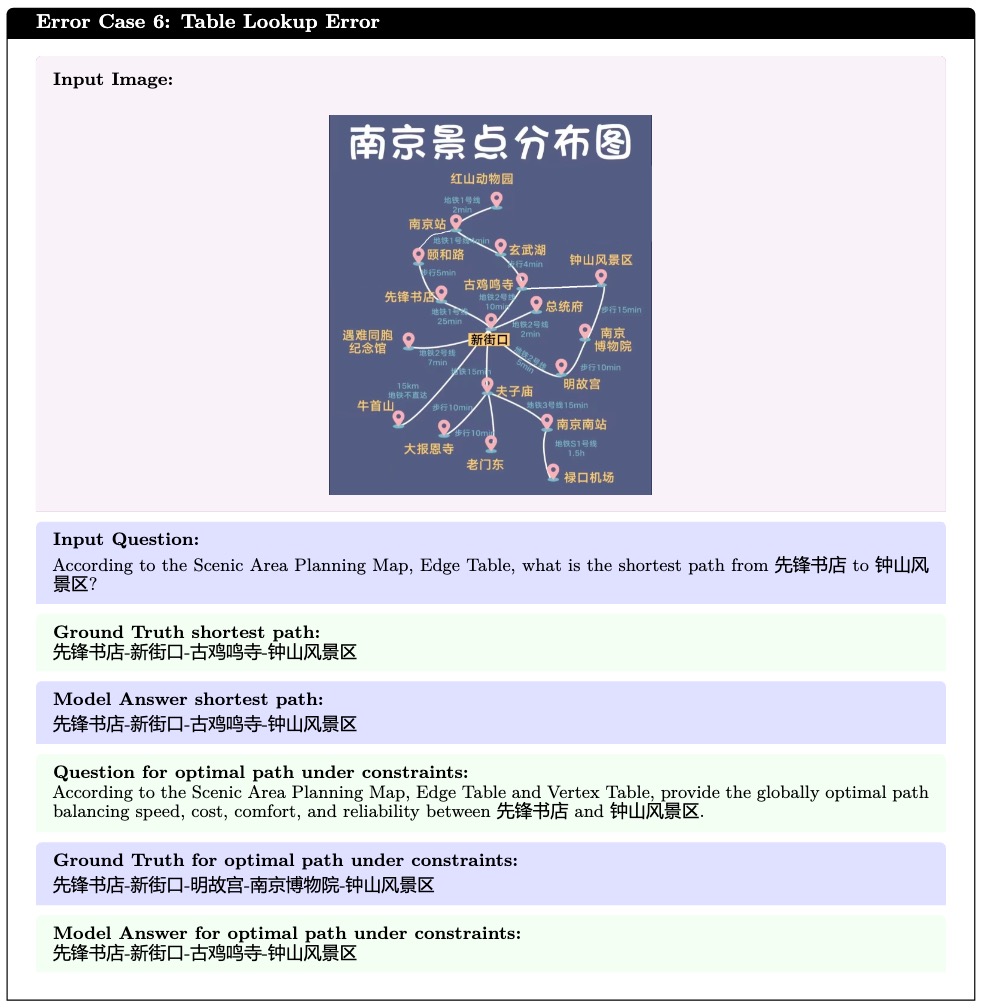} 
    \caption{Error Case 6: Table Lookup Error}
    \label{fig:err_case6}
\end{figure*}

\textbf{Table Lookup.}
Table Lookup errors occur when the model fails to correctly understand the information provided in \texttt{mix\_tab}. As shown in Figure~\ref{fig:err_case6}, the model correctly recognizes the relevant stations and recovers the associated route, indicating that visual perception and graph reasoning are largely successful. However, it fails to correctly understand the tabular content required by the query, leading to an incorrect final answer.

This behavior suggests that the primary difficulty lies in understanding structured tabular information rather than in route planning itself. The model may fail to locate the required entries, associate the correct rows or columns with the queried entities, or understand the semantics of the attributes presented in \texttt{mix\_tab}. As a result, even when the map-related reasoning is largely correct, misunderstanding the table still leads to an incorrect prediction.

This error indicates that current vision-language models still exhibit limited \textbf{table understanding}. Although they can often reason over visual map topology, they remain less effective at retrieving, associating, and interpreting structured information from tables. Therefore, improving table understanding remains an important direction for multimodal route reasoning.

\textbf{Overall}, the six error types reveal a clear break in the capability chain required by MapTab. A model must first recognize task-relevant entities and reconstruct a legal topology from the map, then recover a complete route, identify transfer states, understand the associated tabular information, and finally produce a stable answer under the required protocol. Current models may perform well at an isolated stage, yet they rarely execute the complete chain reliably. More importantly, errors at an early stage can propagate to all subsequent reasoning steps, while excessive reasoning may destroy an otherwise correct solution. These findings suggest that future progress requires more than stronger visual encoders or longer chains of thought. More structured, executable, and verifiable mechanisms are needed for symbolic perception, map-topology extraction, route expansion, transfer-state tracking, table understanding, and final-answer verification.


\section{Limitations}\label{sec:limitations}
Despite providing a large-scale multimodal benchmark for multi-criteria route planning, this study still has several limitations:

\begin{enumerate}
    \item \textit{Limited Map Diversity.} The current version of MapTab mainly focuses on topological map scenarios, including metro maps and scenic-area maps. Although these two scenarios cover different graph structures and route-planning requirements, they cannot fully represent the diversity of real-world cartographic data. More complex map types, such as large-scale urban road networks, indoor maps, campus maps, and remote sensing images, may contain denser visual elements, irregular spatial layouts, and richer semantic information. Extending MapTab to these map types would enable a more comprehensive evaluation of visual perception, topology reconstruction, and spatial decision-making across different domains.

    \item \textit{Partially Synthetic Tabular Data.} The current benchmark is only partially grounded in real-world data. While the map images are collected from real-world scenarios, the numerical attributes provided in the accompanying tables, such as time, price, comfort, reliability, and transfer time, are synthetically generated according to predefined ranges and heuristic rules. This design enables controllable multi-criteria evaluation and prevents models from exploiting external knowledge, but it may not fully capture the correlations, distributions, and irregularities of real transportation and tourism data. Future versions of MapTab will incorporate numerical attributes obtained from real-world sources, such as tourism APIs, public transportation APIs, route-planning services, ticketing platforms, and travel-review systems. Combining real map images with real-time or historically recorded tabular attributes would support the construction of fully realistic multi-criteria route-planning tasks.

    \item \textit{Lack of Dynamic Evaluation.} The current evaluation is based on static maps and predefined criteria, while real-world route planning often involves continuously changing conditions. For example, travel time, ticket price, congestion, service reliability, attraction availability, and transfer conditions may vary over time. Moreover, the underlying graph structure may also change because of road closures, service interruptions, or temporary route adjustments. Future work should therefore introduce temporally evolving graph structures and attributes, requiring models to update their decisions when new information becomes available. Such settings would provide a more realistic evaluation of adaptive planning and decision revision.

    \item \textit{Remaining Limitations of Model Reasoning.} Experimental results show that current MLLMs still exhibit substantial weaknesses in complex visual perception, topology extraction, cross-modal alignment, multi-criteria interpretation, numerical computation, and multi-step route reasoning. Errors introduced during early perception or topology reconstruction may propagate through subsequent stages and ultimately lead to incorrect route decisions. In addition, current models often rely on locally salient or shortest-looking paths rather than systematically comparing all feasible candidates under the specified criteria. These findings suggest that improvements are still required not only in individual perceptual and reasoning capabilities, but also in the stable coordination of these capabilities throughout the complete planning process.
\end{enumerate}

Overall, these limitations indicate that MapTab remains an initial step toward fully realistic multimodal route-planning evaluation. Future extensions should broaden map diversity, incorporate real-world and dynamically updated attributes, and further examine how MLLMs coordinate perception, topology understanding, numerical reasoning, and multi-step planning in complex environments.

\section{Future Work}\label{sec:future_work}

Our findings suggest several promising directions for improving MLLMs on multimodal multi-criteria planning tasks:

\begin{enumerate}
    \item \textit{Modular collaborative frameworks that decouple perception from reasoning.} Future systems could separate visual perception, topology extraction, structured representation, cross-modal alignment, and route optimization into specialized modules~\cite{gou2025reasoning,avogaro2026sparc,wang2025perception}. Since these stages exhibit different failure patterns, independent optimization and verification could reduce error propagation. Passing confidence scores or alternative interpretations between modules may further allow later stages to reconsider uncertain perception results, improving robustness, interpretability, and error traceability.

    \item \textit{Agentic reasoning with selective tool use.} MLLMs could be equipped with tools for OCR, topology extraction, graph construction, numerical calculation, path search, and route verification~\cite{chng2025sensenova,ning2026mc,song2025codedance}. These tools may help address the counting, attribute aggregation, structured search, and multi-step computation errors identified in MapTab. Future work should examine when tools are necessary and whether models can correctly select them, construct valid inputs, and integrate their outputs without replacing native reasoning.

    \item \textit{Targeted post-training based on benchmark failure modes.} Supervised fine-tuning, reinforcement learning, and RLHF could directly target the weaknesses revealed by MapTab~\cite{tan2025reason,liu2025acereason,chen2025beyond}. SFT data could include demonstrations of topology reconstruction, map--table alignment, criterion aggregation, route comparison, and transfer-aware reasoning. Reinforcement learning could reward route correctness, valid topology, accurate numerical computation, and consistent use of multimodal evidence. A curriculum from atomic QA tasks to complete RP tasks may further improve capability coordination.

    \item \textit{Fully realistic and dynamically updated benchmark construction.} The current map images are collected from real-world scenarios, while tabular numerical attributes are generated using controlled heuristic rules. Future versions of MapTab could obtain time, price, accessibility, reliability, opening hours, and congestion information from tourism APIs, transportation APIs, route-planning services, and public data platforms. Periodically updated data could further support dynamic tasks in which graph attributes and feasible routes change over time, requiring models to revise decisions and handle missing or conflicting information.
\end{enumerate}

Together, these directions extend MapTab along both the model and benchmark dimensions. Modular reasoning, selective tool use, and targeted post-training may improve model reliability, while real-world APIs and dynamic data can support more realistic evaluations of multimodal multi-criteria planning.

\clearpage

\end{document}